\documentclass[letterpaper]{article} 
\usepackage[preprint]{aaai2027}  
\usepackage[hyphens]{url}  
\usepackage{graphicx} 
\usepackage{natbib}  
\usepackage{caption} 
\usepackage{algorithm}
\usepackage{algorithmic}

\usepackage{newfloat}
\usepackage{listings}
\usepackage{subcaption}
\usepackage{multirow}
\usepackage{longtable}
\newif\ifarxiv
\arxivtrue

\DeclareCaptionStyle{ruled}{labelfont=normalfont,labelsep=colon,strut=off} 
\floatstyle{ruled}
\newfloat{listing}{tb}{lst}{}
\floatname{listing}{Listing}

\usepackage{booktabs}

\usepackage{amsmath}
\usepackage{amsfonts}

\title{Kastor: An Efficient Fine-Tuning Strategy \\
for Generative Emulation of PDE Simulations}
\author{Guillaume Couairon, Alexis D. Jacq, Yu-Han Wu, Renu Singh,\\ Yana Hasson, Quentin Berthet, Romuald Elie}
\affiliations{Google DeepMind}

\begin{document}

\maketitle

\begin{abstract}

Machine learning offers a promising avenue to accelerate physical simulations by replacing computationally expensive traditional Partial Differential Equation (PDE) solvers with fast, differentiable surrogate models. However, standard auto-regressive ML emulators often suffer from error accumulation over long horizons and struggle to capture the stochasticity of complex physical systems. In this paper, we propose \textit{Kastor}, a comprehensive methodology to adapt a deterministic physics foundation model into a highly efficient and accurate generative surrogate. First, we introduce a two-stage inference scheme that combines a large-stride causal auto-regressive model with a non-causal temporal super-resolution network, significantly reducing error accumulation while minimizing computational cost. Second, we present \textit{Mean prediction regularization} (MPR), a novel training objective that constrains the generative model to predict the deterministic distribution mean under null noise conditioning. This regularization dramatically improves the performance and stability of both Functional Generative Networks (FGN) and diffusion-based emulators. Finally, we demonstrate that incorporating spatial gradient matching improves the accuracy and physical fidelity of the simulations as measured by power spectrum density. Extensive evaluations on diverse simulation datasets of the benchmark \textit{The Well} show that with these components, our model outperforms competing methods in forecasting accuracy, spectral consistency, and computational efficiency. 
Our model achieves a $42.9\%$ average reduction in forecasting error compared to our reference based on the Walrus finetuning methodology, and outperforms Walrus for 8 out of 10 datasets on variance-normalized RMSE (VRMSE).


\end{abstract}

\section{Introduction}

\begin{figure}[ht]
    \centering
    \includegraphics[width=\columnwidth]{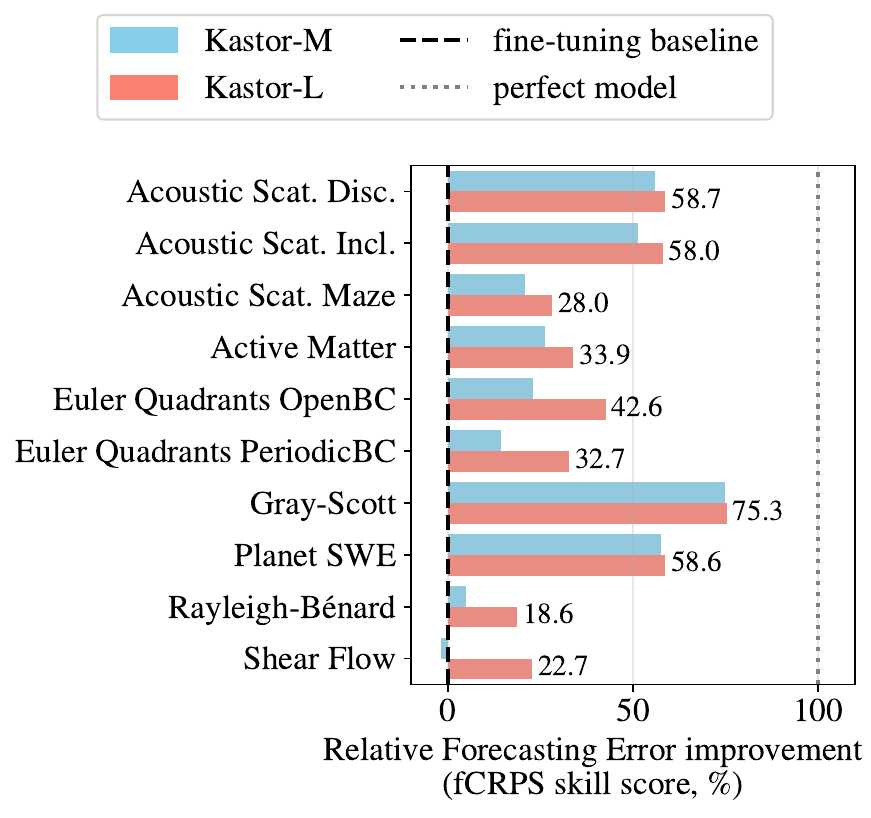}
    \caption{Average relative reduction of forecasting error (fCRPS skill score) between our models and the baseline fine-tuning methodology from Walrus (higher is better). Models trained with our methodology reduce forecasting error by $42.9\%$ on average across datasets and physical variables.}
    \label{fig:main_spider}
\end{figure}

Physical simulation is a cornerstone of modern science, critical for both understanding complex phenomena \cite{dunne2025evolving} and engineering efficient designs \cite{jenko2025accelerating}. Standard Finite Element or Finite Difference methods require fine spatial and temporal meshes to capture complex physics, demanding massive computational costs \cite{morton2005numerical}. Leveraging Machine Learning for Partial Differential Equation (PDE) simulations offers to bypass the computational bottlenecks of traditional numerical solvers by learning the underlying continuous mappings of physical systems \cite{ashton2025fluid, lam2023learning}. ML-based models act as fast surrogate solvers, and can also be used for rapid iterative design loops in engineering and multi-physics optimization \cite{liu2022surrogate, li2024data}.
In this paper, we focus on Neural PDE emulators, that learn the temporal mapping between the state $x_t$ (and optionally additional previous states $x_{t'}$ with $t'<t$) to the future state $x_{t+\Delta t}$ corresponding to the integration of the PDE over the time period $[t, t+\Delta t]$. For training and evaluating models, we use numerical simulations data from \textit{The Well} \cite{ohana2024well}, which contains 19 benchmark datasets covering diverse domains such as biological systems, fluid dynamics and acoustic scattering.
Once trained, ML models can be used auto-regressively to generate trajectories by feeding back forecast states as input states.

To be useful, a surrogate model should satisfy the following constraints:
\begin{itemize}
\item Reproduce the dynamics of the trajectories seen during training, as faithfully as possible;
\item Generalize across unseen initial conditions, boundary conditions, and PDE hyper-parameters values;
\item Generate \emph{stable} rollouts, that remain physically plausible over extended time horizons;
\item Produce correct statistical distributions of simulated physical variables, in particular the power spectrum density;
\item Provide a \emph{set} of possible trajectories given an initial condition, capturing the stochasticity of the system or the uncertainty of the learned model;
\item Significantly reduce computational cost in comparison to the reference numerical solver which it emulates.

\end{itemize}

These requirements are fundamentally challenging, and previous works have sought improvements across distinct areas, such as Fourier Neural Operators \cite{kovachki2023neuraloperator, ahn2025lightweight} for long-rollout stability, foundation models for generalization \cite{mccabe2024multiple, mccabe2025walruscrossdomainfoundationmodel} and generative approaches to enable probabilistic modeling \cite{diaconu2026probabilistic, kassai2026enma}.
In this paper, we propose a strategy to derive strong generative surrogate models from pretrained foundation models, with a particular focus on accuracy, computational efficiency, and statistical consistency.


\paragraph{Efficient inference strategy.} 
ML emulators typically predict states one by one via auto-regressive rollouts. This strategy has many advantages: it focuses the model’s capacity on learning a single state, and allows training causal models efficiently via teacher forcing. However, this strategy can lead to artifact accumulation \cite{mukhopadhyay2026overtone, liu2025bcat} and divergence. We propose a two-stage auto-regressive approach with a causal model that predicts states sub-sampled in time, and a lightweight up-sampling model that resolves the missing states.

\paragraph{Gradient Difference Loss (GDL).} 
The standard method for predicting physical fields with neural networks is Mean Absolute Error (MAE) between the forecast and ground truth. However, derived quantities such as spatial gradients are often imperative for governing the dynamics. To explicitly preserve these features over long horizons, we propose to use the Gradient Difference Loss \cite{mathieu2015deep} to match the spatial gradients between the forecast and ground truth states \cite{czarnecki2017sobolev}.

\paragraph{Generative modeling from foundation models.} 
Physics foundation models, pretrained on a large variety of PDE simulations, greatly improve generalization upon a model trained from scratch \cite{ren2026foundation}. In this paper, we leverage the strongest foundation model up to date, Walrus \cite{mccabe2025walruscrossdomainfoundationmodel}. We turn it into a generative model by training with the Continuous Ranked Probability Score (CRPS) scoring rule~\cite{FGN_Alet2025SkillfulJP,diaconu2026probabilistic}, injecting noise into the model with Adaptive Layer-Normalization (AdaLN) \cite{perez2018film} and using patch jittering as an additional source of stochasticity.

\paragraph{Mean Prediction Regularization (MPR).} 
In contrast to natural video generation, which exhibits large inter-sample variability, generative components in PDE emulation often have a smaller magnitude relative to the underlying deterministic dynamics.
To successfully generate both accurate and realistic physical fields, we motivate and propose a form of regularization where the network both produces generative samples from the distribution $p(x_{t+\Delta t}|x_{t-k:t})$ and estimates the distribution mean, 
improving stability and accuracy.

In summary, our technical contributions are:

\begin{itemize}
    \item We propose a two-stage inference scheme to improve accuracy and efficiency in auto-regressive rollouts.
    \item We introduce \textit{Mean prediction regularization}, a regularization objective for generative models, and demonstrate its efficacy on Functional Generative Models (FGN) and diffusion models.
    \item We demonstrate that training a model by matching spatial gradients of the output increases accuracy while improving power spectrum density and reducing artifacts.

\end{itemize}

With all these contributions combined, the performance of our models reduce the forecasting error (CRPS) by 42.9\% compared to the reference fine-tuning methodology \cite{mccabe2025walruscrossdomainfoundationmodel}, on average across ten 2D benchmark datasets from \textit{The Well} \cite{ohana2024well} (see Figure~\ref{fig:main_spider}).

\begin{figure*}[t]
\centering
\includegraphics[width=\linewidth]{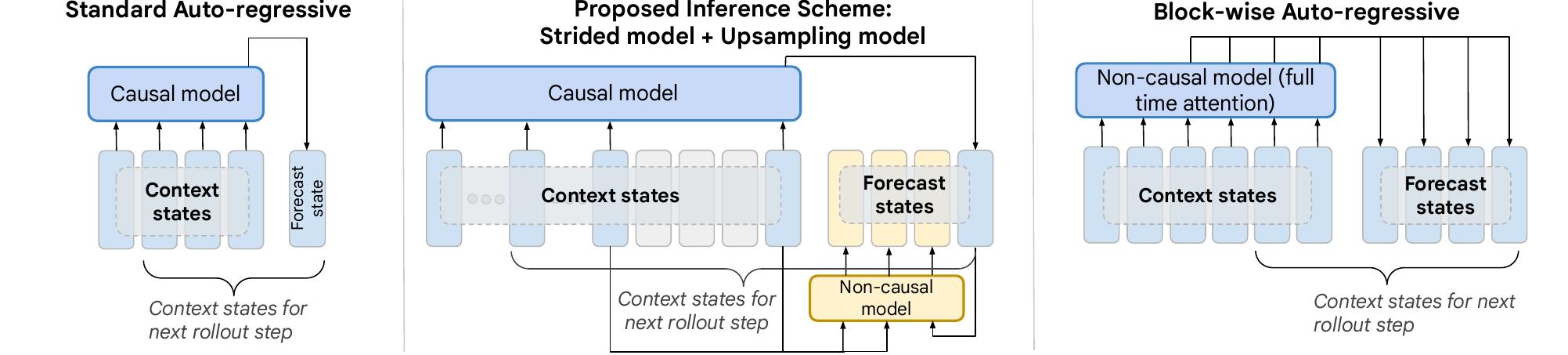}
\caption{Diagram of our proposed method to recover full trajectories with non-causal temporal super-resolution (middle), compared to the baseline with time stride $\tau=1$ (left), and the block-wise auto-regressive strategy (right).}
\label{fig:inference_strat_figure}
\end{figure*}

\section{Related Work}

\paragraph{Physics foundation models.} The push for generalization has led to the widespread development of massive foundation models \cite{nguyen2025physix, mccabe2024multiple, mccabe2025walruscrossdomainfoundationmodel, ren2026foundation}, capable of transferring priors across multi-modal PDE representations \cite{rautela2025morph, liu2024prose}, varying equation families \cite{wiesner2025towards, ye2025pdeformer}, and complex spatiotemporal domains \cite{soares2025towards}. Such models are commonly trained as sequence-to-sequence transformers with a next-state prediction objective \cite{mccabe2024multiple}, such as Walrus \cite{mccabe2025walruscrossdomainfoundationmodel} that we leverage in this paper.

\paragraph{Generative modeling for spatio-temporal data.} To capture intrinsic stochasticity, the field has increasingly adopted generative models as PDE surrogates \cite{diaconu2026probabilistic, kassai2026enma}, utilizing latent diffusion \cite{rozet2026lost}, flow matching \cite{archesweather}, or noise-conditioned functional generation \cite{FGN_Alet2025SkillfulJP}.

\paragraph{Adapting deterministic models to generative models.} A few works have investigated how to adapt existing models into generative models. In U-Cast \cite{cachay2026u}, authors propose to further fine-tune deterministic models using CRPS and dropout for stochasticity, which we evaluate in this work. Probabilistic retro-fitting \cite{diaconu2026probabilistic} injects noise via Adaptive Layer-Normalization, similarly to us, but without leveraging patch jittering.

\paragraph{Rollout stability.} To enforce stability over long rollouts, structural innovations have been proposed for Fourier Neural Operators \cite{kovachki2023neuraloperator, ahn2025lightweight},
alongside auto-regressive correction schemes like cyclic patch modulation \cite{mukhopadhyay2026overtone} and block causal masking \cite{liu2025bcat}, or even test-time operator composition \cite{serrano2026test, morel2025disco}.

\section{Methods}

\paragraph{General framework.}
We use the Walrus architecture \cite{mccabe2025walruscrossdomainfoundationmodel} as the backbone for our models. This allows us to leverage the pretrained weights of this strong foundation model.
Walrus is a model with a convolutional encoder, a causal transformer with spatial attention and time attention, and a convolutional decoder. It takes a sequence of $N_\text{in}$ states of input $x_{t-k:t} = (x_{t-k}, x_{t-(k-1)}, ..., x_t)$ (with $k=N_\text{in}-1$) and is trained to predict the normalized state difference $d_{t-k+1:t+1}$ with teacher forcing, where $d_{t+1} =(x_{t+1} - x_t) * \sigma_\text{res}^{-1}$ and $\sigma_{\text{res}}$ is a per-variable normalization factor. At inference time, the model is used auto-regressively from a given set of states (context) to make trajectory forecasts. The time index difference between a given forecast state and the last context state is called the \textit{lead time}.

Walrus also introduced \textit{patch jittering}, where input states are randomly shifted along spatial dimensions (with some padding for dimensions with non-periodic boundary conditions) before processing, and then shifted back. This addresses the accumulation of "grid-imprinting" artifacts on long horizons, and introduces stochasticity in the model that we leverage for our generative models.

\begin{figure*}
    \includegraphics[width=\linewidth]{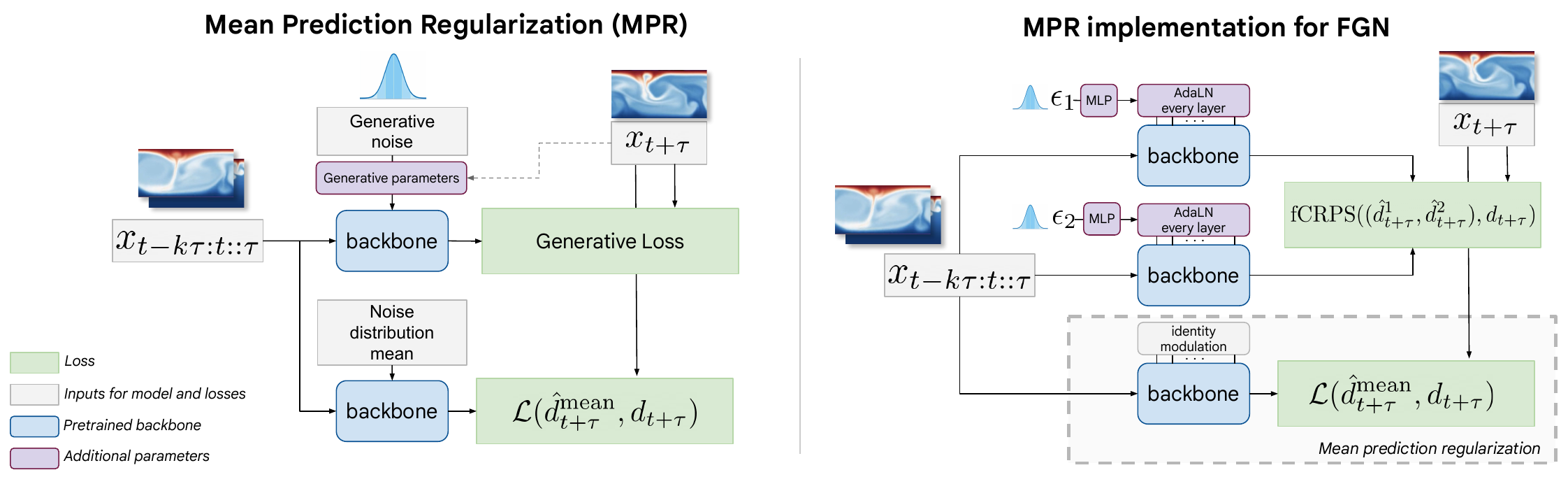}
\caption{Diagram of Mean Prediction Regularization principle (left) and its implementation with FGN (right). The training loss $\mathcal{L}$ is either Mean Absolute Error (MAE) or a combination of MAE and our Gradient Difference Loss (GDL).}
\label{fig:generative_methods}
\end{figure*}

\subsection{Training and Inference strategy}

\paragraph{Time stride.} 

A simple strategy to improve computational efficiency is to sub-sample states in time, with a stride parameter $\tau$, so that the model is trained to forecast $x_{t+\tau}$ from $x_{t-k\tau:t::\tau}$. On one hand, this reduces the number of auto-regressive rollout steps needed to reach a given lead time $T$, which mitigates the risks of error accumulation and divergence; on the other hand, it requires the model to learn a more complicated transition function, with less inference-time compute budget, and a higher distribution shift when re-using model outputs for the next rollout steps. We evaluate this trade-off in the Results section.

If the primary goal is maximizing accuracy at a specific lead time while lowering inference costs, sub-sampling alone is sufficient.
For use-cases where complete trajectories are still required, such as computing time-integrated statistics, we introduce a two-stage approach that generates complete trajectories both efficiently and accurately.

\paragraph{Two-stage approach.} 
We first train a causal model with a time stride $\tau$, typically $\tau=4$.
We then train a temporal up-sampling model to fill in missing states, leaving the causally predicted boundary states untouched. For instance with lead time $\tau=4$, 3 missing states $x_{t+1:t+3}$ are predicted using 3 input states $(x_{t-4}, x_t, x_{t+4})$. Being non-causal allows every predicted state to use all 3 inputs as context.

This strategy is illustrated in Figure \ref{fig:inference_strat_figure}. These two stages can be performed sequentially: the first model can predict a complete sub-sampled sequence, and the second model can up-sample that sequence in time.
We compare it to the more costly default auto-regressive strategy (time stride $\tau=1$) and a block-wise auto-regressive strategy that predicts $N_\text{out} = \tau$ forecast states at once, which splits model capacity into learning a set of states instead of learning only state $x_{t+\tau}$.
Our model improves computational efficiency by reducing the number of auto-regressive rollout steps, even accounting for the upsampling model. To lighten inference, the up-sampling model is half the size of the causal model. At long lead times, the resulting cost is reduced by factor $\sim 37.5\%$ (details in appendix).

\subsection{Mean Prediction Regularization}
\label{subsec:methods_mpr}

We now present Mean Prediction Regularization (MPR).
Our motivation is to help the model better capture the main deterministic mode of the next state $x_{t+\tau}$ and transfer that knowledge to produce better samples.

In many generative approaches for spatio-temporal modeling, a noise distribution $p_\text{noise}(\epsilon) = \mathcal{N}(\epsilon;0, I_{d_\text{noise}})$ is mapped to the transition distribution $p(x_{t+\tau}|x_{t-k\tau:t::\tau})$. Different frameworks realize this mapping, either explicitly (with adversarial training for GANs or CRPS training for FGN \cite{FGN_Alet2025SkillfulJP}) or implicitly (solving the underlying probability flow ODE from noise to data manifold, for diffusion models).

We propose to combine generative losses with the following regularization objective: \textit{the noise distribution mean should map to the output distribution mean}. Conditioned on the noise distribution mean $0_{d_\text{noise}}$, the model's output should match the distribution mean $\mathbb{E}_{x_{t+\tau} \sim p}[x_{t+\tau} \mid x_{t-k\tau:t::\tau}]$,
which we supervise with a standard regression objective to the ground truth $x_{t+\tau}$:

\begin{align}
    \hat{d}^\text{mean}_{t+\tau} &= f(x_{t-k\tau:t::\tau}, 0_{d_\text{noise}}) \\
    \mathcal{L}_{\text{MPR}} &= ||\hat{d}^{\text{mean}}_{t+\tau} - d_{t+\tau}||_1
\end{align}

We implement MPR for both FGN models and diffusion models, which requires an extra deterministic pass at each training step ($\epsilon = \mathbf{0}_{d_\text{noise}}$) alongside the stochastic loss, as illustrated in Figure \ref{fig:generative_methods}.

\subsection{Generative modeling from a deterministic foundation model with FGN}

One of our primary goals is to make \emph{generative} PDE emulators that sample multiple trajectories given an initial state. Our Kastor model leverages the FGN paradigm~\cite{FGN_Alet2025SkillfulJP}, which generates samples supervised against ground truth using a strictly proper scoring rule. The scoring rule is a distance between the empirical and ground truth distributions. When training on PDE simulations, we only have a single data point $x_{t+\tau}$ for the conditional distribution $p(x_{t+\tau} | x_{t-k\tau:t::\tau})$, but the scoring rule can still be computed using this single reference. We use CRPS \cite{gneiting2007strictly}, a metric on the marginals of the predicted and reference distributions.

Samples are created in a single model forward pass by varying a source of stochasticity injected into the architecture.
Similarly to FGN, we mainly use a low-dimensional source of Gaussian noise $\mathcal{N}(0, I_{d_\text{noise}})$ with a dimension $d_\text{noise}=32$, but we also investigate other sources of stochasticity in this paper such as patch jittering and dropout.
This low-dimensional noise is first passed into a 2-layer MLP without bias, and then used as conditioning for AdaLN \cite{perez2018film} at every spatial mixing attention layer in the transformer architecture.
This design constrains the model to generate globally coherent variability~\cite{FGN_Alet2025SkillfulJP}.
With $f_\theta$ being our state difference predictor, the training loss for a single sample is:
\begin{align}
    \epsilon_1, \epsilon_2 &\sim \mathcal{N}(0, I_{d_\text{noise}}) \\
    \begin{split}
    \mathcal{L}_{\text{fCRPS}} &= \frac{1}{2} \Big(||\hat{d}^1_{t+\tau} - d_{t+\tau}||_1 + ||\hat{d}^2_{t+\tau} - d_{t+\tau}||_1 \\
    &\quad\quad\quad- ||\hat{d}^1_{t+\tau} -\hat{d}^2_{t+\tau}||_1 \Big) 
    \end{split}
    \label{eq:fcrps_train}
\end{align}
We keep the stochasticity introduced by patch jittering implicit in the notations, but we show in the experiments that it significantly contributes to performance. Finally, we initialize the AdaLN parameters to zero so that the modulation is the identity function at the beginning of training, recovering the behavior of the unconditioned model.

For FGN, we add the constraint that the null conditioning $0_{d_\text{noise}}$ makes the AdaLN modules behave as identities, which we implement by removing the biases in the noise embedding MLP and in the AdaLN layers.

\subsection{Generative baselines}

\paragraph{MAE-training with patch jittering.} 
The Walrus model is trained with MAE with \textit{patch jittering} as a form of data augmentation. We show that patch jittering can be used as a source of stochasticity to generate an ensemble of forecasts, which we use as default ensembling method when evaluating our other proposed contributions. These models will be refered to as "MAE" or "MAE-trained" in the paper.

\paragraph{Diffusion modeling baseline.}
To establish a robust generative baseline, we adapt the Walrus architecture into a diffusion model.
This involves turning it into a conditional denoiser, providing the context states as well as the noised next state as inputs that the model is trained to denoise.
We implement this using a second set of channel embeddings for the noisy next state.
We also adapt our proposed MPR in this context: to compute the regularization objective, the model only sees the context states without any noised inputs (\textit{i.e.}, only the original set of channels), and the noise level is set to zero.
The complete description can be found in appendix.

\subsection{Gradient Difference Loss}
\label{subsec:methods_gdl}

While MAE is often used as a training loss for operator learning, it can fail to capture some quantities that are important for a dynamical system, like the spatial derivative of the target fields.
We address this by implementing the Gradient Difference Loss (GDL) from \cite{mathieu2015deep}: the GDL between the forecasting state $\hat{d}_{t+\tau}$ and the ground truth $d_{t+\tau}$ is
\begin{equation}
\mathcal{L}_\text{GDL}(\hat{d}_{t+\tau}, d_{t+\tau}) = \mathcal{L}(\partial\hat{d}_{t+\tau},  \partial d_{t+\tau})\,,
\end{equation}
where $\partial$ is the derivative w.r.t the spatial coordinate.
The choice of the base loss function $\mathcal{L}$ and its corresponding inputs vary depending on the primary training objective:
\begin{itemize}
    \item MAE training: We take prediction $\hat{d}_{t+\tau}$ and $\mathcal{L} = \text{MAE}$.
    \item FGN w/o MPR: We use $(\hat{d}_{t+\tau}^1, \hat{d}_{t+\tau}^2)$, and $\mathcal{L} = \text{CRPS}$.
    \item FGN with MPR: We use $\hat{d}^\text{mean}_{t+\tau}$ and  $\mathcal{L} = \text{MAE}$.
\end{itemize}


\subsection{Evaluation}

We evaluate our model on the 10 2D datasets from \textit{The Well}, which cover simulations from various physical domains, and represent various challenges (e.g. capturing stochasticity, shocks, sharp edges).
All simulations for a given dataset have the same spatial and temporal resolutions, which vary across datasets.

The evaluation protocol is as follows: given a set of context states $(x_{t-k:t})$, we make a set of $M$ trajectory forecasts with our model $(\hat{x}^i_{t+1:t+T})_{1 \leq i \leq M}$ up to a final lead time $T$, with auto-regressive rollouts. We then compare these outputs to the ground truth trajectory $x_{t+1:t+T}$. For all our evaluation we use $M=8, T=32$, and choose two initial context per dataset trajectory: one at the beginning, and one at the middle to evaluate models on a non-transient regime. Further details are provided in appendix.
We evaluate trajectories separately for each lead time $1\leq t\leq T$, with the metrics below.
We report metrics on the validation sets for the ablations.
To reduce computational costs, we perform most ablations using strided models for datasets where they outperform the autoregressive baseline.
For consistency across datasets and experiments, we always report aggregated metrics on trajectories sub-sampled at $\tau=4$.

\subsection{Metrics}

The formal definition of all metrics is in appendix.
Following standard practice, we compute RMSE between the ground truth and the empirical mean over our $M$ forecast members, which we refer to as \textbf{EnsembleMeanRMSE}.

To evaluate the quality of predicted distributions, we use the fair Continuous Ranked Probability Score metric (\textbf{fCRPS}) \cite{ferro2014fair}, minimized when the marginals of the empirical and ground truth distributions match.

We report the Spread-skill ratio \textbf{SpSkR}, which is the ratio of the ensemble variance and the mean squared error of the ensemble mean, as a measure of calibration.
For a perfectly calibrated forecast, this ratio should equal 1.
Finally, we evaluate spectral error with the Log Spectral Distance (\textbf{LSD}), which is the error of the radially-averaged power spectrum density (\textbf{RAPSD}) compared to the ground truth.

\textbf{Skill Score.} The \textit{Skill Score} is defined for metrics $\mathcal{M}$ that should be minimized (e.g. \textit{EnsMeanRMSE} and \textit{fCRPS}) as
\begin{equation}\label{eq:ss}
\mathcal{M}_{ss}(\text{model}) = 1 - \frac{1}{|\mathcal{V}|} \sum_v \frac{\mathcal{M}_v(\text{model})}{\mathcal{M}_v(\text{reference})}
\end{equation}
where $\mathcal{M}_v(\text{model})$ is the metric value for $v$, one of the $|\mathcal{V}|$ physical variables of the dataset.

The \textit{skill score} reflects the relative gain of metric $\mathcal{M}$ compared to a reference model across all physical variables: A skill score of $0.1$ means that the metric value is on average 10\% lower than the reference model. Unlike VRMSE \cite{ohana2024well}, skill scores have a comparable range of values across physical variables, datasets and lead times, and can be compared more easily. In this paper, we use as reference a model retrained with the fine-tuning methodology of Walrus \cite{mccabe2025walruscrossdomainfoundationmodel}. Our retrained models perform comparably to Walrus's reported numbers, see details in appendix, validating our reference finetuning methodology.

\subsection{Implementation details}

\paragraph{Fine-tuning.} Unlike the Walrus pretrained model that was trained with sample-dependent statistics, we normalize states by the global dataset mean and standard deviation, and state differences by their standard deviation. Similarly to Walrus, we fine-tune models for 62,500 steps with batch size 8. We use the Muon optimizer with a learning rate of 1e-3 and a linear warmup with cosine decay learning rate scheduler. When using GDL, we use $\lambda_G=2000$ (since gradient differences are very small).

\paragraph{Adapting the model for different compute budgets.} We train models with 3 different compute budgets: Small model (S) with $(N_{in}, N_\text{layers})=(2, 20)$, Medium (M) with $(N_{in}, N_\text{layers})=(4, 20)$, and Large (L) $(N_{in}, N_\text{layers})=(4, 40)$. Medium and Large models have respectively 2x and 4x the inference cost of the Small model. To reduce the computational cost of our emulators, we deliberately choose operating points less costly than the default regime for the pretrained Walrus model $(N_{in}, N_{layers}) = (6, 40)$. For up-sampling with $\tau=4$, we use a model with $(N_{in}, N_\text{layers})=(3, 20)$, so that the total cost for forecasting four states is less than the stride 1 baseline.

\paragraph{Walrus pretrained weights loading strategy.} When loading layers from the small (S) and medium (M) models that contain only half of the layers of the pretrained Walrus model, our strategy is to load weights from even layers only. Our up-sampling model (20 layers) is also initialized from the pretrained weights, and fine-tuned as a non-causal model.

\section{Results}

\subsection{Inference Strategy}
In this section, we analyze our strategy to make efficient forecasts described in Figure \ref{fig:inference_strat_figure}.
We use the MAE models which we evaluate with the \textit{Ensemble Mean RMSE} skill score relative to the reference model.

\begin{figure}[ht]
    \centering
    \includegraphics[width=\linewidth]{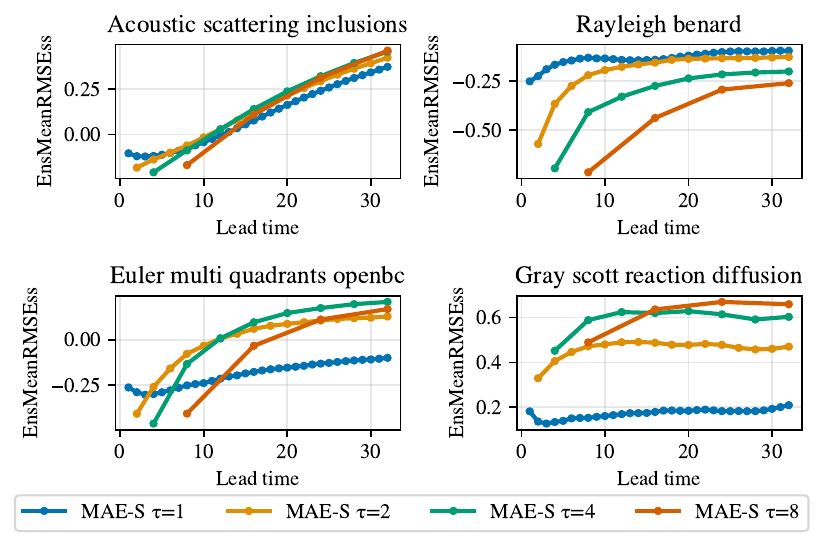}
    \caption{Ensemble Mean RMSE skill score $\uparrow$ per lead time, for four representative datasets and MAE-trained models (size S) at strides $\tau \in \{1, 2, 4, 8\}$. While stride $T=1$ often performs best at early lead times, on a majority of datasets using larger strides is beneficial for longer horizons.}
    \label{fig:time_stride_ablation_repr}
\end{figure}

\begin{figure}[ht]
    \centering
    \includegraphics[width=\columnwidth]{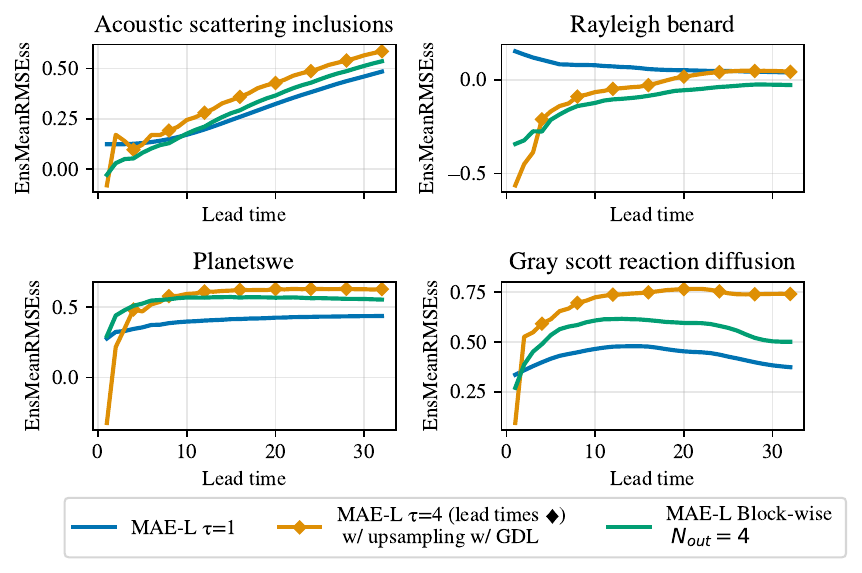}
    \caption{Ensemble Mean RMSE skill score per lead time for our MAE-trained model (L, $N_{in}=4$) with temporal upsampling, compared to the block-wise and the standard auto-regressive methods (see  Figure~\ref{fig:generative_methods}). Temporal upsampling allows to effectively recover dense trajectories, and compares favorably to block-wise regression for $T > 8$ across datasets.} 
    \label{fig:inference_strat_results}
\end{figure}

We first analyze the impact of time stride $\tau$ on forecasting accuracy.
A higher value of $\tau$ can reduce error accumulation, but increases the complexity of the function to approximate since it requires capturing state changes over longer temporal jumps, which can lead to higher errors given a fixed model capacity.
We find that the best strategy depends on the data the model is trained on, and present a set of representative scenarios in Figure \ref{fig:time_stride_ablation_repr}. On our benchmark datasets, using a higher time stride of $\tau=4$ is beneficial for 8 out of 10 domains.
For some datasets like \textit{Gray-Scott Reaction Diffusion} and \textit{Active Matter}, the ranking of methods does not depend on lead time, with one time stride being consistently better across lead times.
For most datasets, like \textit{Acoustic Scattering Inclusions} and \textit{Euler Multi-quadrants}, the optimal stride depends on the lead time: at small lead times, it is better to leverage the $\tau=1$ model with a few rollout steps; at longer lead times, the models with larger time stride benefit from less error accumulation and outperform the other methods.
Qualitatively, we find that the evolution of sharp shock fronts is often better predicted using longer time steps, which we illustrate in Figure~\ref{fig:qual_dt1_vs_dt4} and in appendix. on the \textit{Euler Multi-quadrants} and \textit{Gray-Scott Reaction Diffusion} datasets.

\begin{figure}[ht]
    \centering
    \includegraphics[width=0.95\linewidth]
    {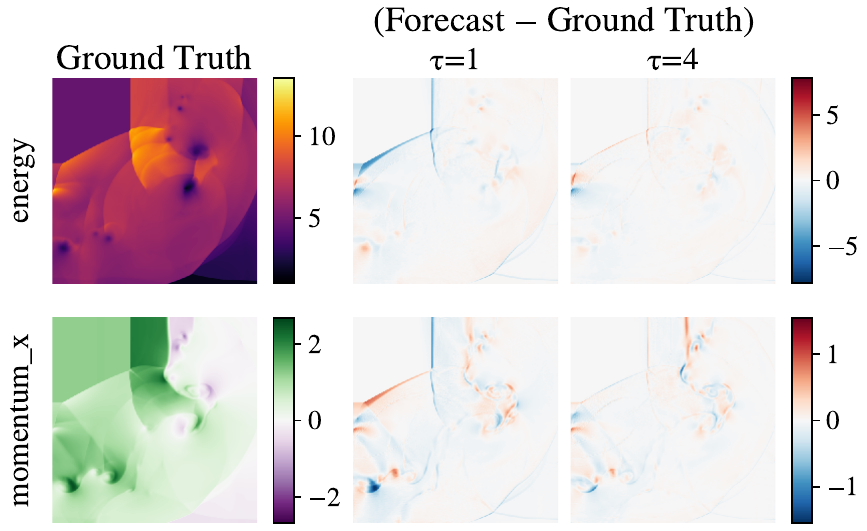}
    \caption{Qualitative comparison between different time strides on the \textit{Euler Multi-quadrants OpenBC} dataset. At lead time 32, errors around the shock fronts are noticeably reduced for the longer stride $\tau=4$.}
    \label{fig:qual_dt1_vs_dt4}
\end{figure}

The complete ablation for all datasets is presented in appendix. 
Based on these findings, we find that a good general strategy is to use $\tau=4$, with the exception of two datasets, \textit{Active Matter} and \textit{Rayleigh-Bénard}, where using a time stride of 1 is better.
We suspect this is related to highly sub-sampled physical simulations.
For \textit{Active Matter}, only one frame every 625 simulation steps was saved; for \textit{Rayleigh-Bénard}, authors mention that "High Reynolds simulations are very time-consuming because they require very small time-steps to prevent the solution from diverging" \cite{ohana2024well}.
We therefore chose to keep autoregressive rollouts with $\tau=1$ for those two datasets. Some datasets would have benefited from a higher time stride (e.g. $\tau=8$), but we retain $\tau=4$ as it strikes a good balance across lead times and datasets.

\paragraph{Strategies for predicting the full trajectories.}

We now evaluate our strategy to forecast missing states when a time stride $\tau>1$ is used. Figure \ref{fig:inference_strat_results} shows that our MAE-trained models with $\tau=4$ combined with the small up-sampling model obtains the best performance, except on \textit{Rayleigh-Bénard}. We observe experimentally that supervising the super-resolution model with GDL is important to generate high quality intermediate frames.
A block-wise auto-regressive model with 4 output steps (Figure \ref{fig:inference_strat_figure} right) has a similar computational budget as our model and makes the same number of rollout steps. However, the up-sampling approach allows to best forecast some specific states instead of splitting the model capacity across 4 states. Furthermore, our strided inference strategy can leverage the causal structure when training the low-resolution $\tau=4$ model, which is beneficial as we show in appendix.

\subsection{Impact of Mean Prediction Regularization}

For further analysis, we evaluate our size S generative models based on FGN and diffusion models on four representative datasets of \textit{The Well}: \textit{PlanetSWE, Acoustic Scattering Maze, Active Matter and Rayleigh-Bénard}.
We use a time stride of 4 for the two former, and 1 for the two latter, which we found optimal in the previous section.

\begin{figure}[ht]
    \centering
    \includegraphics[width=\columnwidth]{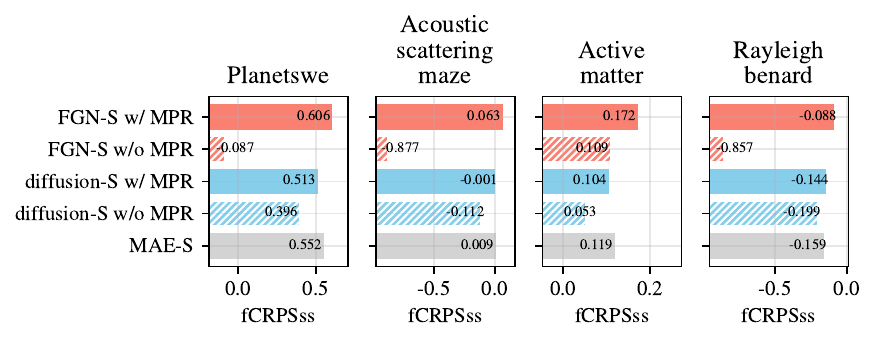}
    \caption{Performance of S models trained with and without MPR, for both our FGN and diffusion-based models. MPR is helpful for both methods, and critical for reaching best performance with FGN.}
    \label{fig:mean_pred_reg}
\end{figure}

\begin{figure}[htbp]
    \centering
    \includegraphics[width=1\linewidth] {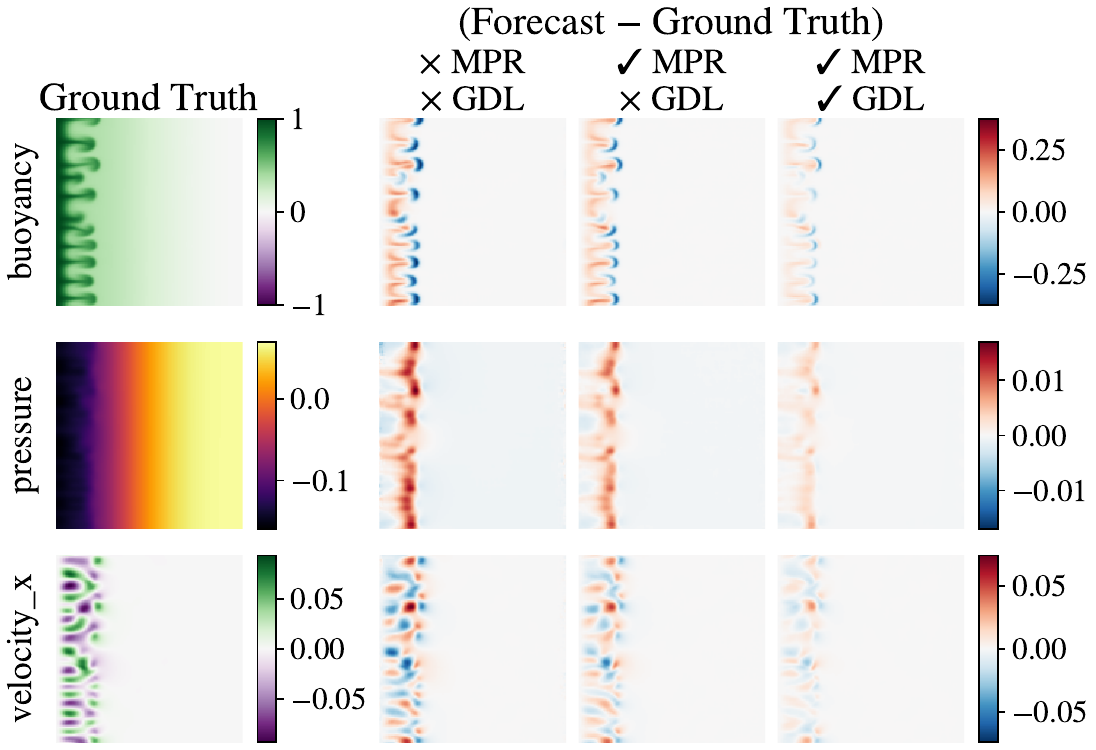}
    \caption{Comparison of FGN models trained with different regularizations. On center crops from the \textit{Rayleigh-Bénard} dataset at lead time $t=16$, we observe reduced errors from MPR across channels, and further improvements from GDL.}
    \label{fig:qual_gdl_mpr_comparison_main}
\end{figure}

We analyze in Figure~\ref{fig:mean_pred_reg} the impact of Mean Prediction Regularization (MPR) for our FGN model.
Without regularization, FGN has poor performance and performs worse than diffusion or even the MAE-trained baseline, which we also observe qualitatively, as shown in Figure~\ref{fig:qual_gdl_mpr_comparison_main} and illustrated on additional datasets in appendix.
MPR is greatly beneficial for all datasets and lead times and sets regularized FGN as the best performing model, outperforming the MAE-trained baseline by 15\% on average. The diffusion baseline is better than FGN, and MPR improves upon it, but still underperforms FGN with MPR. Diffusion models were evaluated using 5 diffusion steps, which we found to be optimal experimentally (increasing this number tends to degrade the CRPS).

\subsection{Impact of Gradient Difference Loss}

\begin{figure}
    \centering
    \includegraphics[width=\columnwidth]{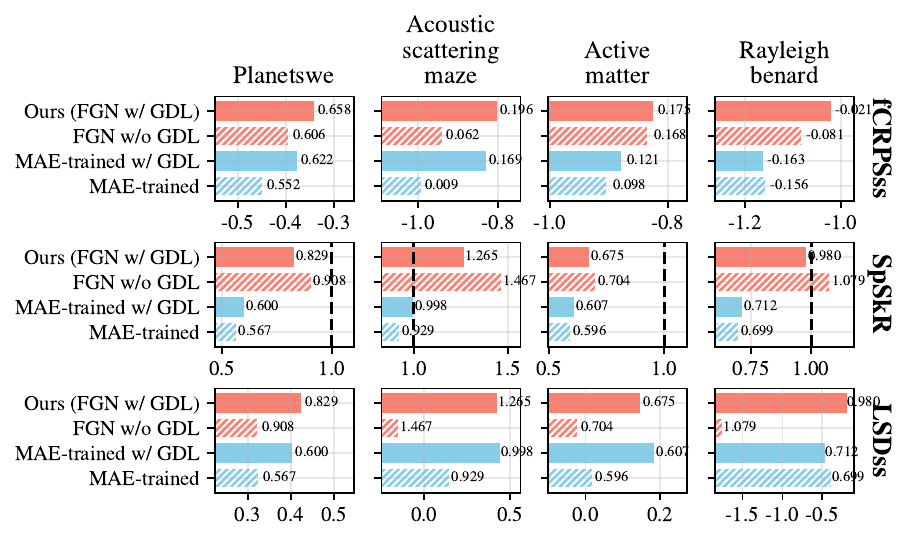}
    \caption{Comparison of FGN and MAE-trained models (size S) with and without GDL. We find that FGN improves the spread-skill ratio and GDL significantly improves both fCRPS and LSD skill scores, when averaged across lead times for a stride of $\tau=4$.}
    \label{fig:gradient_difference_loss}
\end{figure}

We now evaluate the impact of the Gradient Difference Loss (GDL) in Figure \ref{fig:gradient_difference_loss}, where the MAE-trained model (size S) is shown for reference. We find that the loss brings significant improvements for 3 out of four datasets, and marginal improvements for \textit{Active Matter}. The best improvements occur for \textit{Acoustic Scattering Maze} (+13.3\% fCRPS skill score), where sharp maze walls create discontinuities well-suited for gradient matching. We can see that the FGN method without GDL improves forecast dispersion compared to MAE on three datasets (spread-skill score closer to 1) but worsens the spectral error measured with LSD (lower skill score) compared to MAE training. Using GDL allows to compensate and improves spectral consistency. A breakdown of score improvement per leadtime is in appendix. 
\ifarxiv
(Figure \ref{fig:gdl_summary_per_leadtime}).
\fi

\section{Limitations}

Despite the improved speed, accuracy, and power spectrum density of our model, there is still room for improvements. In particular, we find that the calibration of our models can still be improved, as reflected by the Spread-Skill ratios presented in Figure~\ref{fig:gradient_difference_loss}. Qualitatively, we observed that, although to a lesser extent than for the baseline, trajectories can still accumulate artifacts in some cases and diverge to unphysical behavior.
Finally, a standalone emulator should ideally start rollouts from a \textit{single} state; and then progressively increase the context length. Following existing practices in the field, we rely on a fixed-length context window to start rollouts, implicitly relying on a physics solver to provide the initial context.

\section{Conclusion}

In this work, we presented a robust framework for training and deploying generative surrogate models for PDE simulations, leveraging the representation power of the Walrus pre-trained physics foundation models.
We showed that converting a deterministic foundation model into a generative one yields superior predictive capabilities, provided the training is appropriately constrained. To this end, our novel Mean Prediction Regularization proved highly effective, enabling FGN-based architectures to substantially outperform both standard deterministic baselines and more computationally heavy diffusion models. Combined with spatial gradient matching, our approach successfully preserves the spectral distribution of physical variables.
Finally, by shifting from standard single-step auto-regressive rollouts to a two-stage inference strategy combining advancements at longer time strides with non-causal temporal super-resolution, we demonstrate gains both in accuracy and computational efficiency.

\section{Acknowledgements}

We would like to thank Clément Crepy and Klaus Greff for various contributions related to this work. Many thanks to Patrick Perez, Valentin De Bortoli and Michael McCabe for proofreading the manuscript.

\bibliography{bib}

\appendix
\clearpage
\section{Qualitative Results}
\label{sec:qual_results}

We investigate qualitatively the effect of several method choices, in particular the effect of training with a larger model, comparing Kastor-L and Kastor-M in Section~\ref{subsec:qual_model_size}, differences arising from varying the time stride, comparing $\tau=1$ and $\tau=4$ in Section~\ref{subsec:qual_time_stride}, and the impact of the added MPR and GDL regularizations in Section~\ref{subsec:qual_impact_gdl_mpr}.

\subsection{Impact of model size}
\label{subsec:qual_model_size}

\begin{figure*}[htbp]
    \centering
    \begin{subfigure}{\textwidth}
        \centering
        \includegraphics[width=\linewidth]{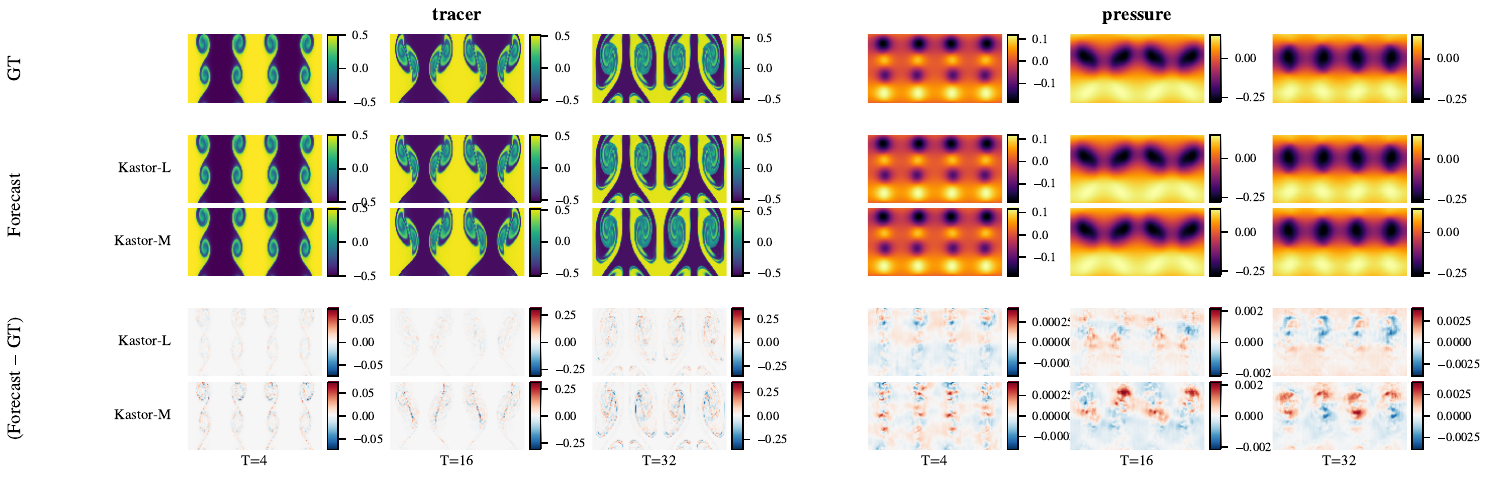}
        \caption{Sample 1}
        \label{fig:shear_flow_m_vs_l_sample1}
    \end{subfigure}
    \begin{subfigure}{\textwidth}
        \centering
        \includegraphics[width=\linewidth]{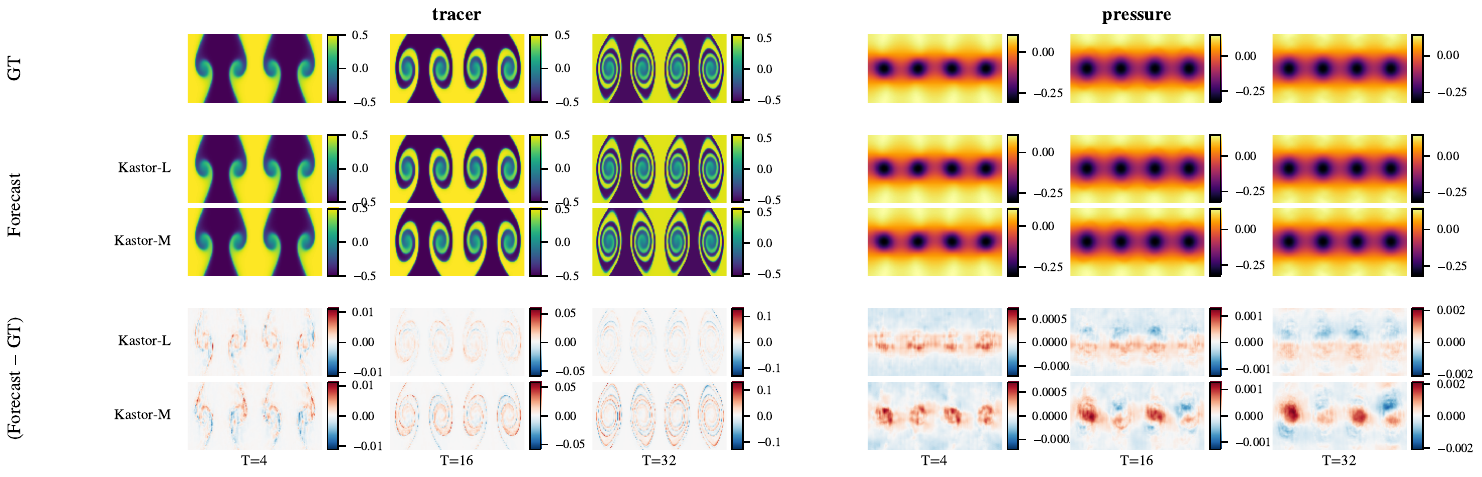}
        \caption{Sample 2}
        \label{fig:shear_flow_m_vs_l_sample2}
    \end{subfigure}
     \begin{subfigure}{\textwidth}
         \centering
         \includegraphics[width=\linewidth]{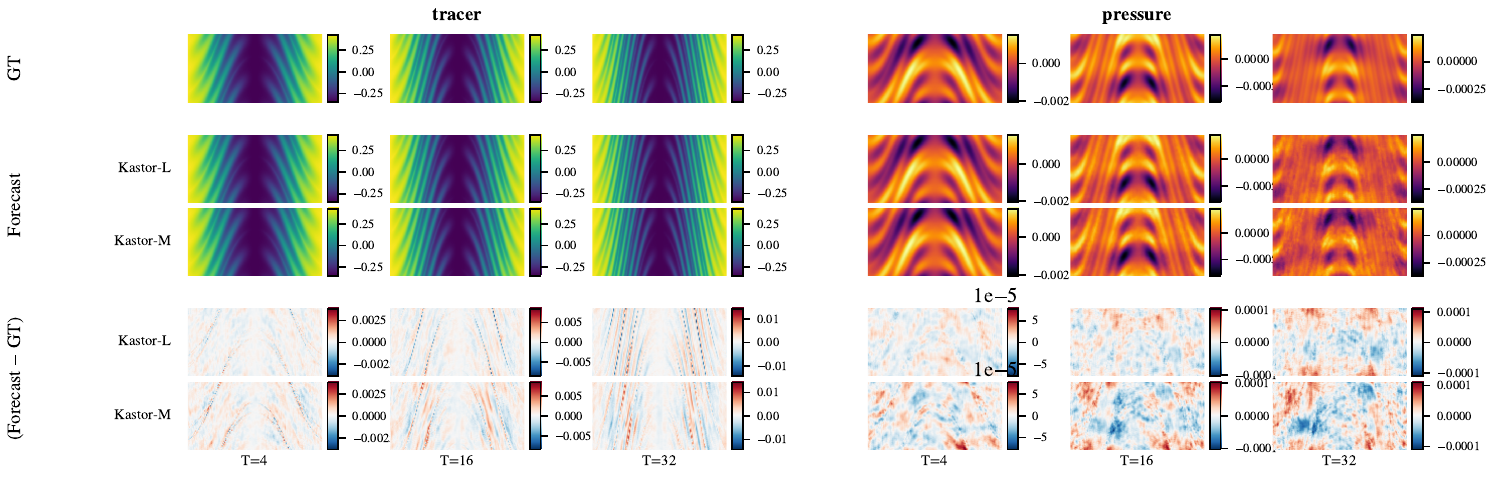}
         \caption{Sample 3}
         \label{fig:shear_flow_m_vs_l_sample3}
     \end{subfigure}
    \caption{Qualitative comparison of Kastor-M (20 input layers) and Kastor-L (40 input layers) on samples from the Periodic Shear Flow dataset on the tracer and pressure fields. Consistently, we observe that Kastor-L outperforms the smaller model.}
    \label{fig:app_qual_scaling_shear_flow}
\end{figure*}

\begin{figure*}[htbp]
    \centering
    \includegraphics[width=0.95\textwidth]{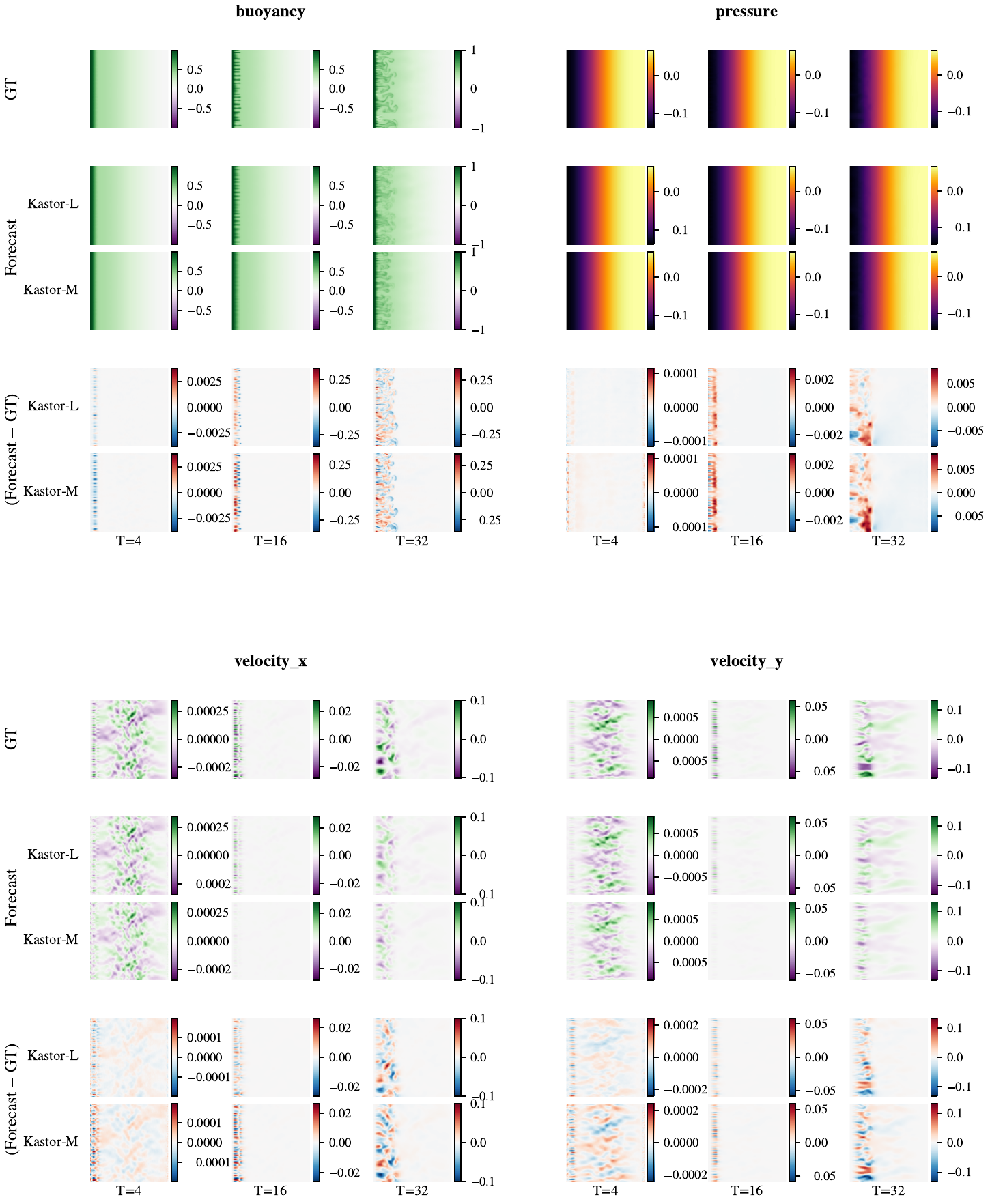}
    \caption{Qualitative comparison of Kastor-M (20 input layers) and Kastor-L (40 input layers) on the center crop of a validation sample from the \textit{Rayleigh-Bénard} dataset. At both early ($T=4$) and late ($T=32$) lead times, across channels, we observe in the (Forecast - GT) rows that model size reduction results in larger errors.}
    \label{fig:app_qual_scaling_rayleigh_benard}
\end{figure*}

\begin{figure*}[htbp]
    \centering
    \begin{subfigure}{\textwidth}
        \centering
        \includegraphics[width=\linewidth]{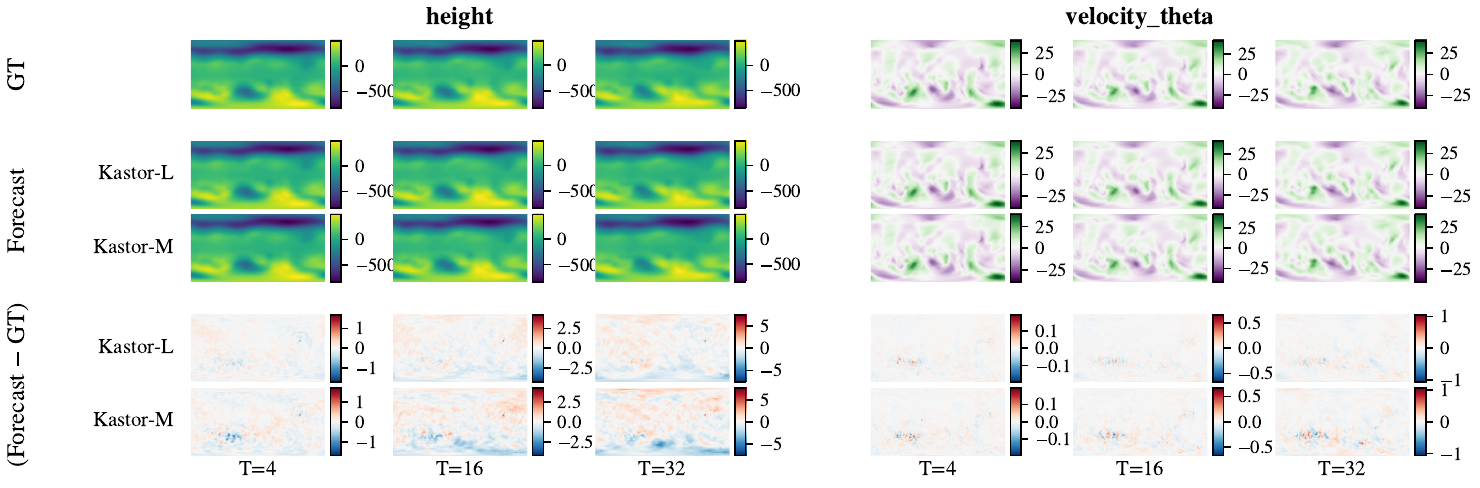}
        \caption{Sample 1}
        \label{fig:planetswe_m_vs_l_sample1}
    \end{subfigure}
    \begin{subfigure}{\textwidth}
        \centering
        \includegraphics[width=\linewidth]{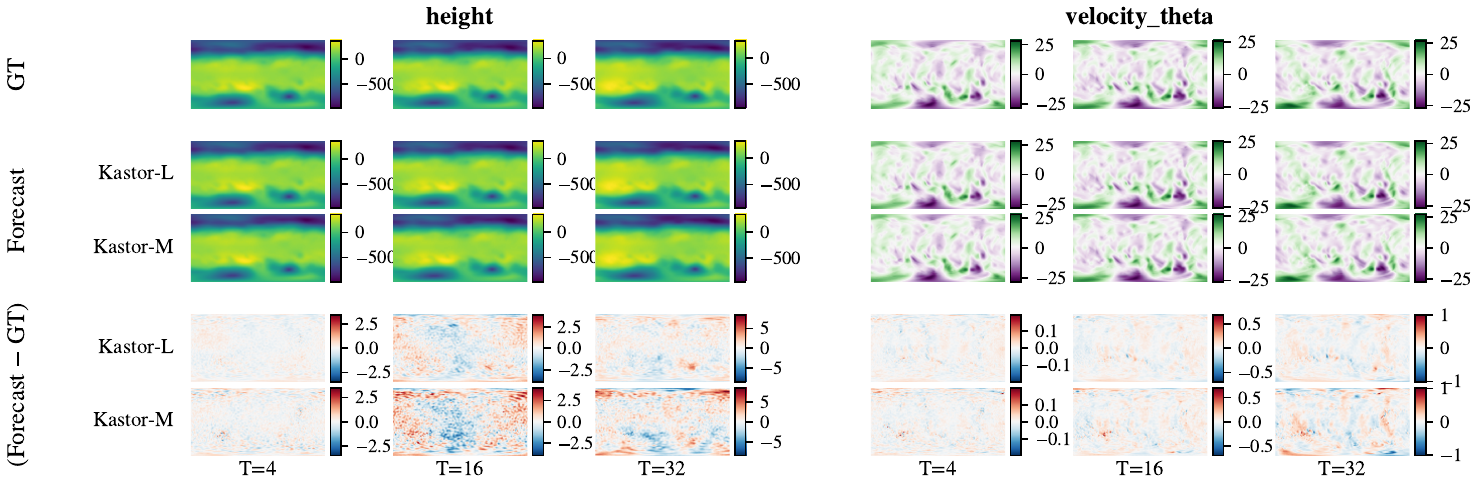}
        \caption{Sample 2}
        \label{fig:planetswe_m_vs_l_sample2}
    \end{subfigure}
    \begin{subfigure}{\textwidth}
        \centering
        \includegraphics[width=\linewidth]{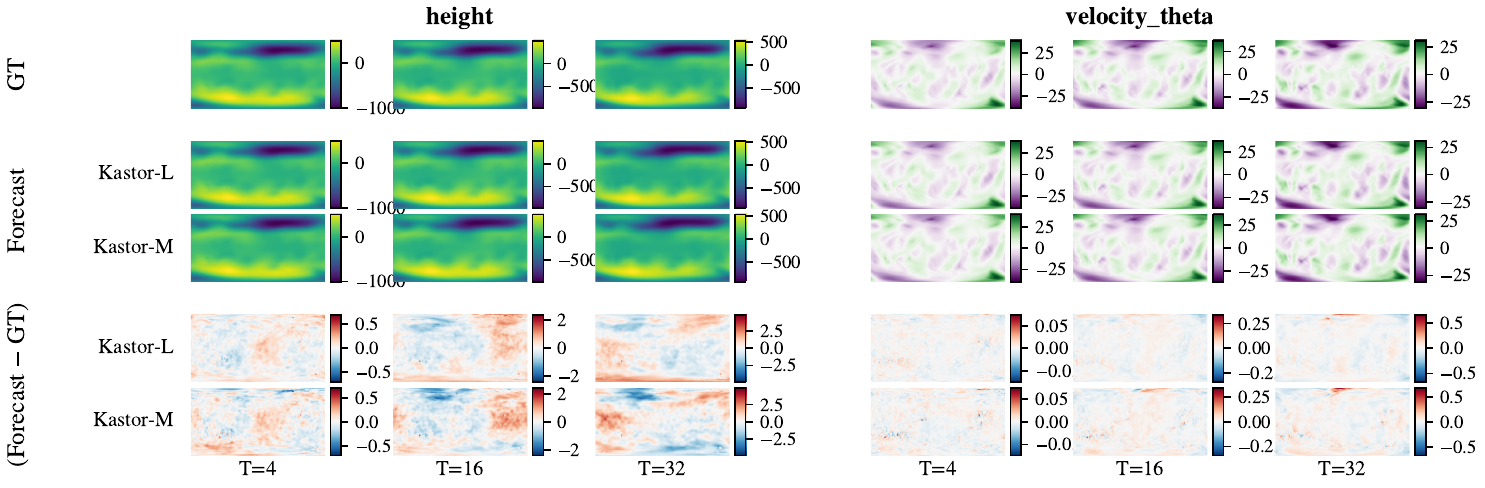}
        \caption{Sample 3}
        \label{fig:planetswe_m_vs_l_sample3}
    \end{subfigure}
    \caption{Qualitative comparison of Kastor-M (20 input layers) and Kastor-L (40 input layers) on samples from the \textit{PlanetSWE} dataset. At both early ($T=4$), intermediate ($T=16$) and later ($T=32$) lead times, we observe on two representative fields that Kastor-L qualitatively matches or outperforms the smaller model.}
    \label{fig:app_qual_scaling_planetswe}
\end{figure*}

As we quantitatively observe in Figure~\ref{fig:scaling_experiment}, we illustrate the decrease in performance for the smaller M-size model (20 layers, 4 input frames) compared to the L model (40 layers, 4 input frames).
Consistently across datasets, physical fields, and lead times, we observe larger forecasting errors for the M model compared to Kastor-L. Figure~\ref{fig:app_qual_scaling_shear_flow} shows how forecasting degrades systematically on the \textit{Periodic Shear Flow} dataset, which we also observe on \textit{PlanetSWE} in Figure~\ref{fig:app_qual_scaling_planetswe} and the \textit{Rayleigh-Bénard} (see Figure~\ref{fig:app_qual_scaling_rayleigh_benard}).

\subsection{Impact of time stride $\tau$}
\label{subsec:qual_time_stride}

\begin{figure*}[htbp]
    \centering
    \includegraphics[width=0.6\linewidth]{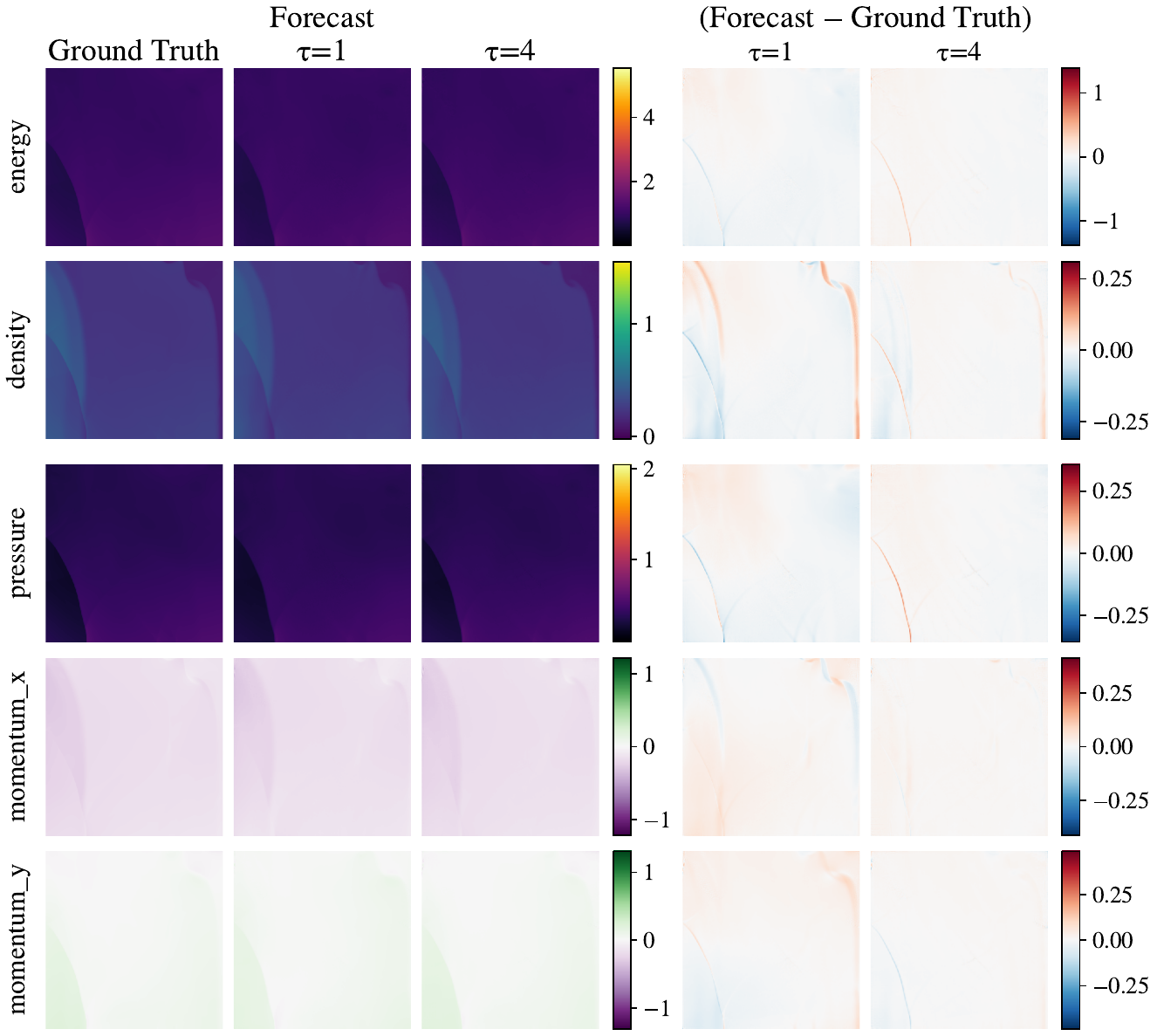}
    \vspace{0.25cm} 
    
    
    \includegraphics[width=0.6\linewidth]{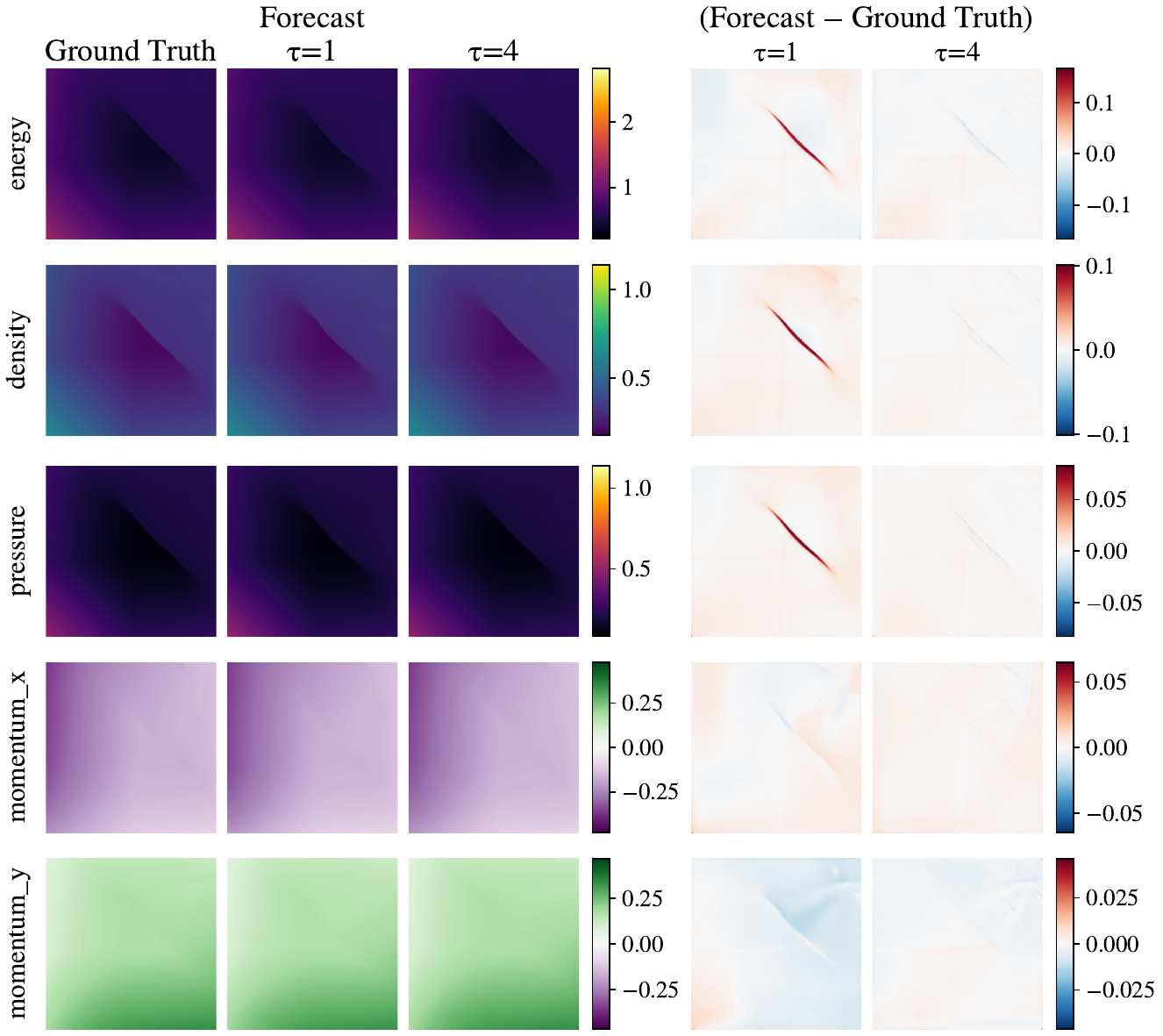}
    \vspace{0.25cm}

    \caption{Additional qualitative examples comparing different time strides on validation samples from the Euler Multi-quadrants OpenBC dataset. Consistently with the observation in the main paper, errors around the shock fronts at lead time $T=32$ are noticeably reduced for the longer stride $\tau=4$.}
    \label{fig:qual_dt1_vs_dt4_openBC}
\end{figure*}

\begin{figure*}[htbp]
    \begin{subfigure}{\textwidth}
        \centering
        \includegraphics[width=\linewidth]{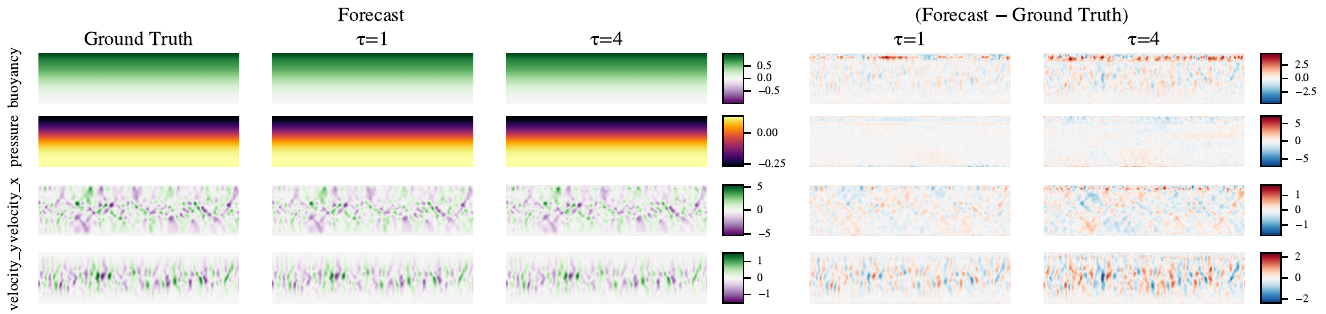}
        \caption{Sample 1, Lead time $T=4$}
    \end{subfigure}
    \begin{subfigure}{\textwidth}
        \centering
        \includegraphics[width=\linewidth]{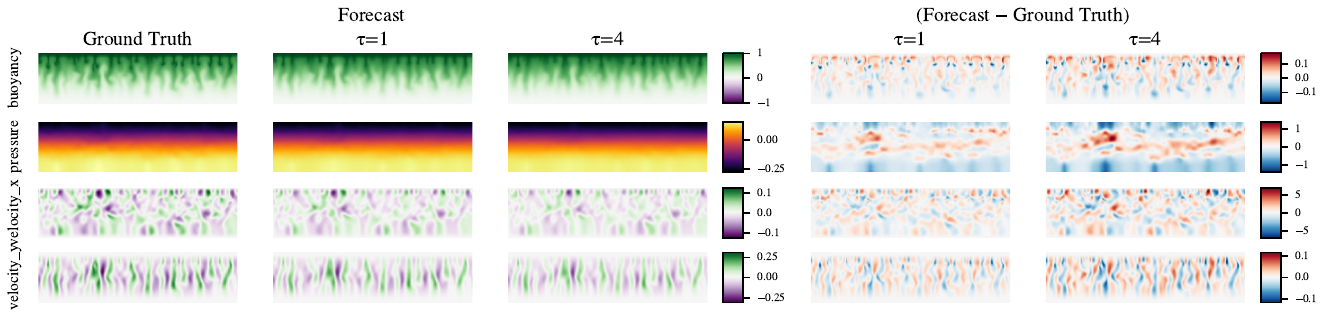}
        \caption{Sample 1, Lead time $T=32$}
    \end{subfigure}
    \begin{subfigure}{\textwidth}
        \centering
        \includegraphics[width=\linewidth]{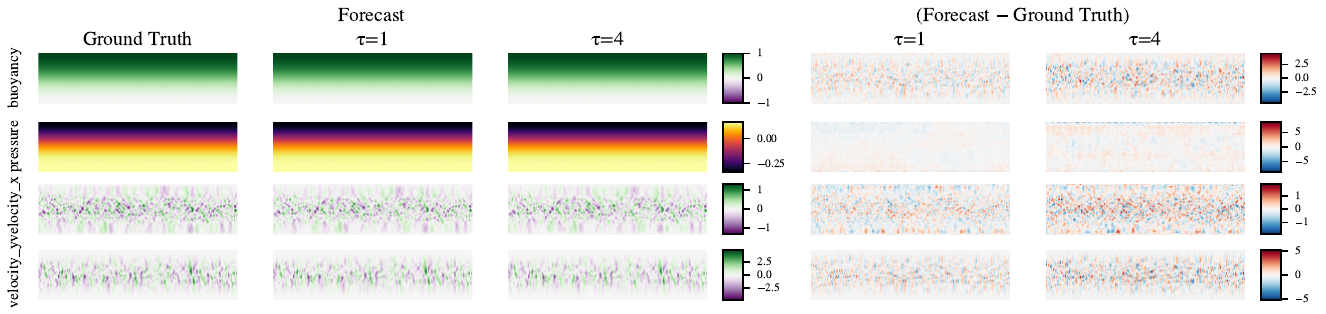}
        \caption{Sample 2, Lead time $T=4$}
    \end{subfigure}
    \begin{subfigure}{\textwidth}
        \centering
        \includegraphics[width=\linewidth]{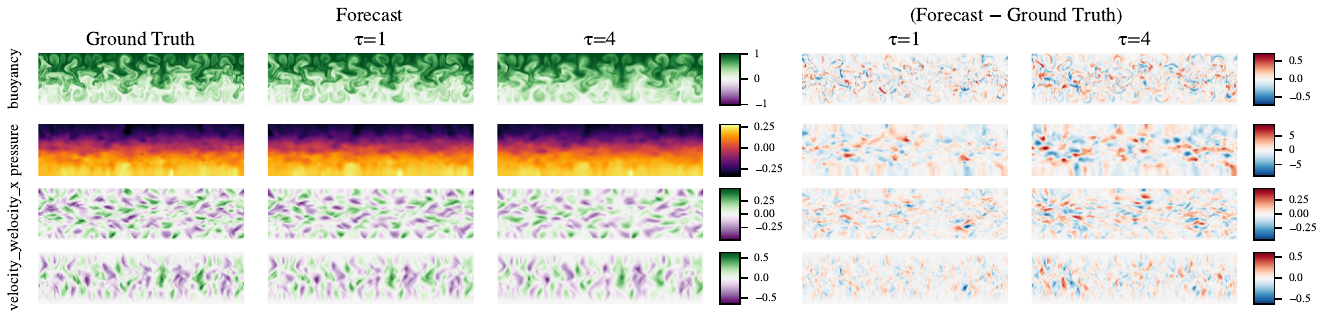}
        \caption{Sample 2, Lead time $T=32$}
    \end{subfigure}
    
    \caption{Qualitative examples comparing different time strides on two validation samples from the  Rayleigh-Bénard dataset. For models trained with a stride $\tau=4$, visibly increased errors are observed as early as lead time $T=4$ and persist at $T=32$.}
    \label{fig:qual_dt1_vs_dt4_rb}
\end{figure*}

\begin{figure*}[htbp]
    \centering
    \begin{subfigure}{\textwidth}
        \centering
        \includegraphics[width=\linewidth]{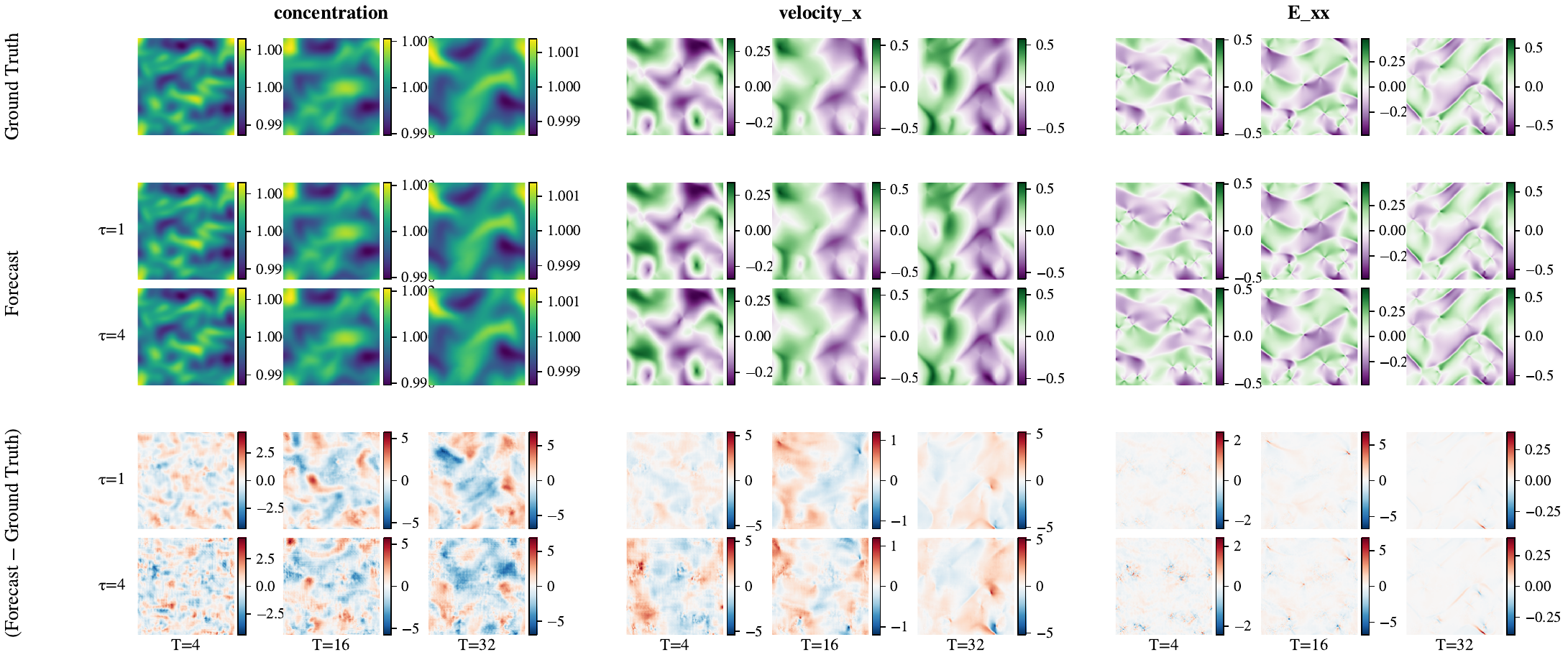}
        \caption{Sample 1}
    \end{subfigure}
    \begin{subfigure}{\textwidth}
        \centering
        \includegraphics[width=\linewidth]{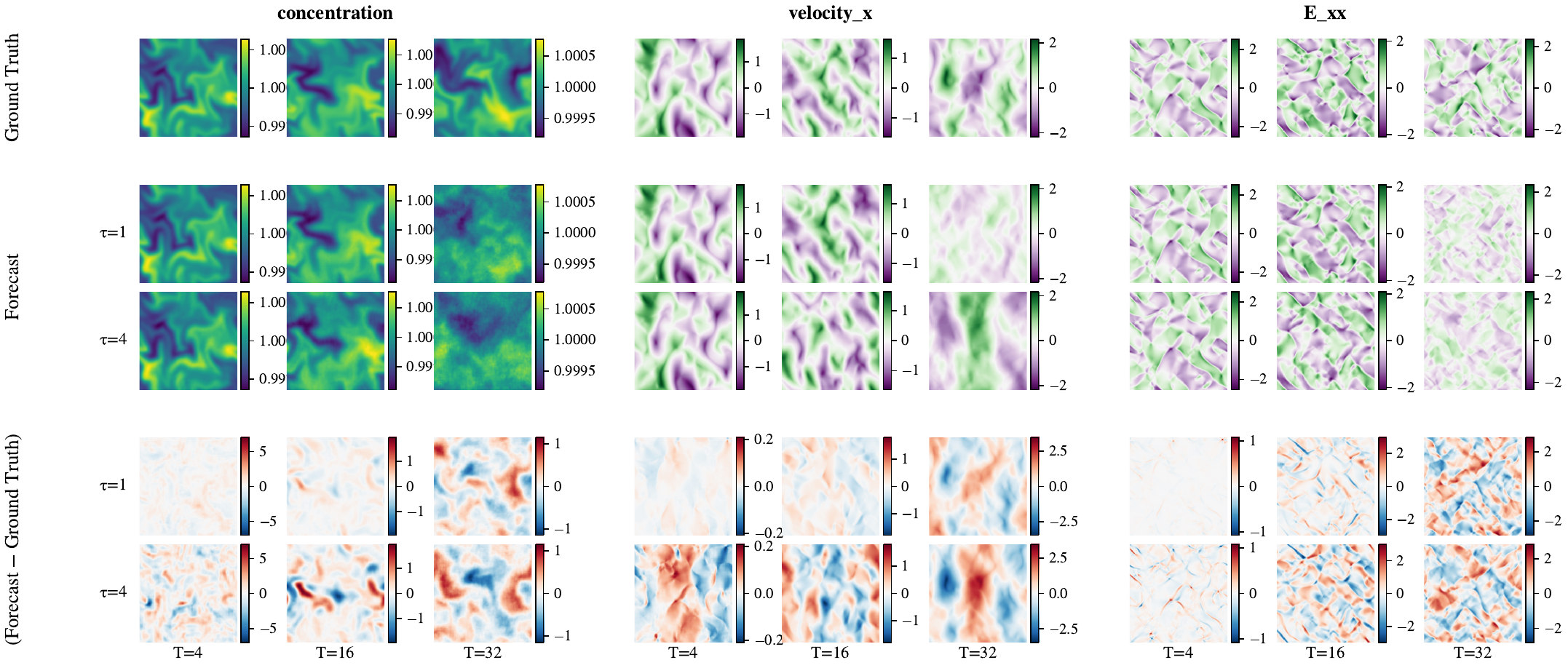}
        \caption{Sample 2}
    \end{subfigure}
    
    \caption{Qualitative examples comparing different time strides on two validation samples from the  \textit{Active Matter} dataset. Compared to stride $\tau=4$, we observe larger forecasting errors and more visible artifacts across lead times $T=4,16,32$.}
    \label{fig:qual_dt1_vs_dt4_am}
\end{figure*}

\begin{figure*}[htbp]
    \centering
    \includegraphics[width=0.65\textwidth]{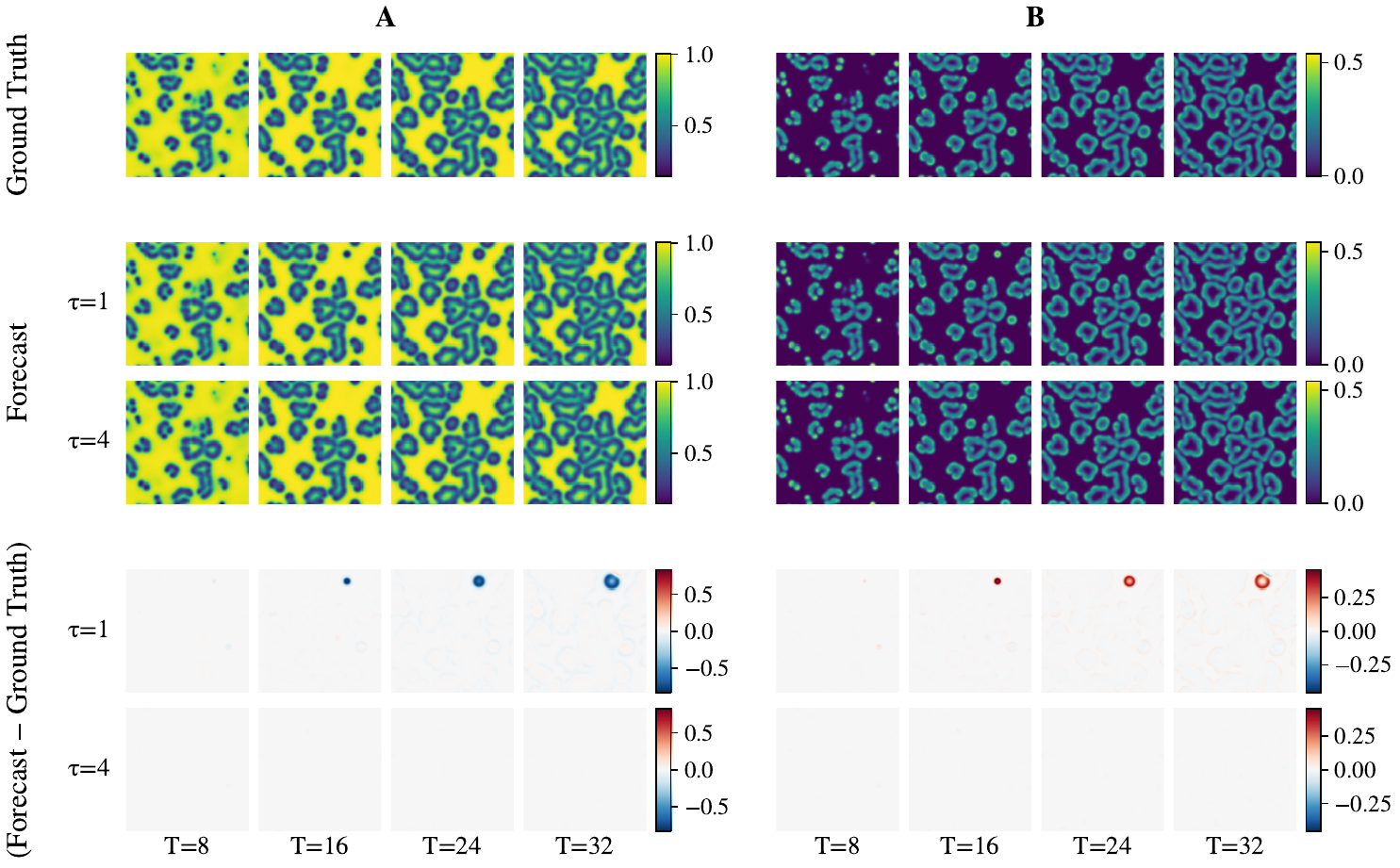}
    
    \includegraphics[width=0.65\textwidth]{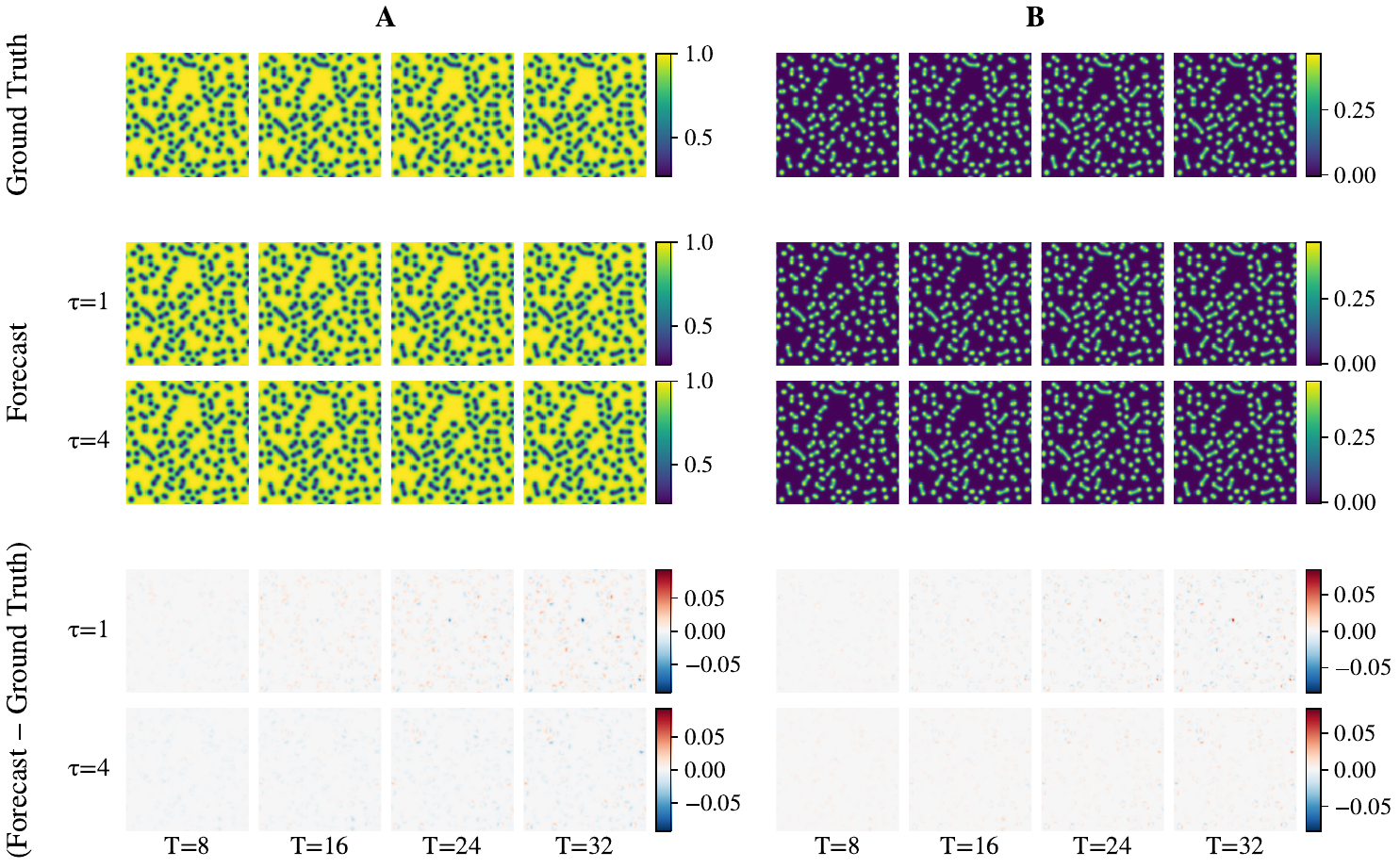}
    
    \includegraphics[width=0.65\textwidth]{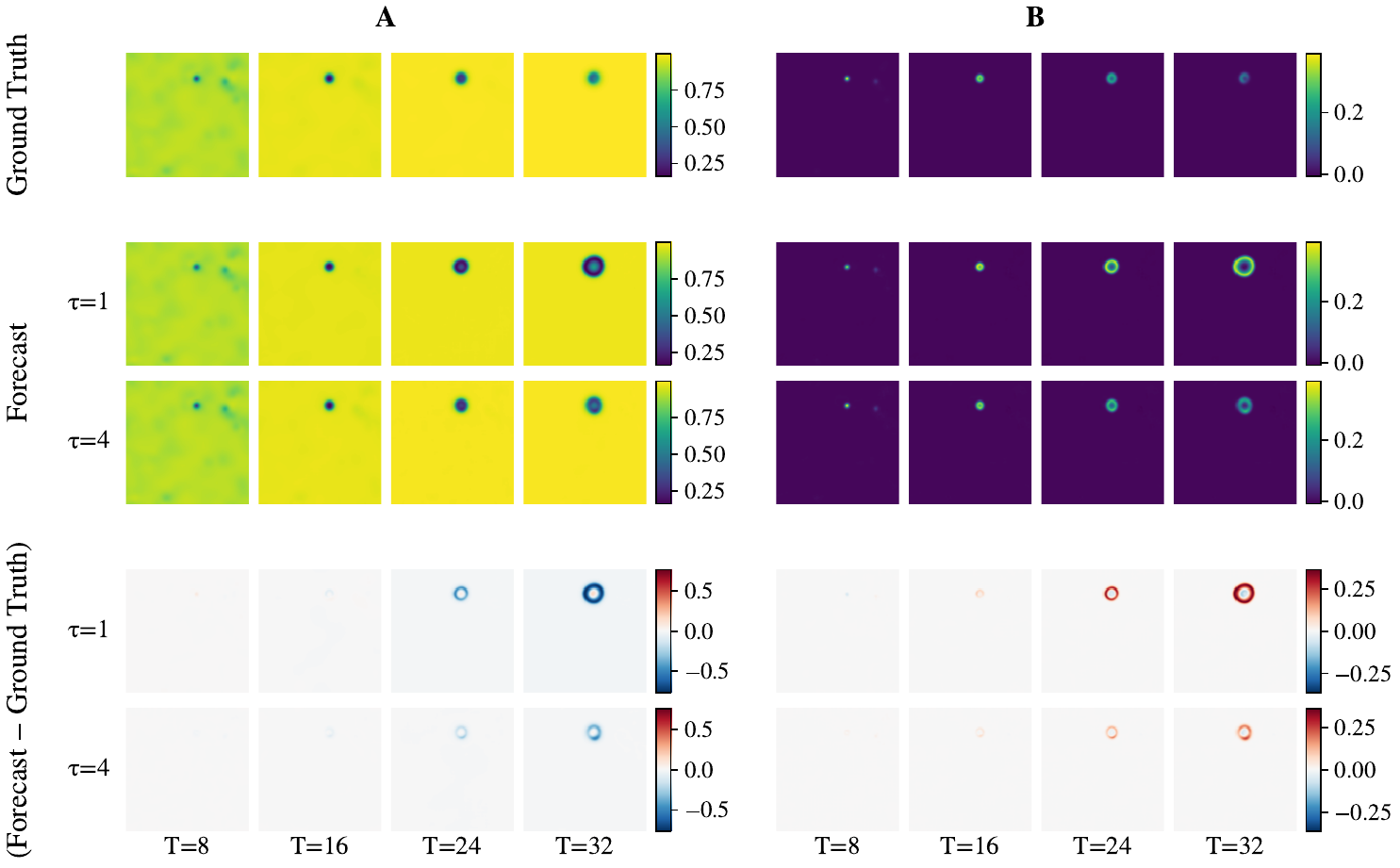}
    
    \caption{Qualitative comparison of forecasting on the \textit{Gray-Scott Reaction Diffusion} dataset for 3 validation samples. The figure displays ground truth and predictions at lead times 8, 16, 24, and 32 for the standard auto-regressive baseline ($\tau=1$) and our large-stride model ($\tau=4$). By forecasting fewer steps to reach the same horizon, the $\tau=4$ model effectively mitigates error amplification as lead time advances.}
    \label{fig:qual_gray_scott_qualitative_dt1_vs_dt4}
\end{figure*}

We investigate qualitatively the effect of using longer time strides $\tau=4$, compared to the $\tau=1$ reference on two datasets from the Well.
In Figure~\ref{fig:qual_dt1_vs_dt4_openBC}, we illustrate how the longer time stride mitigates the appearance of large errors along shock fronts.
At long lead times, we consistently observe reduced errors on the \textit{Euler Multi-quadrants OpenBC} dataset across fields.
We also observe visibly reduced error accumulation as lead time increases on both channels of the \textit{Gray-Scott Reaction Diffusion} dataset in Figure~\ref{fig:qual_gray_scott_qualitative_dt1_vs_dt4}, in line with the quantitative results reported in Figure~\ref{fig:full_time_stride}.
Also supporting the results from Figure~\ref{fig:full_time_stride}, we observe visibly higher errors are soon as lead time $T=4$ which persist at longer horizons for the \textit{Rayleigh-Bénard} dataset (Figure~\ref{fig:qual_dt1_vs_dt4_rb}) and \textit{Active Matter} (Figure~\ref{fig:qual_dt1_vs_dt4_am}). For \textit{Active Matter}, we note that artifacts across lead times are amplified by the larger stride.

\subsection{Impact of regularization}
\label{subsec:qual_impact_gdl_mpr}

\begin{figure*}[htbp]
    \centering
    \begin{subfigure}{\textwidth}
        \centering
        \includegraphics[width=\linewidth]
        {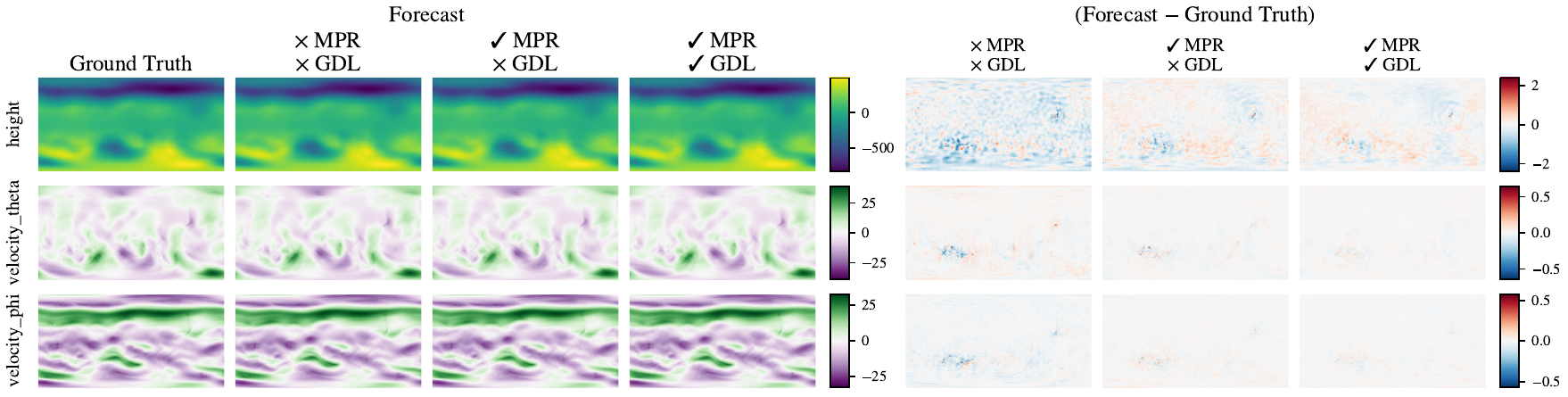}
        \caption{Sample 1, Lead time $T=4$}
        \label{fig:planetswe_sample1_t4}
    \end{subfigure}
    \begin{subfigure}{\textwidth}
        \centering
        \includegraphics[width=\linewidth]{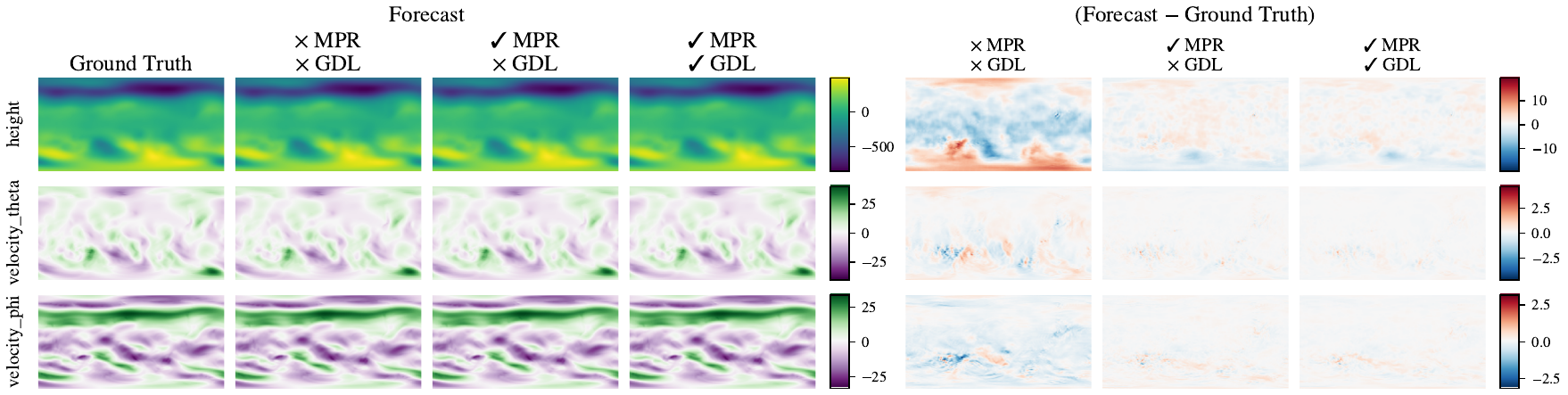}
        \caption{Sample 1, Lead time $T=32$}
        \label{fig:planetswe_sample1_t32}
    \end{subfigure}
    \begin{subfigure}{\textwidth}
        \centering
        \includegraphics[width=\linewidth]{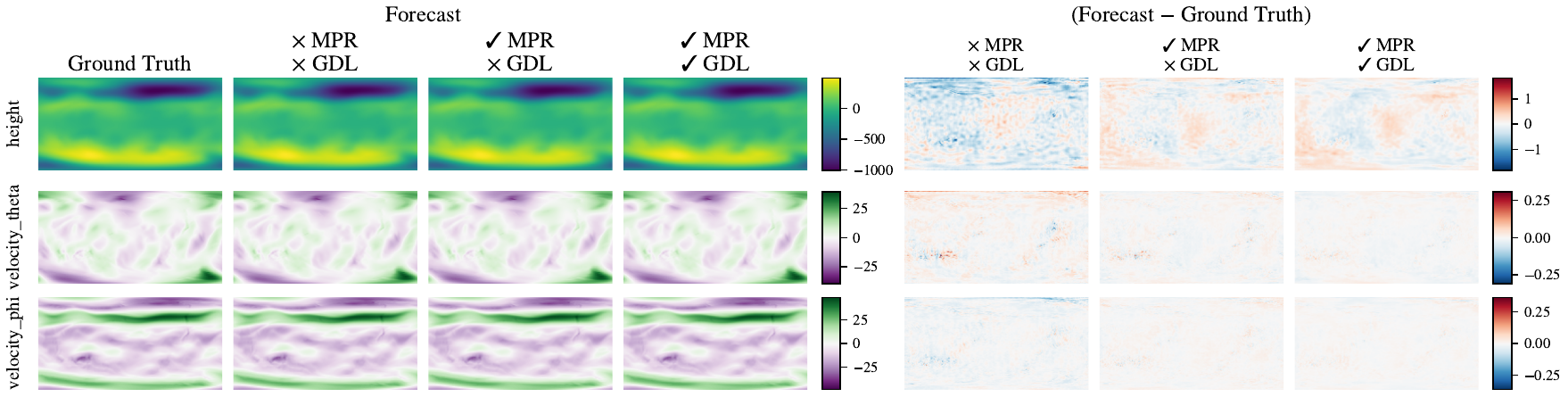}
        \caption{Sample 2, Lead time $T=4$}
        \label{fig:planetswe_sample2_t4}
    \end{subfigure}
    \begin{subfigure}{\textwidth}
        \centering
        \includegraphics[width=\linewidth]{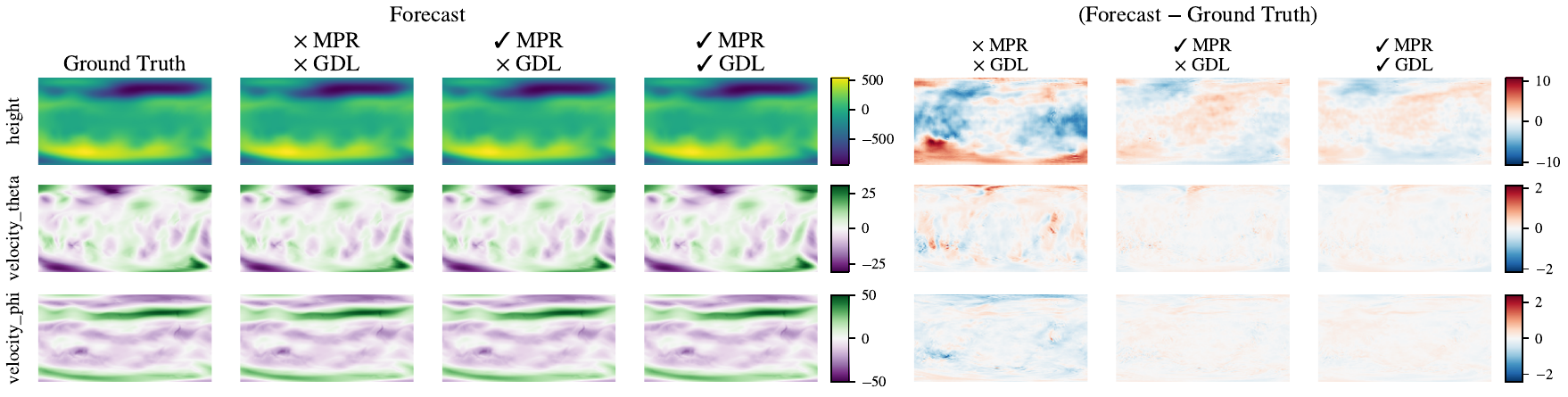}
        \caption{Sample 2, Lead time $T=32$}
        \label{fig:planetswe_sample2_t32}
    \end{subfigure}
    \caption{Qualitative comparison of FGN regularization on two samples from the \textit{PlanetSWE} dataset at short ($T=4$) and long ($T=32$) lead times. We compare the base FGN method without MPR or GDL, FGN with MPR and without GDL, as well as the fully regularized Kastor model with both MPR and GDL. We observe significant artifact reduction for instance in the height variable at short lead times from MPR. Adding GDL further reduces errors visibly in the velocity\_theta channel, especially at short lead times.}
    \label{fig:qual_gdl_mpr_planetswe}
\end{figure*}

\begin{figure*}[htbp]
    \centering
    \begin{subfigure}{\textwidth}
        \centering
        \includegraphics[width=0.8\linewidth]{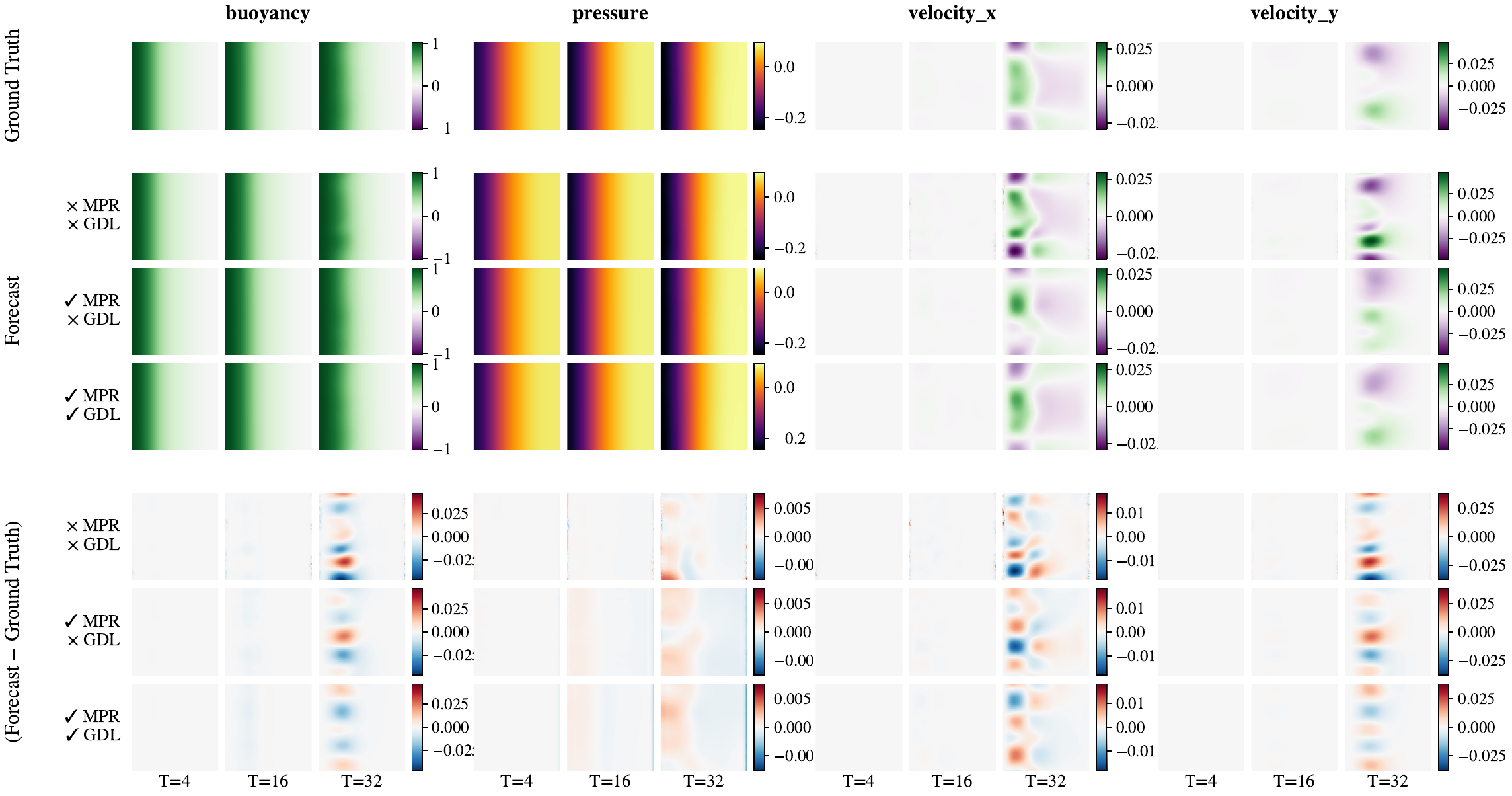}
    \end{subfigure}

    \begin{subfigure}{\textwidth}
        \centering
        \includegraphics[width=0.8\linewidth]{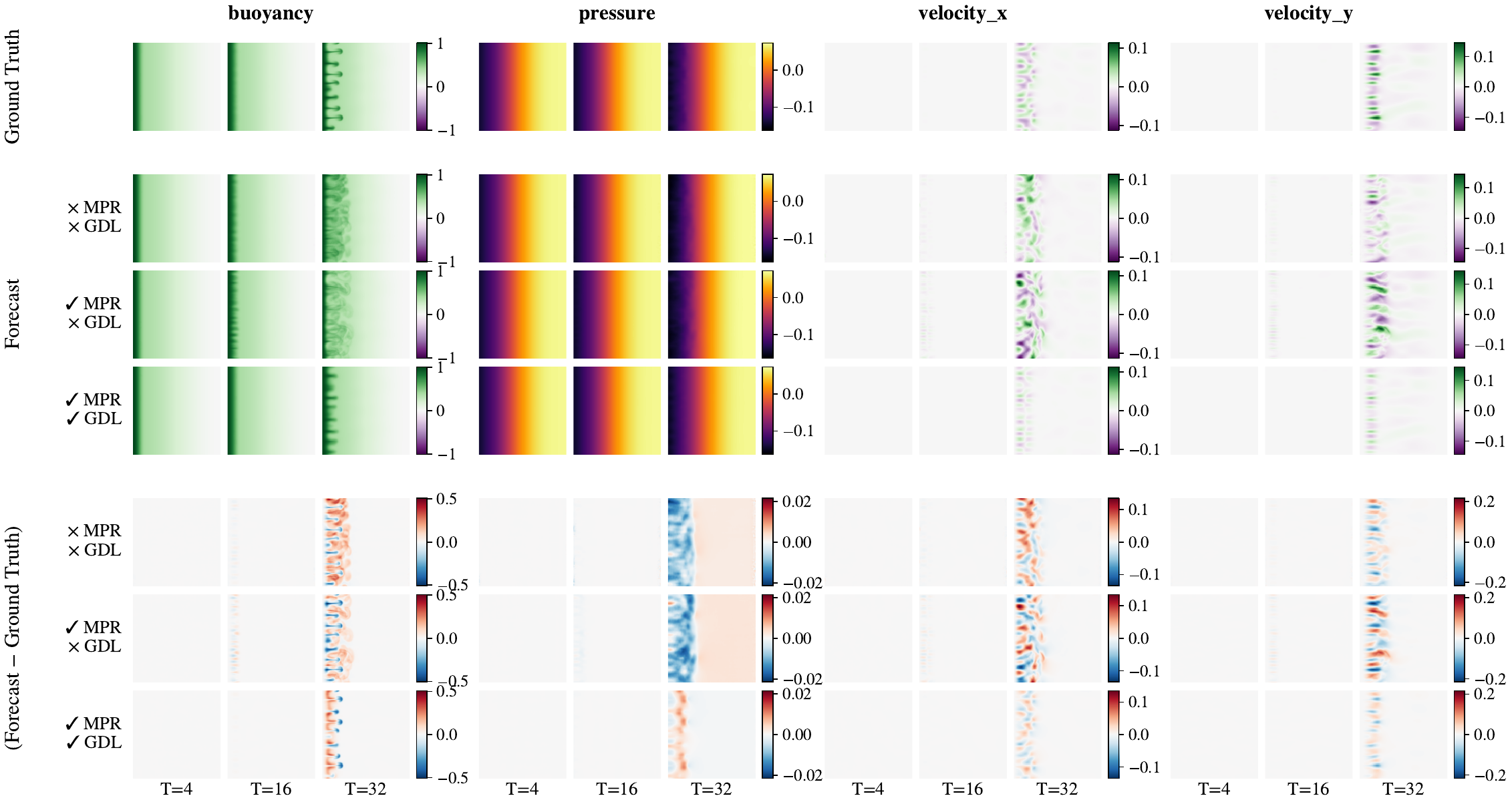}
    \end{subfigure}

    \begin{subfigure}{\textwidth}
        \centering
        \includegraphics[width=0.8\linewidth]{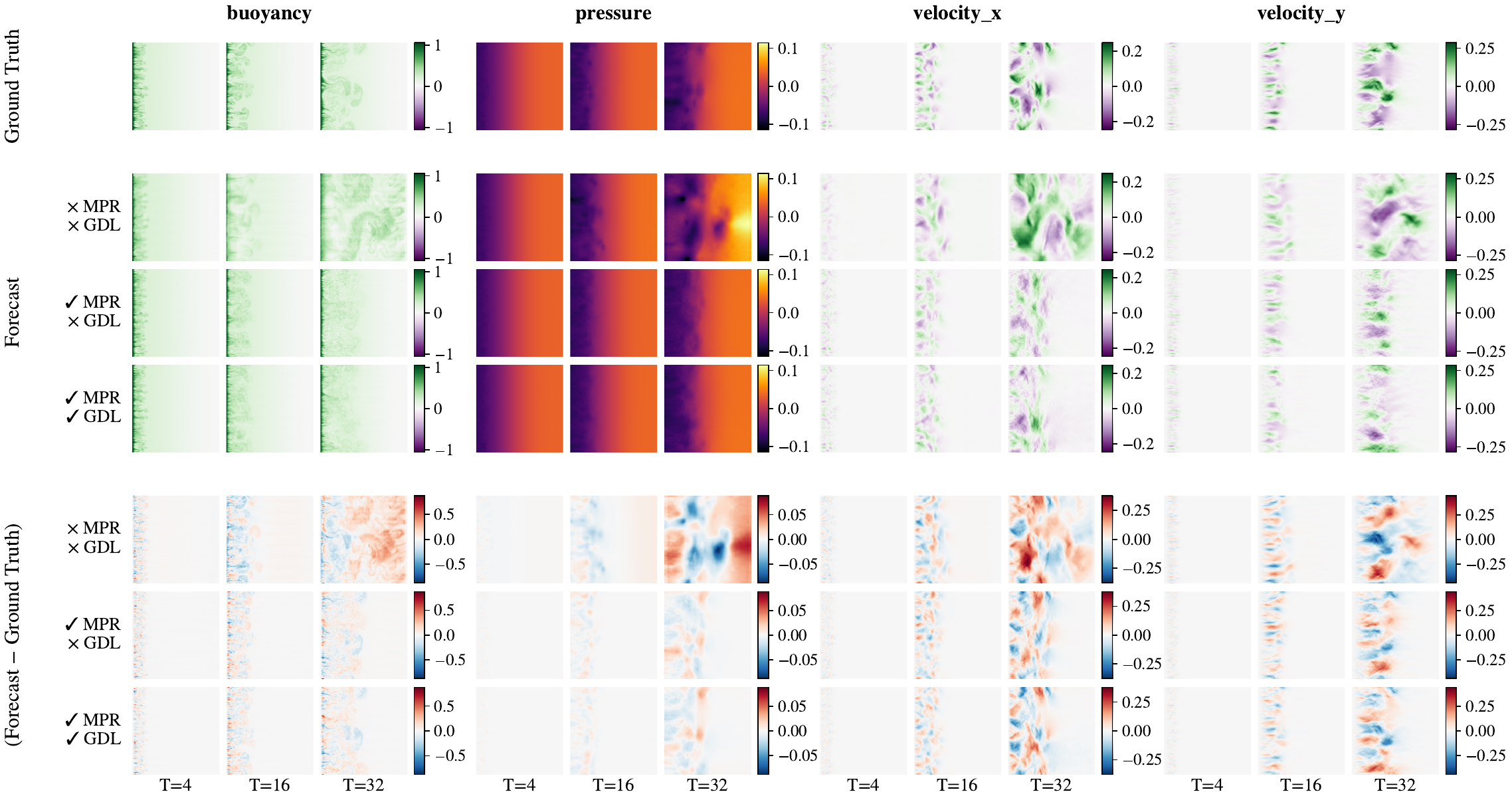}
    \end{subfigure}
    \caption{Qualitative study of the effect of regularization of FGN models on the center-crops of three samples from the \textit{Rayleigh-Bénard} validation set. We illustrate the forecasting performance across different lead times ($T=4$, $T=16$ and $T=32$) and observe that adding MPR is critical to reduce border artifacts across lead times and that GDL further reduces errors at longer horizons.}
    \label{fig:qual_gdl_mpr_rayleigh_benard}
\end{figure*}

\begin{figure*}[htbp]
    \centering
    \begin{subfigure}{\textwidth}
        \centering
        \includegraphics[width=0.7\linewidth]{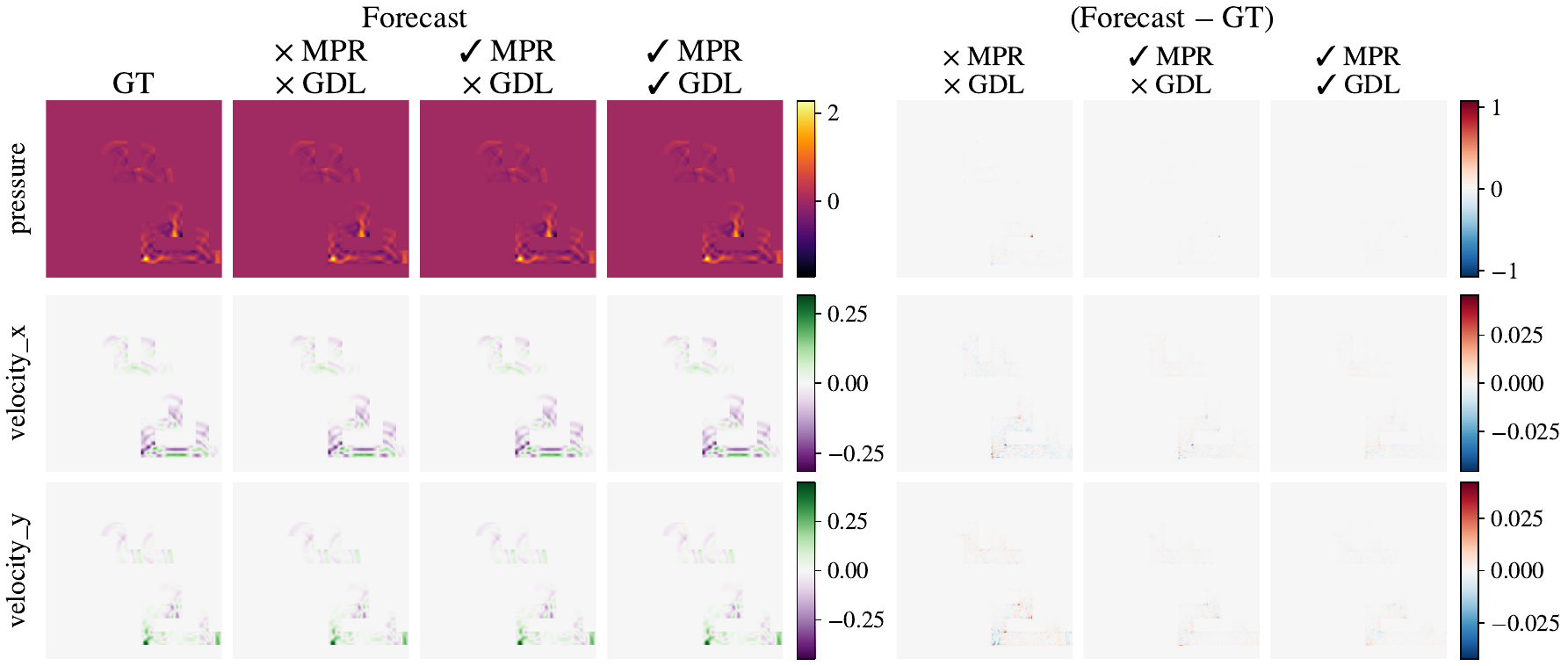}
        \vspace{-0.05cm}
        \caption{Sample 1, Lead time $T=4$}
        \label{fig:acoustic_scattering_maze_sample1_t4}
    \end{subfigure}
    \vspace{-0.05cm}
    \begin{subfigure}{\textwidth}
        \centering
        \includegraphics[width=0.7\linewidth]{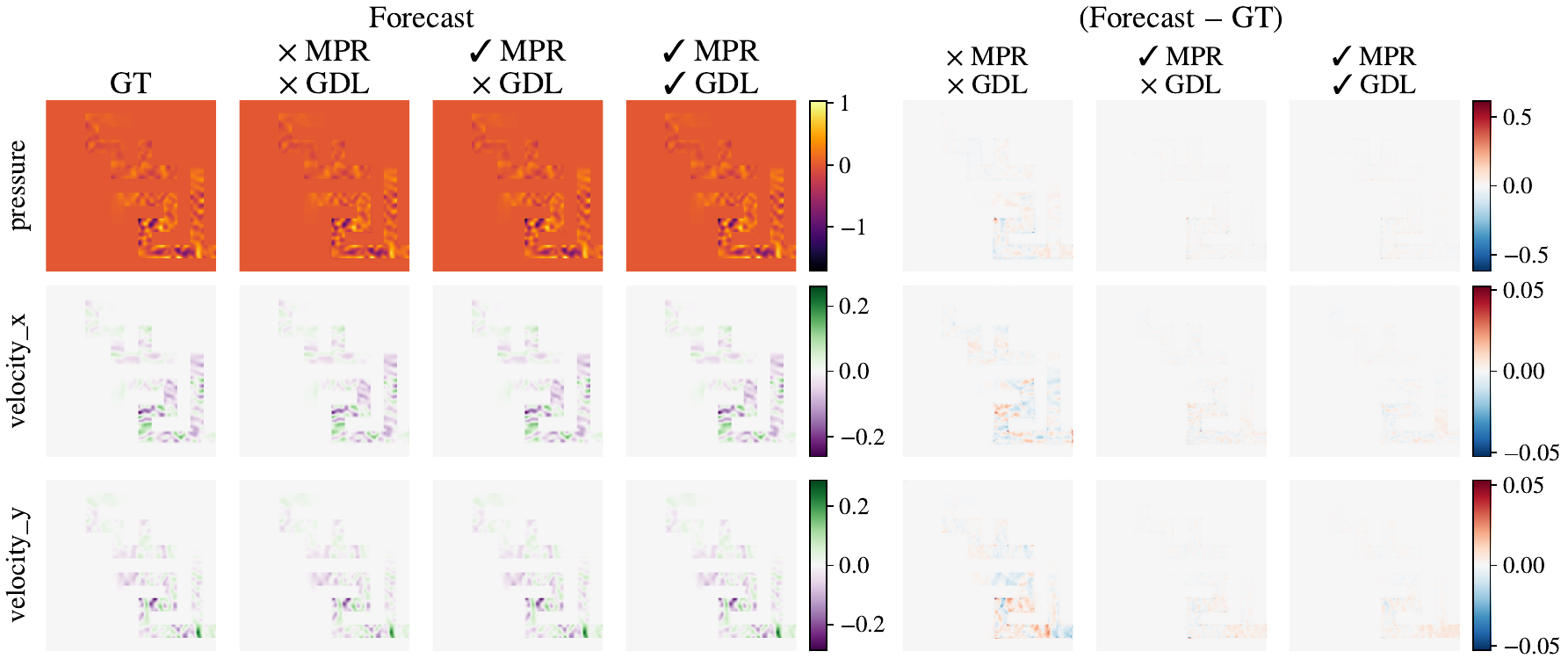}
        \vspace{-0.05cm}
        \caption{Sample 1, Lead time $T=32$}
        \label{fig:acoustic_scattering_maze_sample1_t32}
    \end{subfigure}
    \vspace{-0.05cm}
     \begin{subfigure}{\textwidth}
         \centering
         \includegraphics[width=0.7\linewidth]{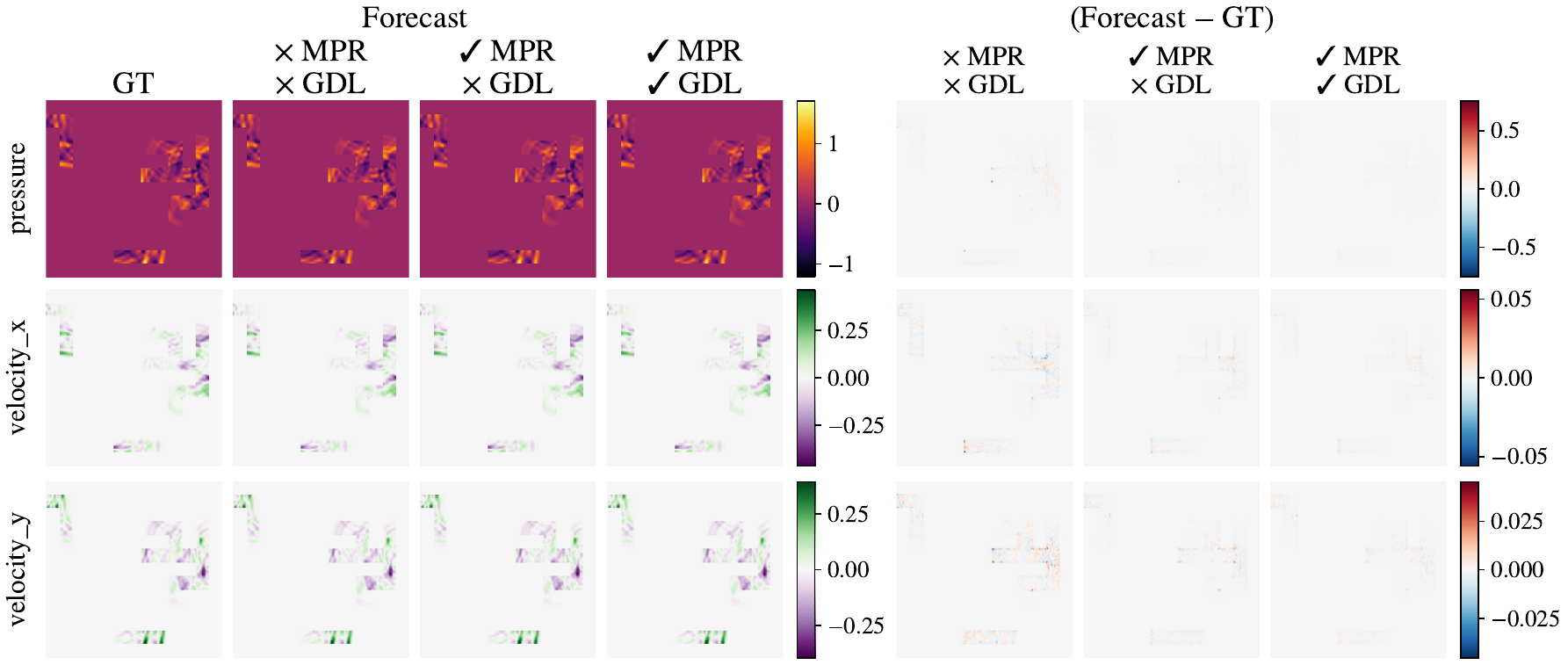}

         \vspace{-0.05cm}
         \caption{Sample 2, Lead time $T=4$}
         \label{fig:acoustic_scattering_maze_sample2_t4}
     \end{subfigure}
     \vspace{-0.05cm}
     \begin{subfigure}{\textwidth}
         \centering
         \includegraphics[width=0.7\linewidth]{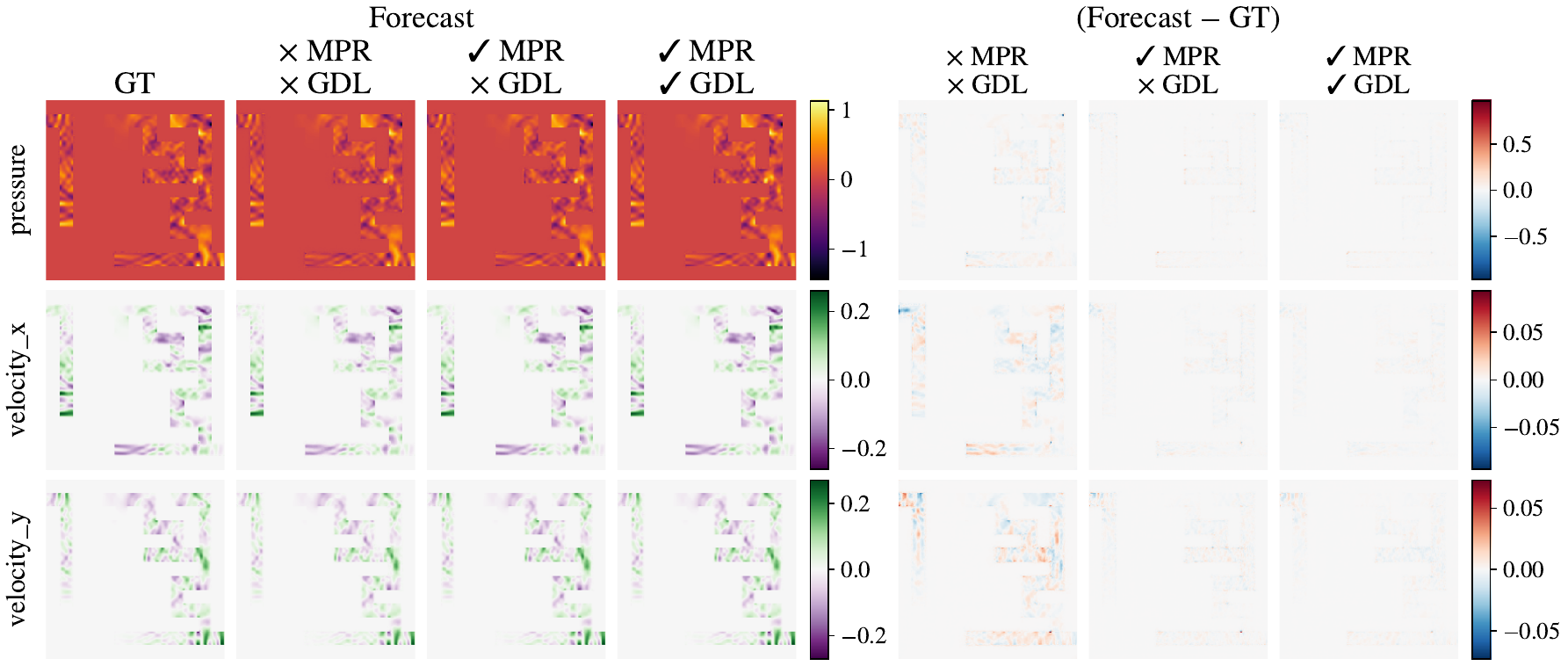}
         \vspace{-0.05cm}
         \caption{Sample 2, Lead time $T=32$}
         \label{fig:acoustic_scattering_maze_sample2_t32}
     \end{subfigure}

    \caption{Qualitative comparison of FGN regularization methods on samples from the Acoustic Scattering Maze dataset across short ($T=4$) and long ($T=32$) lead times. The figures compare the base FGN method without MPR or GDL, FGN with MPR and without GDL, as well as models regularized with both MPR and GDL. We observe visible error reductions both at lead times $T=4$ and $T=32$ when adding MPR and marginal added benefits from further adding GDL, for instance in the $velocity$ fields.}
    \label{fig:qual_gdl_mpr_scattering_maze}
\end{figure*}

We also look into the qualitative changes observed when introducing MPR and GDL (described in Section~\ref{subsec:methods_mpr} and Section~\ref{subsec:methods_gdl} of the main paper).
Consistently with the quantitative results presented in Figure~\ref{fig:mean_pred_reg}, we observe significantly improved forecasts for most samples from \textit{PlanetSWE}, which we illustrate in Figure~\ref{fig:qual_gdl_mpr_planetswe}. We also observe reduced reconstruction errors in a  majority of samples from the \textit{Acoustic Scattering Maze} dataset, for which samples are provided in Figure~\ref{fig:qual_gdl_mpr_scattering_maze}, in particular for the velocity fields at later time steps.
On the \textit{Rayleigh-Bénard} dataset, we find some examples where adding MPR qualitatively degrades long-term forecasts compared to the unregularized baseline, such as in the central sample of Figure~\ref{fig:qual_gdl_mpr_rayleigh_benard}.  Generally, however, forecasts improve, as is most clearly seen at lead time $T=32$ which predictably displays the largest errors.
For the \textit{Rayleigh-Bénard} dataset, we observe consistent improvements from the GDL loss in Figure~\ref{fig:qual_gdl_mpr_rayleigh_benard},  while its effect is less noticeable on \textit{Acoustic Scattering Maze}.
On \textit{PlanetSWE}, GDL seems to reduce artifacts at  both lead times $T=4$ and $T=32$, which is visible for instance in the height fields from the samples in Figure~\ref{fig:qual_gdl_mpr_planetswe},

\section{Additional Details}

\subsection{Datasets}

We evaluate our models on the 2D datasets of the Well: 
\begin{itemize}

\item \textit{Acoustic Scattering}: Models how sound waves propagate through a domain that contains sharply variable densities, such as maze-like walls or pockets of different materials.

\item \textit{Active Matter}: Simulates suspensions of active particles in a viscous fluid that can turn chemical energy into mechanical work, creating complex spatiotemporal dynamics from steric and hydrodynamic interactions.

\item \textit{Euler Multi-quadrants}: Describes the behavior of inviscid, compressible fluids starting from multiple initial discontinuities, leading to interacting shocks and rarefactions.

\item \textit{Gray-Scott Reaction-Diffusion}: Models the spontaneous pattern formation and self-assembly of two chemical species governed by reaction and spatial diffusion.

\item \textit{Planetary Shallow Water Equations} (PlanetSWE): Approximates incompressible fluid flows on a global scale, incorporating Earth's topography (bathymetry) and featuring daily and annual periodic forcings.

\item \textit{Rayleigh-Bénard} Convection: Simulates the fluid dynamics that occur in a horizontal fluid layer when it is heated from below and cooled from above, creating convection currents.
\end{itemize}

We set aside the 3 remaining 2D datasets for the following reasons: \textit{Turbulent Radiative Layer 2D}, which is significantly smaller than the other datasets (only 5GB of data, 10x less than the next smaller datasets \textit{Active Matter} (50GB) and falls outside our target data regime); \textit{Helmholtz Staircase}, because the time-dependence of the system has an analytic solution given the $t=0$ initial condition; and \textit{Visco-Elastic Instability}, for practical reasons, which contains trajectories of variable size and in particular many which cannot be evaluated on 32-steps rollouts.

The datasets have been generated with a deterministic process; however, training generative surrogate models is still useful for two reasons: First, if the data is partially observed (e.g. downsampled), then that introduces stochasticity that we would like to capture; Second, even if the data is completely observed, an ML model cannot represent exactly the underlying function and has some error. A generative model allows to capture this error and to produce a set of trajectories covering potential scenarios that the underlying numerical solver could have produced.

\subsection{Model details}

\paragraph{Architecture.}
We use the architecture of Walrus \cite{mccabe2025walruscrossdomainfoundationmodel}, composed of a shallow convolution-based encoder, a transformer backbone, and a shallow convolutional decoder. The encoder and decoders have variable strides to be able to encode data of various dimensions into the same feature map of size $(32, 32)$. A transformer layer consists of 2 main blocks: first spatial attention is applied independently for every frame in the input sequence; then time attention is applied independently at each spatial location.
Causality is implemented in the time attention via a causal mask, where each frame can only attend to earlier frames in the input sequence.

\paragraph{Patch jittering.} One of the key technique introduced in Walrus is \textit{patch jittering}, which consists in shifting the input data by a small number of pixels before processing with the architecture, and then shifting back by the opposite value at the end. For dimensions with periodic boundary conditions, the input fields can be rolled in that dimension without any problem. For dimensions with open boundary conditions, some padding is added to the data so that it can be shifted and embedded as a slightly larger feature map (33 instead of 32).
In Walrus, this technique was introduced to perform data augmentation and to stabilize rollouts.
We show in this paper that it is also effective both in producing an ensemble of simulation members for the MAE-trained models, and for improving our generative models.

\paragraph{Computational efficiency.}
We show here that our proposed inference scheme with $\tau=4$ is more efficient than $\tau=1$ even accounting for the additional cost of the up-sampling module. We note $C_\text{base}$ the cost of forwarding a single frame through the base model, and $C_{\text{up}}$ the same cost for the up-sampling model. For a single inference on a set of $N_\text{in}$ frames, the forward cost is $C_\text{base}N_{in}$. Through KV caching, we can assume to simplify that causal models only have an additional cost of $C_{\text{base}}$ for each frame during an auto-regressive rollout (In theory it would be slightly more due to time attention along input frames, but this would apply both to our model and the reference). The inference costs for an auto-regressive rollout up to lead time $T$ for the 3 methods in Figure \ref{fig:inference_strat_figure} are:
\begin{align}
    C_\text{standard}(T) &= C_\text{base}* (N_\text{in} + T - 1) \\
    C_\text{kastor}(T) &= C_\text{base}*(N_\text{in} + T/\tau - 1) \\ &\quad\quad + C_\text{up} * (\tau-1)T/\tau\\
    C_\text{blockwise}(T) &= N_\text{in} C_\text{base}T/\tau
\end{align}
Indeed, our model only requires $T/\tau$ auto-regressive steps and uses the non-causal up-sampling model to predict $(\tau-1)$ frames at each rollout step. For the blockwise model, we consider that we predict $\tau$ frames at once with a single model forward so that the number of rollout steps is the same as Kastor. The blockwise model cannot be causal (all forecast frames should have the information of all context frames), no KV caching can be used. With a value of $\tau=4$ and using a half-size up-sampling model ($C_\text{up} = 0.5 C_\text{base}$), we then have

\begin{align}
    C'_\text{standard}(T) &= a + C_{\text{base}}T \\
    C'_\text{kastor}(T) &= a + \frac{5}{8}C_\text{base}T\\
    C'_\text{blockwise}(T) &= \frac{N_\text{in}}{\tau}C_\text{base}T
\end{align}
where $a = C_\text{base}* (N_\text{in} - 1)$. Since $N_\text{in}$ must be larger than $\tau$ for the blockwise auto-regressive baseline, the Kastor model has the smallest linear dependency in $T$. We note that we could also use a full-size model for the up-sampling model $C_\text{up} = C_\text{base}$, in which case our dual-stage model would have the same cost as the standard auto-regressive baseline  $C'_\text{kastor}(T) = a + C_{\text{base}}T$.

\subsection{Training details}

We fine-tune models for 62,500 steps with batch size 8 on a node of 8 H100s. We use the Muon optimizer \cite{jordan2024muon} with learning rate 1e-3 for the 2D matrices and Adam for the embedding weights with a learning rate of 1e-4, $(\beta_1, \beta_2) = (0.9, 0.999)$ and a weight decay of 1e-4. Gradient updates are clipped to a maximum global norm of 10. We use a learning rate scheduler with linear ramp-up for 5000 steps, and cosine decay for the remaining steps.

The combined loss for the FGN model combines the CRPS-based loss with MPR and GDL:
\begin{equation}
    \mathcal{L}_{\text{FGN}} = \lambda_{\text{CRPS}} \mathcal{L}_{\text{CRPS}} + \lambda_{\text{MPR}} \mathcal{L}_{\text{MPR}} + \lambda_{\text{GDL}} \mathcal{L}_{\text{GDL}}
\end{equation}
where we set the loss weights to $\lambda_{\text{CRPS}} = 100$, $\lambda_{\text{MPR}} = 100$, and $\lambda_{\text{GDL}} = 2000$.

\subsection{Diffusion model}

To establish a robust generative baseline, we adapt the Walrus architecture into a diffusion model. While diffusion models require multiple denoising steps—making inference slower than a single forward pass—they provide a strong generative framework that can later be accelerated via distillation \cite{jacq2026diffusion}. Diffusion models require two inputs: context frames $x_{t-k\tau:t}$ and noised \emph{next} frames $\text{Noise}(x_{t-(k-1)\tau:t+\tau}, t_\text{diff}, \epsilon_\text{diff})$, where $0<t_\text{diff}<1$ controls the balance between pure noise and clean frames, and $\epsilon_\text{diff}\sim \mathcal{N}(0, I)$ is sampled pure noise. 

Usually, diffusion models are trained to predict either the original clean data $x_{t-(k-1)\tau:t+\tau}$ or the noise $\epsilon_\text{diff}$. At inference, the model starts from pure noise and denoises it step-by-step until obtaining a clean sample of the next frames. In our case, to better exploit the pretrained Walrus model, we train our diffusion model to denoise the physical state differences $d_{t-(k-1)\tau:t+\tau}$. The noise level $t_\text{diff}$ is provided as an input, encoded with a sinusoidal embedding, and injected into the network via AdaLN at every spatial mixing attention layer in the transformer. We denoise by predicting the clean data ($x_0$-prediction), using the same MAE loss as the one used for pretraining:
\begin{align}
    \hat{d}_{t+\tau}^\text{diff} &= f(x_{t-k\tau:t}, \text{Noise}(d_{t-(k-1)\tau:t+\tau}, t_\text{diff}, \epsilon_\text{diff}), t_\text{diff}) \\
    \mathcal{L}_{\text{MAE}}^\text{diff} &= \lambda(t_\text{diff})||\hat{d}^{\text{diff}}_{t+\tau} - d_{t+\tau}||_1,
\end{align}
where $\lambda(t_\text{diff})$ is a weighting term that depends on the noise level.

To integrate the context and noised inputs, we concatenate them along the channel axis. This allows us to fine-tune separate embedding weights for the context frames and the noised differences (respectively `frames' and `diff' embedding blocks on Figure \ref{fig:generative_methods}). Because this concatenation doubles the number of input channels, the Walrus architecture—which generates one prediction per channel by design—outputs a doubled set of predictions $[\hat{d}_{t+\tau}^1, \hat{d}_{t+\tau}^2]$. Rather than adding a linear projection head or summing these outputs, we strictly use the first half ($\hat{d}_{t+\tau} = \hat{d}_{t+\tau}^1$) to maximally preserve the behavior of the pretrained weights.

We also adapt mean prediction regularization for our diffusion model. For this objective, the model only sees the context frames without any noised inputs (\textit{i.e.}, using just the original subset of channels), and the noise level is set to zero. We compute this loss using the first half of the doubled predictions ($\hat{d}_{t+\tau}^1$), as these weights were already pretrained with a similar deterministic objective. Figure \ref{fig:diffusion} in appendix illustrates the implementation of this diffusion modeling using MPR.

We display on Figure \ref{fig:diffusion} our implementation of the diffusion model for Walrus, and how we applied mean prediction regularization.

\begin{figure*}
\centering
    \includegraphics[width=0.75\linewidth]{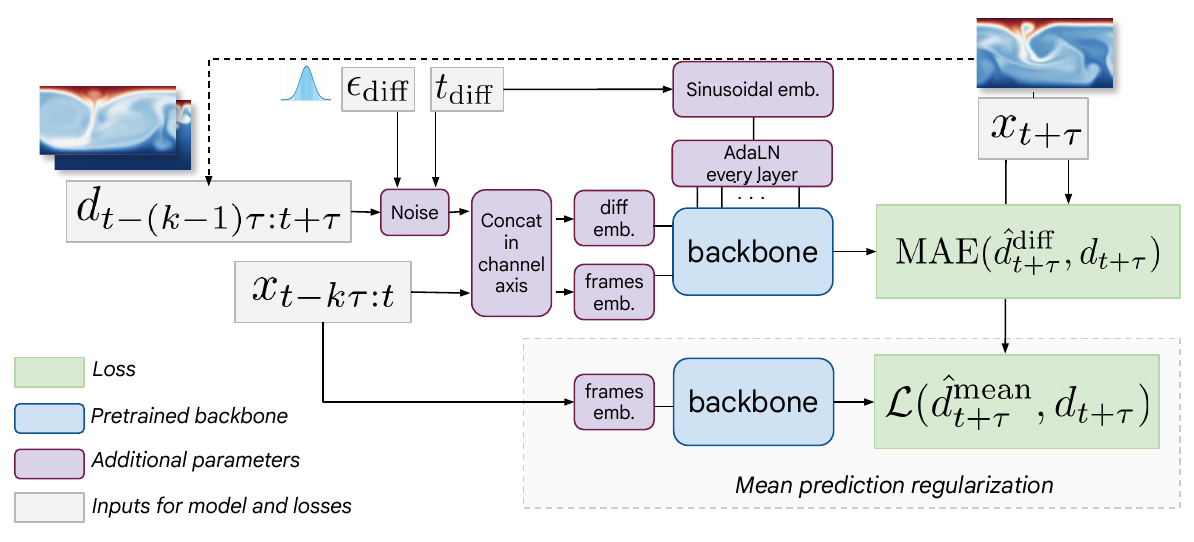}
    \caption{Diagram of Mean Prediction Regularization applied to diffusion modeling.}
    \label{fig:diffusion}
\end{figure*}

\subsection{Normalization}

Unlike Walrus, which uses RMS normalization, we use mean/standard deviation normalization using statistics computed over the full dataset.

Given a dataset of $M$ trajectories containing $N$ total states $x \in \mathbb{R}^{C \times H \times W \times D}$, we compute channel-wise mean and standard deviation vectors for both the states and state deltas across all spatial dimensions $(H, W, D)$ and trajectory transitions:

\begin{equation}
\mu_{x,c} = \frac{1}{N \cdot HWD} \sum_{n=1}^{N} \sum_{i=1}^{H} \sum_{j=1}^{W} \sum_{k=1}^{D} x^{(n)}_{c, i, j, k}
\end{equation}

\begin{equation}
\sigma_{x,c} = \sqrt{\frac{1}{N \cdot HWD} \sum_{n=1}^{N} \sum_{i=1}^{H} \sum_{j=1}^{W} \sum_{k=1}^{D} \left(x^{(n)}_{c, i, j, k} - \mu_c^x\right)^2}
\end{equation}

The statistics for the state deltas ($\mu_{\text{res}}, \sigma_{\text{res}}$) are computed analogously by replacing $x^{(n)}$ with within-trajectory differences $d^{(m,t)}$:
\begin{equation}
d^{(m,t)} = x^{(m,t+1)} - x^{(m,t)}
\end{equation}

where $x^{(m,t)}$ denotes state $t$ in trajectory $m \in M$ of length $T_m$.

To generate a training pair of input state sequence and output delta sequence for training with teacher forcing, we follow the pseudo code presented in Algorithm \ref{alg:global_norm}.

\begin{algorithm}[tb]
\caption{Global normalization for one training sample}
\label{alg:global_norm}
\textbf{Input}: Input context length $N_{\text{in}}$, lead time $\tau$ ($N_{\text{out}} = \tau$), raw trajectory sequence $x = (x_{t-k}, x_{t-k+1}, \dots, x_{t+\tau})$, global statistics $(\mu_x, \sigma_x, \mu_{\text{res}}, \sigma_{\text{res}})$\\
\textbf{Output}: Normalized input sequence $x_{\text{norm}}$, normalized delta target sequence $d_{\text{norm}}$
\begin{algorithmic}[1]
\STATE $k \leftarrow N_{\text{in}} - 1$
\STATE $x_{t-k:t} \leftarrow (x_{t-k}, x_{t-k+1}, \dots, x_t)$ \COMMENT{Extract window of $N_{\text{in}}$ input states}
\STATE $x_{\text{norm}} \leftarrow (x_{t-k:t} - \mu_x) \oslash \sigma_x$ \COMMENT{Normalize input state sequence}
\STATE $d_{t+1:t+\tau} \leftarrow (x_{t+1} - x_t, x_{t+2} - x_{t+1}, \dots, x_{t+\tau} - x_{t+\tau-1})$ \COMMENT{Extract raw delta sequence}
\STATE $d_{\text{norm}} \leftarrow (d_{t+1:t+\tau} - \mu_{\text{res}}) \oslash \sigma_{\text{res}}$ \COMMENT{Normalize target delta sequence}
\STATE \textbf{return} Training pair $(x_{\text{norm}}, d_{\text{norm}})$
\end{algorithmic}
\end{algorithm}

\subsection{Experimental protocol}

Similarly to the methodology in \cite{mccabe2025walruscrossdomainfoundationmodel}, we perform as many rollouts as individual trajectories in the validation/test datasets, choosing the initial context to have the 16th trajectory frame as the last input frame.

Since the beginning of trajectories is often a transient regime, we also evaluate our models starting from the middle frame of dataset trajectories, and report metrics averaged over these two evaluation modes.

\subsection{Metrics}

We depart from the VRMSE metric introduced by the Well \cite{ohana2024well} for two reasons: First, the Root Mean-Square Error (RMSE) is not truly aligned with our goals in as it can be optimized by producing smoother and smoother outputs as the uncertainty grows at higher lead times. Furthermore, the VRMSE score \cite{ohana2024well} removes units by normalizing the RMSE by the spatial variance of the ground truth, however we have found that this gives us scores that can differ by several orders of magnitude and cannot be reliably averaged across datasets and lead times.

We note $||\cdot||_1$ and $||\cdot||_2$ the 1- and 2-norms over the physical space.

\textbf{EnsMeanRMSE.} The ensemble Mean RMSE error ($\operatorname{EnsMeanRMSE}$) for a physical variable $v$ and at a given lead time $\delta$ is calculated as
\begin{equation}
\operatorname{EnsMeanRMSE}(v, \delta) = \frac{1}{|\mathcal{T}|} \sum_{(i,t)\in \mathcal{T}} ||\bar{\hat{x}}_{i, t, \delta} - x_{i, t+\delta} ||_2,
\label{eq:ens_mean_rmse}
\end{equation}
where $\mathcal{T}$ is the set of couples $(i, t)$ with $i$ the trajectory index and $t$ the initialization time in that trajectory. The ensemble mean prediction $\bar{\hat{x}}_{i,t,\delta}$ is the average of the individual predictions across the $M$ members.

\textbf{fCRPS.} Beyond capturing the ensemble mean, we measure how well the empirical distribution marginals matches the ground truth distribution. We use the fair version of the CRPS \cite{ferro2014fair} that is 

\begin{align}
\text{fCRPS}(v, \delta) &= \frac{1}{|\mathcal{T}|} \sum_{(i,t)\in \mathcal{T}} \Big(\frac{1}{M}\sum_{j=1}^M ||\hat{x}^j_{i,t,\delta} - x_{i, t+\delta}||_1 \\
                & - \frac{1}{2M(M-1)} \sum_{j,k} ||\hat{x}^j_{i,t,\delta} - \hat{x}^k_{i,t,\delta}||_1 \Big)
\label{eq:fcrps}
\end{align}

\paragraph{Spread-Skill Ratio.} To evaluate whether the ensemble is reliable and well-calibrated, we measure the relationship between the ensemble spread and the accuracy of the ensemble mean. The ensemble variance ($\operatorname{EnsVariance}$) for a physical variable $v$ and at a given lead time $\delta$ is defined as the average distance of the individual ensemble members to the ensemble mean prediction:
\begin{equation}
\operatorname{EnsVariance}(v, \delta) = \frac{1}{|\mathcal{T}|} \sum_{(i,t)\in \mathcal{T}} \frac{1}{M-1} \sum_{j=1}^M ||\hat{x}^j_{i,t,\delta} - \bar{\hat{x}}_{i,t,\delta}||_2^2.
\label{eq:ens_spread}
\end{equation}
The Spread-Skill Ratio ($\operatorname{SSR}$) is then the ratio of the ensemble spread to the ensemble mean MSE (computed similarly to Equation \ref{eq:ens_mean_rmse} but with squared norm):
\begin{equation}
\operatorname{SSR}(v, \delta) = \sqrt{\frac{M+1}{M}} \sqrt\frac{\operatorname{EnsVariance}(v, \delta)}{\operatorname{EnsMeanMSE}(v, \delta)}.
\label{eq:ssr}
\end{equation}
For a perfectly calibrated ensemble, the spread should match the error of the ensemble mean, resulting in an $\operatorname{SSR}$ close to $1$. An $\operatorname{SSR} < 1$ indicates that the ensemble is under-dispersive, while an $\operatorname{SSR} > 1$ indicates an over-dispersive ensemble.

\paragraph{Log spectral distance (LSD)}
LSD measures the distance of the log of the radially averaged power spectra (RAPSD) $P$, averaged over all isotropic wavenumbers $k$ excluding the DC component ($k=0$). For a physical variable $v$ and at a given lead time $\delta$, LSD is calculated as the average absolute difference over trajectory initializations $\mathcal{T}$:

\begin{equation}
\operatorname{LSD}(v, \delta) = \frac{1}{|\mathcal{T}|} \sum_{(i,t)\in \mathcal{T}} \frac{1}{K} \sum_{k=1}^K \left| \log{\hat{P}_{i, t, \delta}(k)} - \log{P_{i, t+\delta}(k)} \right|
\label{eq:lsd}
\end{equation}

where the RAPSD at isotropic wavenumber $k$ is calculated as the average power of wave components of the 2D power spectrum that share the same radial distance (rounded) from the center. To avoid artifacts that arise from non-square image aspect ratios, we calculate the RAPSD per square patch of the image and average across patches.

\section{Additional Experimental results}

\subsection{Reproduction of Walrus fine-tuning methodology}

\begin{figure*}[htpb]
    \centering
    \includegraphics[width=\linewidth]{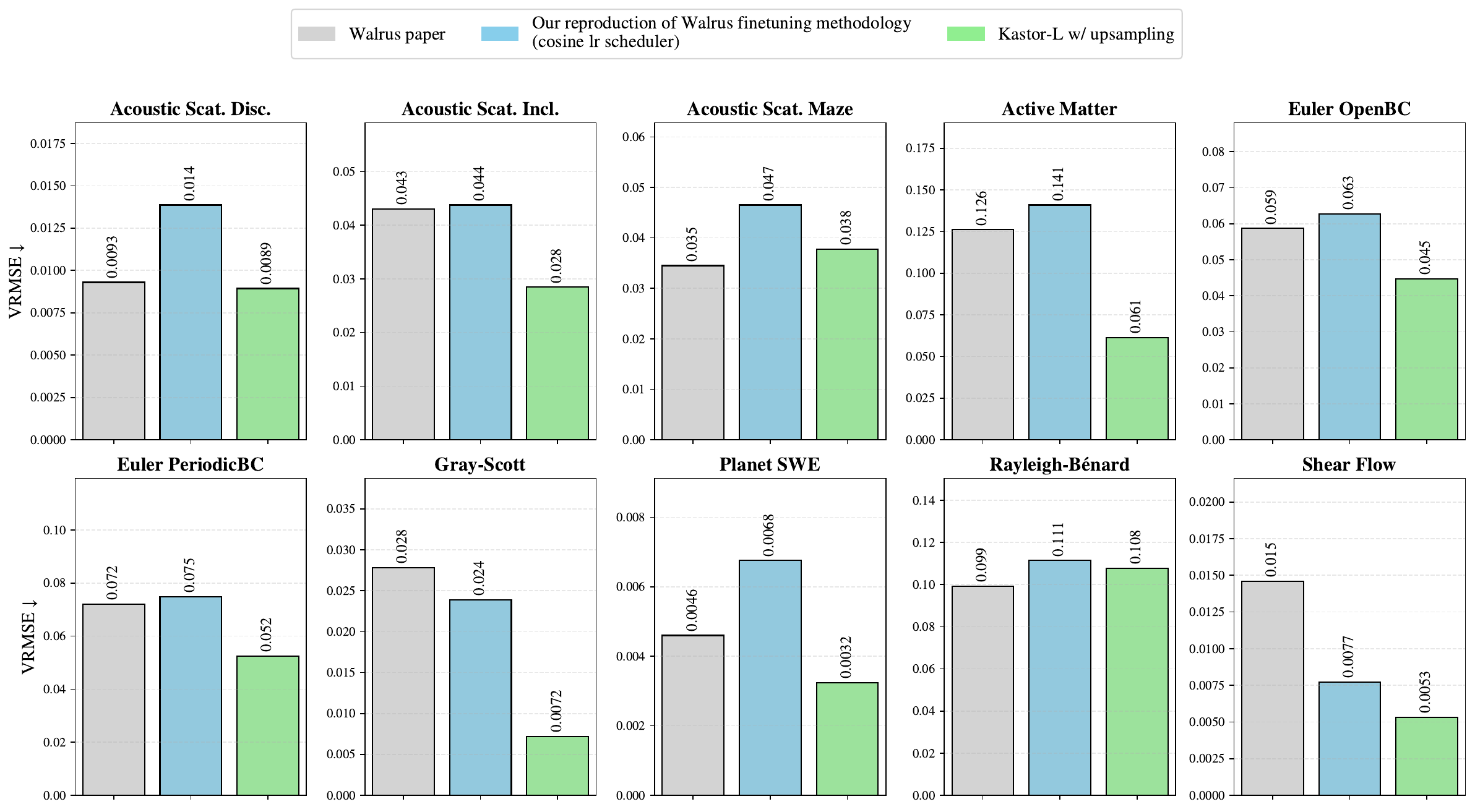}
    \caption{Median VRMSE for lead times $T=1-20$ for our reproduction of the Walrus finetuning methodology, which we used as reference for skill score computations, Kastor-L w/ upsampling compared to the scores reported in \cite{mccabe2025walruscrossdomainfoundationmodel}. Our reproduction performs similarly to the reported Walrus paper on most datasets, with the exceptions of Acoustic Scattering Discontinuity, Acoustic Scattering Maze and PlanetSWE where it performs noticeably worse, and Shear Flow, where it outperforms the reference.}
    \label{fig:median_vrmse}
\end{figure*}

In the paper, we retrain a model with the fine-tuning methodology of Walrus, that we use as reference to compute the skill scores of our other models. We analyze here this retrained model compared to the scores reported in \cite{mccabe2025walruscrossdomainfoundationmodel} using the Median VRMSE metric.
While trying to reproduce exactly the methodology, the model had worse performance than reported in \cite{mccabe2025walruscrossdomainfoundationmodel}, and early experiments indicated that the learning rate scheduler is an important design choice that was left ambiguous in the paper (it was simply indicated that it is the same as for the pretraining phase, with inverse square root ramp-up and inverse square root decay). To bridge part of the performance gap, we retrained the model with a scheduler which performs 5000 steps of linear warmup and applies cosine decay for the remaining steps, which performs close to the reported model as we see in Figure \ref{fig:median_vrmse}: Comparing the 1-step median VRMSE and VRMSE averaged over frames 1-20, our retrained model performs better than the reference on 2/3 datasets and worse on the remaining 7/8 datasets.
This model with cosine learning rate scheduler is the one we chose as base for computing skill scores.
We also compare to the final Kastor-L w/ upsampling model, and find that our model outperforms the reported scores from Walrus for 8 out of 10 datasets (all except \textit{Acoustic Scattering Maze} and \textit{Rayleigh-Bénard}), showing the strength of our proposed method.

\subsection{Fundamental model choices}

\begin{figure*}
    \centering
    \includegraphics[width=\linewidth]{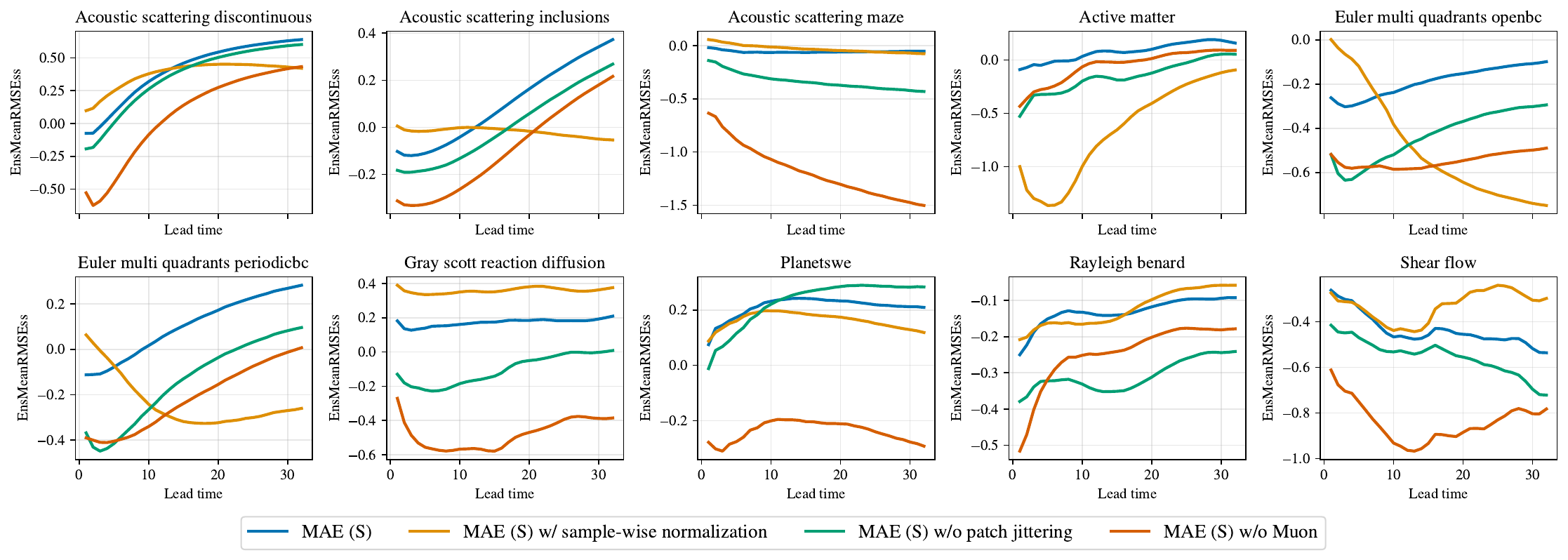}
    \caption{RMSE relative improvement (skill score) of the Ensemble mean for various configurations. Patch jittering, global normalization, and the Muon optimizer are essential components for best performance.}
    \label{fig:fundamental_exps_dt1}
\end{figure*}

In Figure \ref{fig:fundamental_exps_dt1}, we analyse the impact of some fundamental training choices, with our MAE-trained models (size S) a fixed time stride of 1. We can see that patch jittering helps to get better scores across all datasets, except \textit{PlanetSWE}. For normalization, interestingly, using sample-wise normalization is better than global normalization (our default) at small lead times, but becomes less competitive at higher lead times. The pretrained model being trained with sample-wise normalization, it is probable that the fine-tuned model better leverages pretrained features; however since sample-wise statistics are re-computed at each rollout step, it might be a less stable strategy than global normalization. Combining the two for best overall performance is an interesting avenue for future work. Finally, using the Muon optimizer \cite{jordan2024muon} is critical to get good performance, rather than the original Adam optimizer \cite{kingma2014adam} used for pre-training Walrus.

\subsection{Full time stride ablation}

\begin{figure*}[htbp]
    \centering
\includegraphics[width=\linewidth]{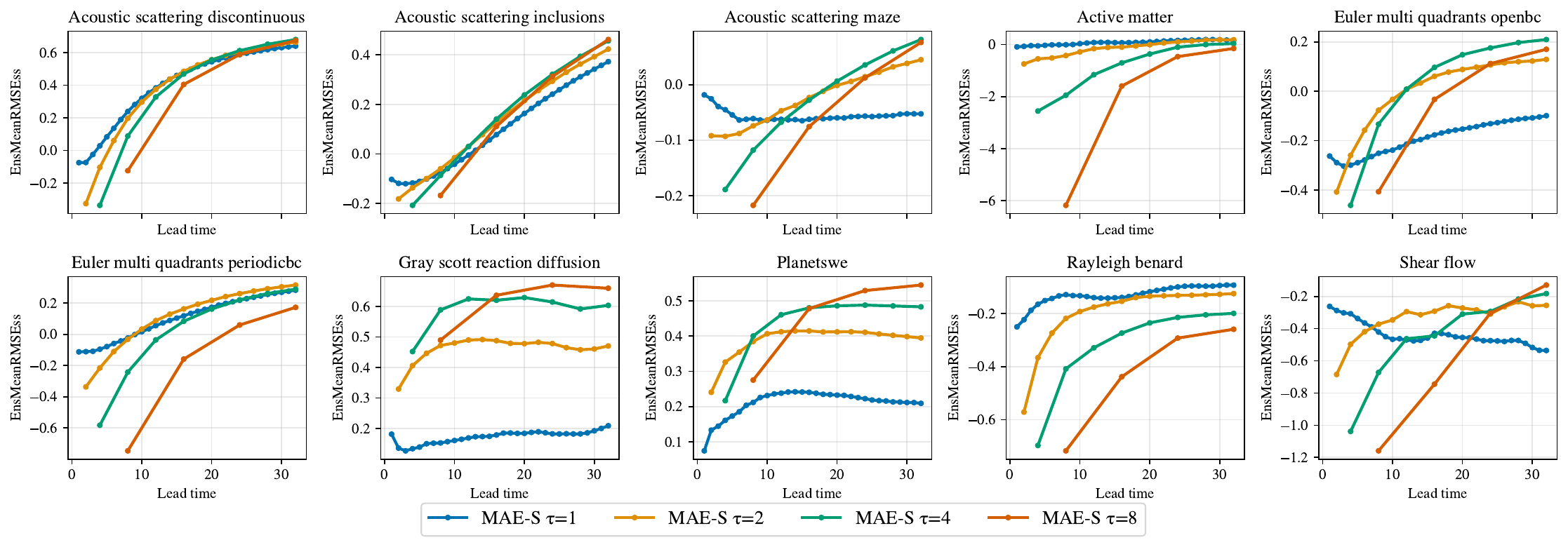}
    \caption{Per-dataset comparison of models trained with different time stride $\tau$. Models are evaluated with Ensemble Mean RMSE skill score per lead time. The best time stride is dataset-dependent, and $\tau=4$ strikes a reasonable trade-off for all datasets except \textit{Active Matter} and \textit{Rayleigh-Bénard}, where $\tau=1$ outperforms the other strides for all lead times.}
    \label{fig:full_time_stride}
    
\end{figure*}

In Figure \ref{fig:full_time_stride}, we show the performance of MAE-trained models with different time stride $\tau$, for different datasets. We observe that generally, the datasets with time stride are better for higher lead times, probably due to less error accumulation. Interestingly, the optimal time stride depends on lead time and in many cases, it is better to use lead time of $\tau=1$ for small lead times: e.g. for lead time 4, in 6 datasets out of 10, rolling out a $\tau=1$ model four times gives better results than training with $\tau=4$, while it is the opposite at longer lead times. Given those results, it might be that a combination of those two strides can give the best results, e.g. starting a rollout with $\tau=1$ for four steps and continuing with $\tau=4$, which we leave for future exploration.

\subsection{Multi-step rollout supervision}

\begin{figure*}[htbp]
    \centering
\includegraphics[width=\linewidth]{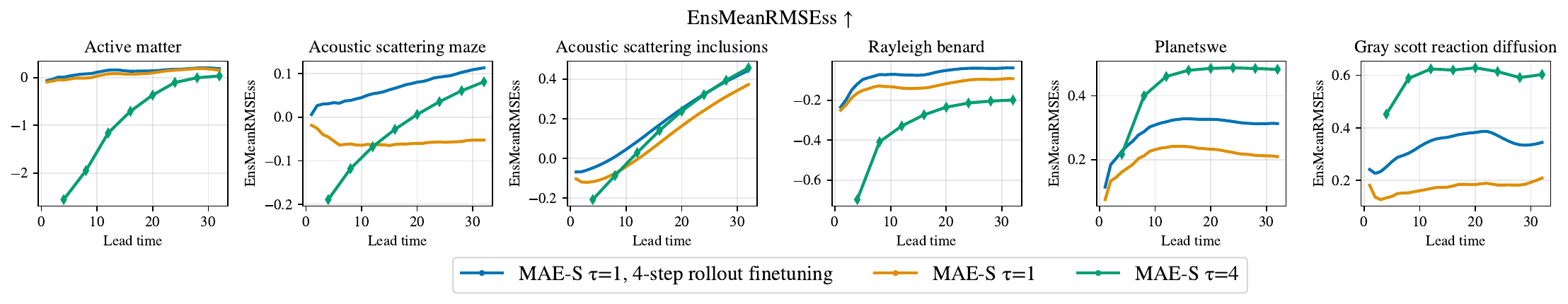}
    \caption{Per-dataset comparison of a $\tau=1$ MAE-S model trained with supervision on 4 consecutive rollout steps, compared to MAE-S models with time strides $\tau=\{1,4\}$. Models are evaluated with Ensemble Mean RMSE skill score per lead time. Multi-step finetuning improves performance compared to the original $\tau=1$ model, at an added computational cost. However, it only partly bridges  the gap for datasets where $\tau=4$ outperforms $\tau=1$ for most lead times.}
    \label{fig:finetune_4_output_steps}
    
\end{figure*}

An alternative method to limit  artifacts and error accumulation for forecasting is to perform multiple rollout steps during training and apply losses on the generated trajectories.
We compare this strategy to our strided training with $\tau=4$.
In practice, we finetune the auto-regressive MAE-S model with stride $\tau=1$ with an additional MAE loss between the 4 rollout steps from the predicted trajectory and the corresponding normalized target trajectory.
We add this additional multi-step loss with the same weight as the original MAE objective, and finetune for 20k additional training steps with a learning rate of $1e^{-5}$ for a cosine decay schedule following 5000 steps of linear warmup, using the same optimizer as the default MAE-S model. 
In Figure \ref{fig:finetune_4_output_steps}, we compare the performance of this further finetuned model with the $\tau=1$ MAE-S model (from which the multi-step rollout finetuning is resumed) and the strided  $\tau=4$ MAE-S model on 6 datasets.
We observe that for the datasets for which $\tau=4$ generally outperforms $\tau=1$, such as \textit{PlanetSWE} and \textit{Gray-Scott Reaction Diffusion}, adding the loss on multiple rollout steps allows to partly bridge the gap between these two models, though the $\tau=4$ model remains stronger for most lead times.
For the Acoustic Scattering Inclusions datasets, where $\tau=1$ performs better for short lead times, we observe that the added 4-step rollout loss improves upon both the $\tau=1$ and $\tau=4$ models, performing as well or better as the $\tau=1$ model at early lead times and competitively to $\tau=4$ at later horizons.
For the Active Matter and Rayleigh-Bénard datasets, multi-step rollouts provide an additional improvement compared to the leading $\tau=1$ models.
Note that multi-step rollout finetuning incurs a significant computational overhead compared to the autoregressive baselines.

\subsection{Scaling experiment}

\begin{figure*}
\centering
    \includegraphics[width=\linewidth]{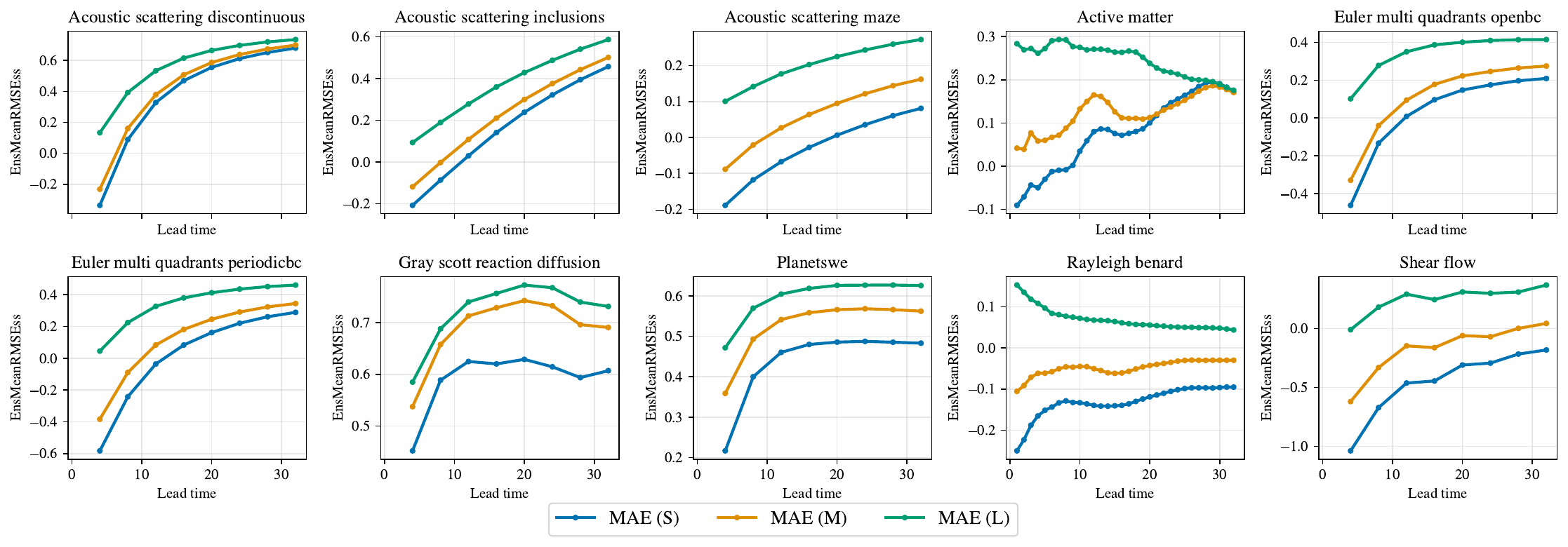}
    \caption{Improvement in forecasting error (EnsMeanRMSE skill score) for our S (20 layers, 2 input frames), M (20 layers, 4 input frames) and L (40 layers, 4 input frames) models.}
    \label{fig:scaling_experiment}
\end{figure*}

In Figure \ref{fig:scaling_experiment}, we show the forecasting accuracy of our MAE-trained models with our three model sizes (S, M, L). As expected, more inference-time compute budget gives higher skill scores (lower forecasting errors).

\subsection{Key FGN design choices}

\begin{figure}
    \centering
    \includegraphics[width=\linewidth]{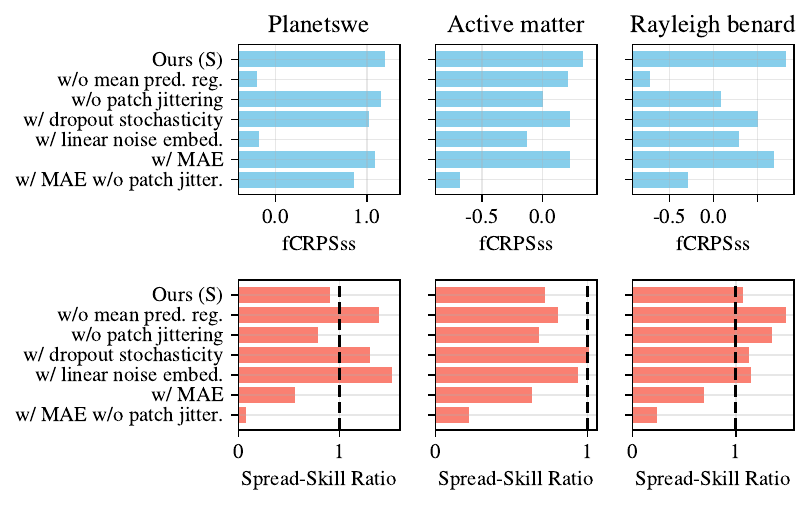}
    \caption{Forecasting error improvement (fCRPS skill score) and Calibration (Spread-skill ratio) for various configurations of our FGN-model (S size, w/o Gradient Difference Loss).}
    \label{fig:ablation_fgn}
\end{figure}

We show in Figure \ref{fig:ablation_fgn} the accuracy gain brought by key method components. First, patch jittering is a form of stochasticity injected into the architecture, and it could be considered as redundant with the AdaLN noise conditioning of FGN. We find that removing patch jittering actually significantly deteriorates performance, and that the model strongly benefits from having these two sources of stochasticity. We also evaluate whether dropout can be used as another form of noise injection instead of AdaLN conditioning, similarly to \cite{cachay2026u}, and find that it does not perform as well. Using an MLP with 2 layers to embed the noise instead of linear embedding is important to correctly map the input noise distribution to the target distribution.

\subsection{Ablation of Gradient Difference Loss}

\begin{figure*}
    \centering
    \includegraphics[width=\linewidth]{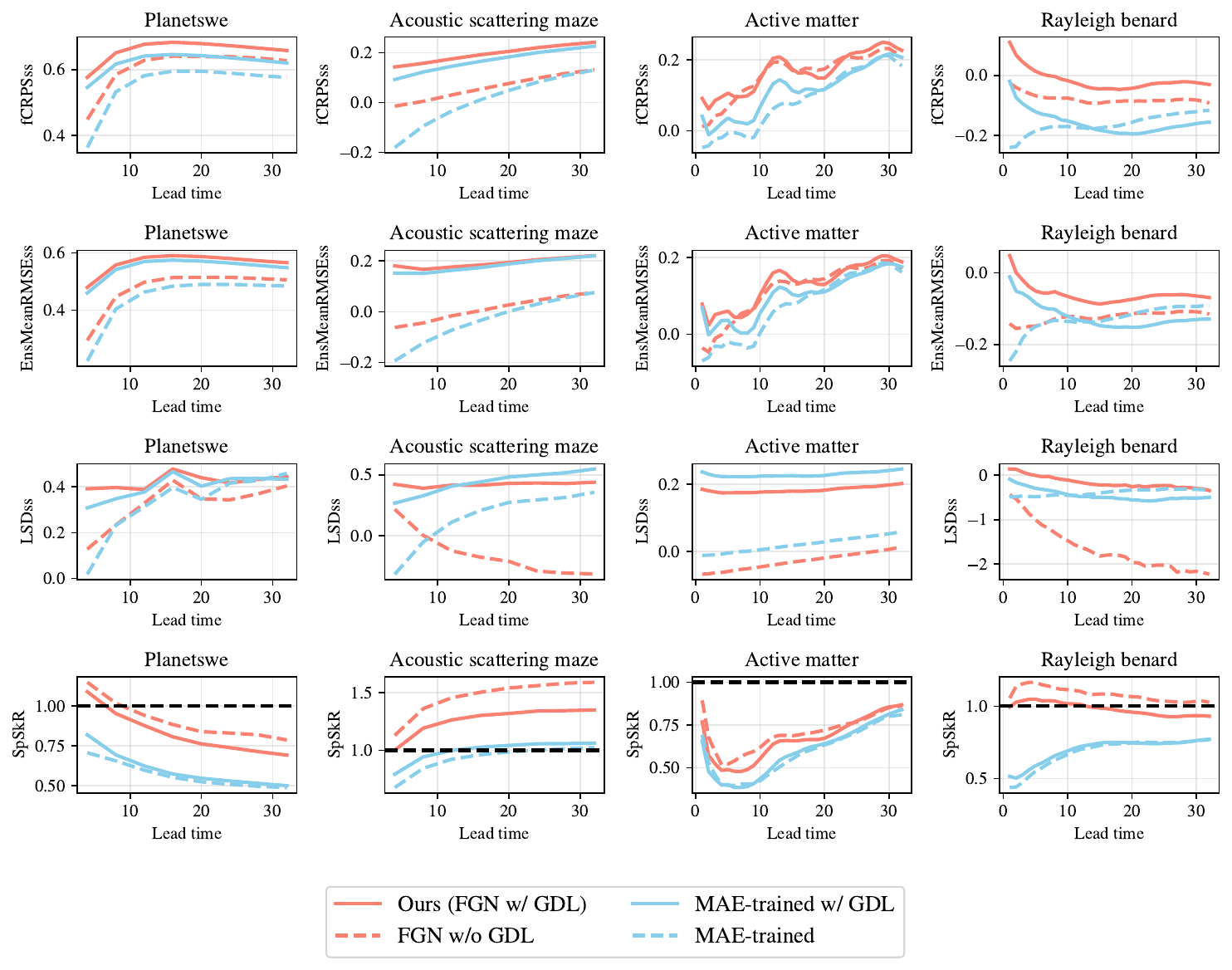}
    \caption{Overall improvement (higher) of forecasting error (fCRPSss and EnsMeanRMSEss), spectral consistency (LSDss), and calibration (SpSkR) for our FGN model (size S) with Gradient Difference Loss (GDL) across lead times. Comparison on MAE-trained baseline shows that GDL can also bring similar improvements to the deterministic model.}
\label{fig:gdl_summary_per_leadtime}
\end{figure*}

\begin{figure}
    \centering
    \includegraphics[width=\columnwidth]{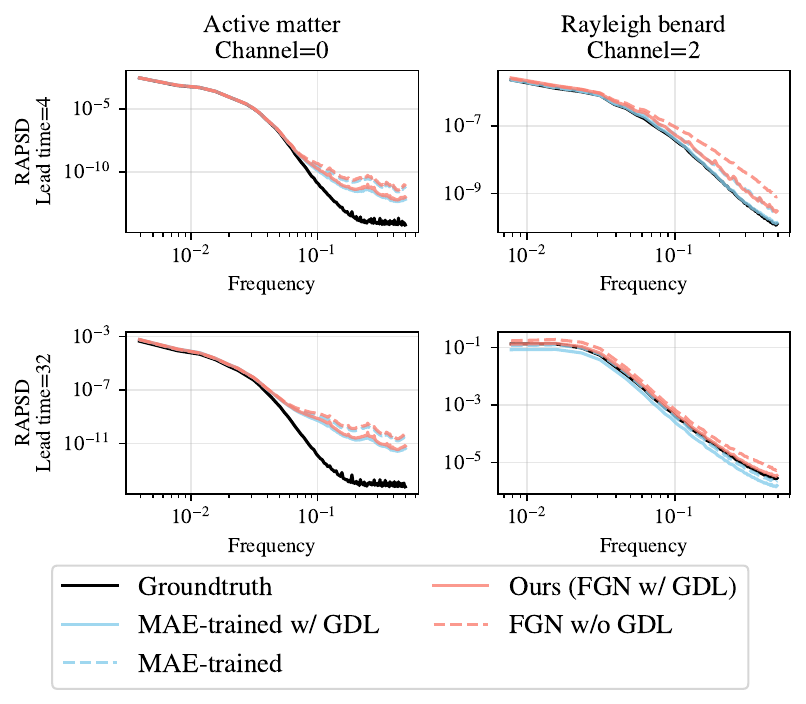}
    \caption{Radially-averaged power spectra (RAPSD) for selected variables over two lead times. GDL brings the FGN power spectra closer to ground truth across datasets.}
    \label{fig:gdl_power_spectrum}
\end{figure}

In Figure~\ref{fig:gdl_summary_per_leadtime}, we show the performance per lead time of FGN and the MAE-trained models (size S) with and without Gradient Difference Loss (GDL).
In general, across the 4 representative datasets, GDL improves forecast error (measured by fCRPSss and Ensemble Mean RMSEss), and spectral error (LSDss) across lead times. For both \textit{Acoustic Scattering Maze} and \textit{Active Matter} datasets, the MAE-trained model with GDL has slightly better LSDss than FGN with GDL, but at the cost of forecast error. Also, FGN without GDL exhibits a larger gap in LSDss with the MAE-trained model, indicating that GDL is crucial in maintaining spectral consistency in the FGN model.

Figure \ref{fig:gdl_power_spectrum} visualizes the improvement that GDL brings to the radially-averaged power spectra (RAPSD) for selected variables. Interestingly, for the \textit{Rayleigh-Bénard} dataset, GDL actually worsens power spectra (more blurry at high frequencies) for the MAE-trained model at longer lead times, which can be visualized in the bottom right.

\subsection{Residual modeling}

\begin{figure}[htbp]
\centering
    \includegraphics[width=\linewidth]{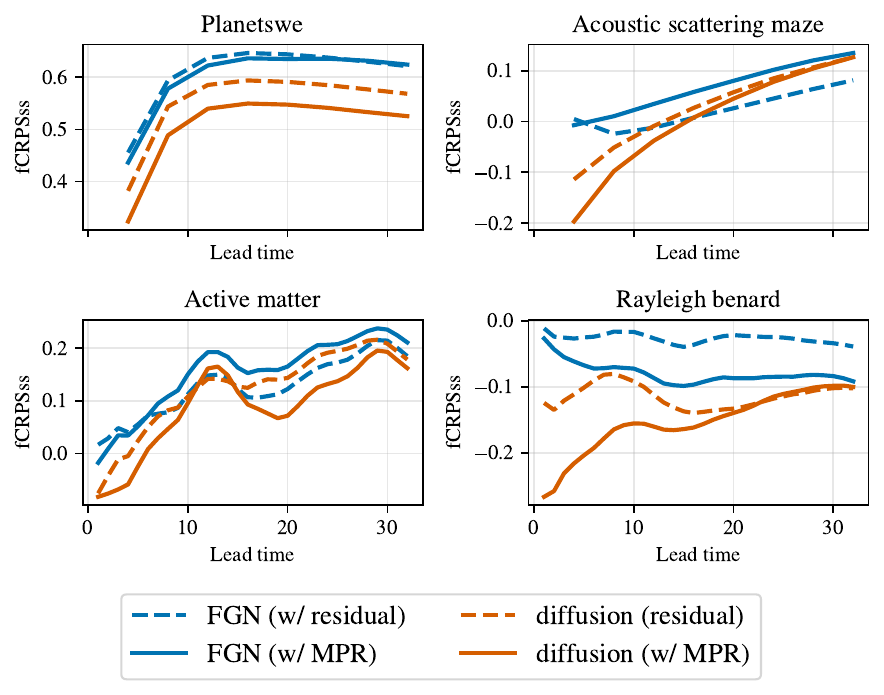}
    \caption{Forecasting error (fCRPSss) of our models compared to residual models.}
    \label{fig:exp_with_res}
\end{figure}

In this paper, we have used \textit{Mean Prediction Regularization} as a way to regularize generative models with the mean of the ground truth distribution. Another method developed with the same goal is residual models \cite{archesweather}, where a deterministic MAE-trained approximates the distribution mean $\hat{d}_{t+\tau}^\text{mean}$, and a generative model can sample from the distribution of rescaled residuals. While this can improve generative modeling by rescaling the distribution, it is also more costly as it requires at least two model evaluations for a single rollout step, one for the deterministic model and one for the generative. We implement residual models for both FGN and diffusion. We note that for residual models, \textit{Mean Prediction Regularization} does not apply, as the mean is given by a separate model.

\paragraph{Residual models with FGN}
Within the FGN framework, residual modeling requires two models. The first model $f_\theta$ is trained with MAE; For the generative model $g_\theta$, we define the predictions
\begin{align}
    \epsilon_1, \epsilon_2 &\sim \mathcal{N}(0, I_{d_\text{noise}}) \\
    \hat{d}^1_{t+\tau} &= \text{SG}[f_\theta(x_{t-k\tau:t::\tau})] + \sigma g_\theta(x_{t-k\tau:t::\tau}, \epsilon_1) \\
    \hat{d}^2_{t+\tau} &= \text{SG}[f_\theta(x_{t-k\tau:t::\tau})] + \sigma g_\theta(x_{t-k\tau:t::\tau}, \epsilon_2) \\
    \label{eq:fcrps_train}
\end{align}
where $\sigma$ is the error of the $f_\theta$ computed on the training set, and $\text{SG}$ is the stop-gradient operator. Those predictions are then trained with fCRPS as in the non-residual case.

\paragraph{Diffusion and residual prediction.}
For diffusion, we define the target residual: 
\begin{equation}
    \varepsilon_{t+\tau}=(d_{t+\tau} - \text{SG}[\hat{d}_{t+\tau}^\text{mean}])/\sigma_{t+\tau},
\end{equation}
where $\sigma_{t+\tau}$ is a moving average of the RMSE between $\hat{d}_{t+\tau}$ and $d_{t+\tau}$. We train the model to predict the residuals: 
\begin{align}
    \hat{\varepsilon}_{t+\tau} &= f(x_{t-k\tau:t}, \text{Noise}(d_{t-(k-1)\tau:t+\tau}, t_\text{diff}, \epsilon_\text{diff}), t_\text{diff}) \\
    \mathcal{L}_{\text{MAE}}^\text{residual} &= \lambda(t_\text{diff})|\hat{\varepsilon}_{t+\tau} - \varepsilon_{t+\tau}|_1.
\end{align}
Because this objective changes the modality of the target, we predict the residual using the \emph{second} half of the doubled predictions ($\hat{d}_{t+\tau}^2$). This allows the first half of the predictions to remain dedicated to the mean prediction task during the same forward pass.

We evaluate residual models for FGN and diffusion in Figure \ref{fig:exp_with_res}. While residual modeling brings improvements for diffusion, the results is more mixed for FGN, with a notable improvement in \textit{Rayleigh-Bénard} and some degradation in \textit{Active Matter} and \textit{Acoustic Scattering Maze}.

\subsection{Impact of model causality structure.} \label{app:causal_vs_noncausal}
\begin{figure}
    \centering
    \includegraphics[width=\columnwidth]{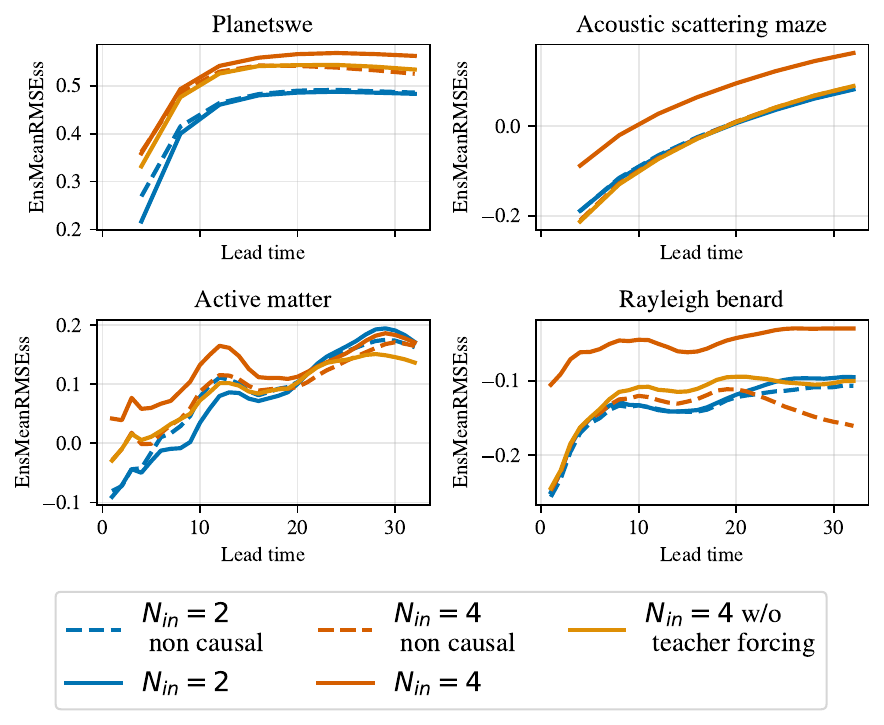}
    \caption{fCRPSsf scores per lead time for 2 and 4 input frames, based on fine-tuning Walrus as a causal or non-causal model. While Walrus can be fine-tuned as a non-causal model with 2 input frames without degradation, we observe that it is sub-optimal compared to causal fine-tuning with 4 input frames, due to the absence of training signal on prefix trajectories.}
    \label{fig:causal_vs_noncausal}
\end{figure}

In Figure \ref{fig:causal_vs_noncausal}, we analyse the impact of using a causal transformer or a non-causal transformer. The causal transformer has a mask where each frame can only attend to previous frames during time-wise attention, which allows the model to be trained with teacher forcing, unlike the non-causal model. The loss signal is therefore richer, with the model predicting the sequence $\hat{x}_{t-(k-1)\tau:t+\tau::\tau}$ from $x_{t-k\tau:t::\tau}$ and receiving a gradient at every frame. Since the pretrained Walrus model was trained as a causal model, we could also expect causal fine-tuning to better transfer knowledge from the pretrained weights. On the other hand, the advantage of the non-causal model is that it has a higher capacity, which is entirely devoted to predicting the next frame $x_{t+1}$ instead of making a sequence prediction $\hat{x}_{t-(k-1)\tau:t+\tau}$. Figure \ref{fig:causal_vs_noncausal} shows that both options are viable when using only two input frames, discarding the hypothesis that causal fine-tuning would better leverage the pretrained weights; however with 4 input frames, causal fine-tuning shows a strong advantage, especially on the \textit{Acoustic Scattering Maze} and \textit{Rayleigh-Bénard} datasets. We show that this advantage is due to the teacher forcing and parallel training on prefix trajectories of inputs: a causal model trained only on predicting the last frame has similar performance to the non-causal model, as shown in Figure \ref{fig:causal_vs_noncausal}.

\subsection{Impact of the number of diffusion steps}
\begin{figure*}
    \centering

    \begin{subfigure}[b]{\linewidth}
        \centering
        \includegraphics[width=\linewidth]{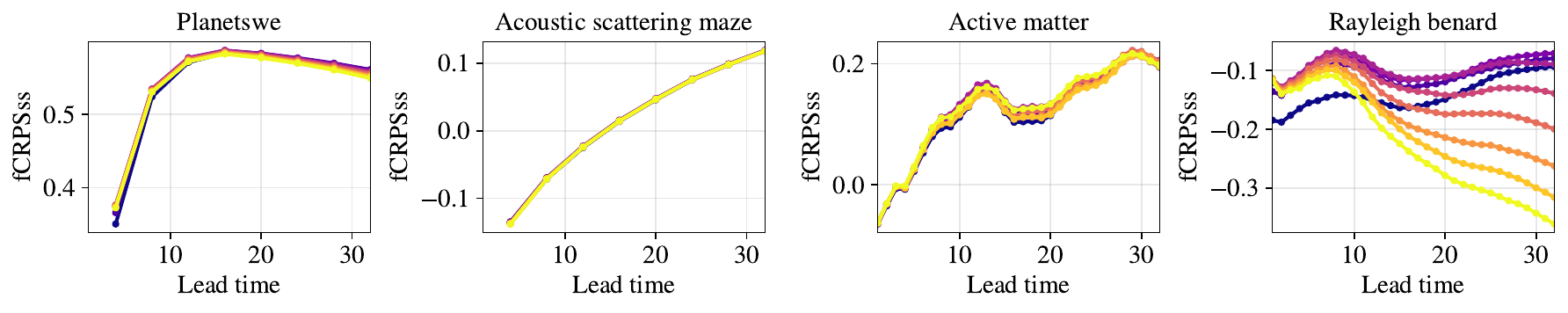}
    \end{subfigure}
    \begin{subfigure}[b]{\linewidth}
        \centering
        \includegraphics[width=\linewidth]{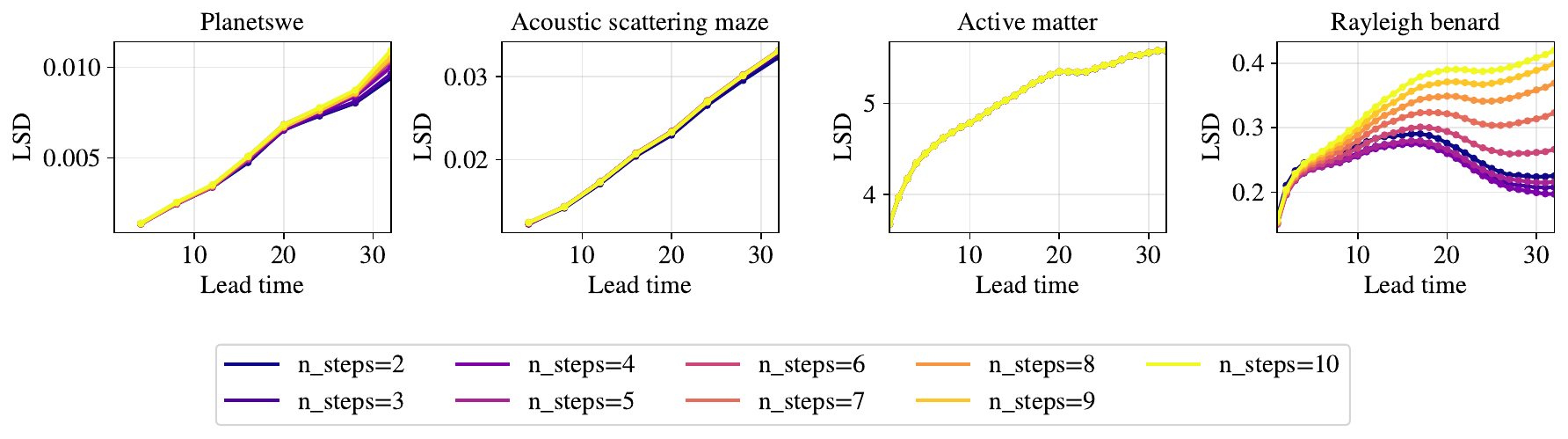}
    \end{subfigure}
    \caption{Impact of the number of diffusion steps on fCRPSss (top) and LSD (bottom). The quality of rollouts increases with the number of diffusion steps up to n=4 steps, then increasing the diffusion steps reduces the performance.}
    \label{fig:diffusion_steps}
\end{figure*}

\begin{figure*}
    \centering

    \begin{subfigure}[b]{\linewidth}
        \centering
        \includegraphics[width=\linewidth]{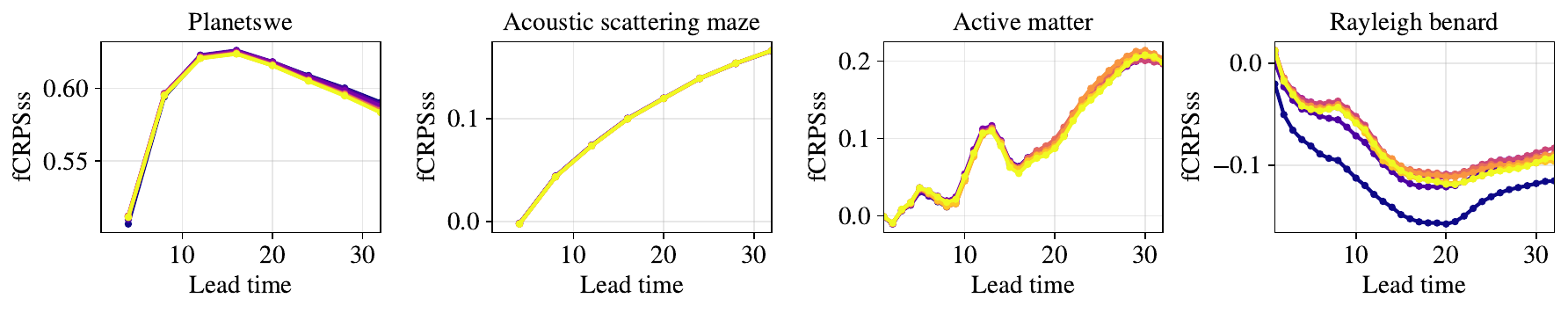}
    \end{subfigure}
    \begin{subfigure}[b]{\linewidth}
        \centering
        \includegraphics[width=\linewidth]{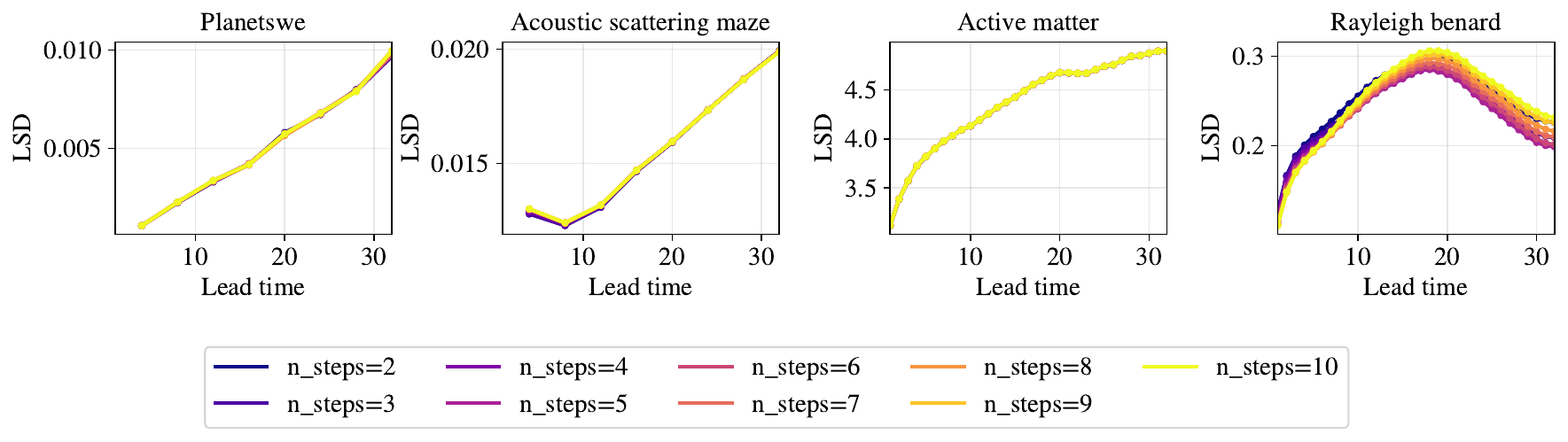}
    \end{subfigure}
    \caption{Impact of the number of diffusion steps, when using GDL. With GDL, the quality of rollouts increases with the number of diffusion steps up to n=6 steps, and the impact of the number of steps is lower.}
    \label{fig:diffusion_steps_gradloss}
\end{figure*}

We study how the number of diffusion steps affects the CRPS along generated trajectories. As shown in Figure~\ref{fig:diffusion_steps}, the number of steps has no significant effect on environments with low stochasticity (\textit{PlanetSWE} and \textit{Acoustic Scattering Maze}), but show a stronger impact on \textit{Rayleigh Bénard}. The number of diffusion steps only increases the quality of generations up to 4 steps, and starts degrading generations after 5 steps. To study this effect, we also display the spectral error (LSD), and observe a similar pattern, the number of steps increasing the LSD with more than 5 steps. As shown in Figure~\ref{fig:diffusion_steps_gradloss}, adding a Gradient Difference Loss to the objective reduces this effect, and the quality of generations is improving up to 6-7 steps with a much lower degrading impact on higher numbers of steps. This suggests that one problem with diffusion models for forecasting could be the creation of high-frequency artifacts worsened by the higher number of model calls.

\subsection{Full table of metrics}

In Tables \ref{tab:table_fcrps} and 
\ref{tab:table_ensrmse}, we show a complete table of metrics for the final configurations of Kastor compared to the finetuning baseline Walrus.

\onecolumn
\begin{longtable}{ll ccc ccc}
\caption{fCRPS per variable and lead time $\tau$ for final Kastor FGN models compared to Walrus finetuning baseline.} \\
\toprule
\endhead
\bottomrule \multicolumn{8}{r}{\textit{Continued on next page}} \\
\endfoot
\bottomrule
\endlastfoot
\multirow{2}{*}{\textbf{Dataset}} & \multirow{2}{*}{\textbf{Variable}} & \multicolumn{3}{c}{\textbf{$\tau = 4$}} & \multicolumn{3}{c}{\textbf{$\tau = 8$}} \\
\cmidrule(lr){3-5} \cmidrule(lr){6-8}
 & & Kastor-L & Kastor-M & Walrus & Kastor-L & Kastor-M & Walrus \\
\midrule
Active Matter & $D_xx$ & 9.11e-06 & 1.05e-05 & 4.64e-05 & 1.38e-05 & 1.66e-05 & 1.26e-04 \\
 & $D_xy$ & 2.64e-03 & 3.35e-03 & 4.28e-03 & 6.83e-03 & 8.16e-03 & 1.22e-02 \\
 & $D_yx$ & 2.28e-03 & 2.93e-03 & 3.82e-03 & 6.99e-03 & 8.99e-03 & 1.23e-02 \\
 & $D_yy$ & 1.58e-03 & 2.03e-03 & 2.41e-03 & 5.09e-03 & 6.27e-03 & 8.35e-03 \\
 & $E_xx$ & 1.59e-03 & 2.01e-03 & 2.45e-03 & 5.10e-03 & 6.29e-03 & 8.34e-03 \\
 & $E_xy$ & 1.59e-03 & 2.01e-03 & 2.45e-03 & 5.10e-03 & 6.29e-03 & 8.34e-03 \\
 & $E_yx$ & 1.58e-03 & 2.03e-03 & 2.41e-03 & 5.10e-03 & 6.28e-03 & 8.35e-03 \\
 & $E_yy$ & 3.05e-03 & 3.96e-03 & 4.71e-03 & 9.32e-03 & 1.12e-02 & 1.57e-02 \\
 & $\text{concentration}$ & 3.14e-03 & 3.94e-03 & 4.87e-03 & 9.29e-03 & 1.14e-02 & 1.57e-02 \\
 & $\text{velocity}_{x}$ & 3.14e-03 & 3.94e-03 & 4.87e-03 & 9.29e-03 & 1.14e-02 & 1.57e-02 \\
 & $\text{velocity}_{y}$ & 3.05e-03 & 3.96e-03 & 4.71e-03 & 9.31e-03 & 1.12e-02 & 1.57e-02 \\
\addlinespace
Rayleigh Benard & $\text{buoyancy}$ & 4.70e-04 & 5.78e-04 & 7.48e-04 & 1.08e-03 & 1.26e-03 & 1.44e-03 \\
 & $\text{pressure}$ & 1.62e-05 & 2.05e-05 & 1.98e-05 & 4.74e-05 & 5.66e-05 & 5.65e-05 \\
 & $\text{velocity}_{x}$ & 1.33e-04 & 1.67e-04 & 2.10e-04 & 4.10e-04 & 4.86e-04 & 5.58e-04 \\
 & $\text{velocity}_{y}$ & 1.22e-04 & 1.58e-04 & 1.97e-04 & 3.96e-04 & 4.76e-04 & 5.52e-04 \\
\addlinespace
Planetswe & $\text{height}$ & 9.94e-02 & 1.08e-01 & 2.02e-01 & 1.70e-01 & 2.00e-01 & 4.59e-01 \\
 & $\text{velocity}_{\theta}$ & 6.44e-03 & 5.99e-03 & 1.48e-02 & 1.19e-02 & 1.22e-02 & 3.47e-02 \\
 & $\text{velocity}_{\phi}$ & 5.60e-03 & 5.29e-03 & 1.32e-02 & 1.07e-02 & 1.09e-02 & 3.19e-02 \\
\addlinespace
Acoustic Scattering Maze & $\text{pressure}$ & 4.58e-04 & 5.43e-04 & 6.77e-04 & 6.21e-04 & 7.26e-04 & 8.94e-04 \\
 & $\text{velocity}_{x}$ & 7.39e-05 & 8.88e-05 & 1.16e-04 & 9.59e-05 & 1.17e-04 & 1.49e-04 \\
 & $\text{velocity}_{y}$ & 7.38e-05 & 8.82e-05 & 1.16e-04 & 9.47e-05 & 1.14e-04 & 1.51e-04 \\
\midrule
\multirow{2}{*}{\textbf{Dataset}} & \multirow{2}{*}{\textbf{Variable}} & \multicolumn{3}{c}{\textbf{$\tau = 12$}} & \multicolumn{3}{c}{\textbf{$\tau = 16$}} \\
\cmidrule(lr){3-5} \cmidrule(lr){6-8}
 & & Kastor-L & Kastor-M & Walrus & Kastor-L & Kastor-M & Walrus \\
\midrule
Active Matter & $D_xx$ & 2.07e-05 & 2.46e-05 & 2.08e-04 & 3.41e-05 & 3.70e-05 & 2.83e-04 \\
 & $D_xy$ & 1.73e-02 & 2.42e-02 & 3.46e-02 & 5.21e-02 & 4.69e-02 & 8.03e-02 \\
 & $D_yx$ & 1.97e-02 & 2.32e-02 & 3.87e-02 & 4.59e-02 & 5.34e-02 & 7.87e-02 \\
 & $D_yy$ & 1.43e-02 & 1.76e-02 & 2.54e-02 & 3.46e-02 & 3.60e-02 & 5.12e-02 \\
 & $E_xx$ & 1.44e-02 & 1.77e-02 & 2.56e-02 & 3.51e-02 & 3.72e-02 & 5.31e-02 \\
 & $E_xy$ & 1.44e-02 & 1.77e-02 & 2.56e-02 & 3.51e-02 & 3.72e-02 & 5.31e-02 \\
 & $E_yx$ & 1.44e-02 & 1.76e-02 & 2.54e-02 & 3.46e-02 & 3.60e-02 & 5.12e-02 \\
 & $E_yy$ & 2.51e-02 & 3.11e-02 & 4.56e-02 & 5.81e-02 & 5.94e-02 & 8.66e-02 \\
 & $\text{concentration}$ & 2.53e-02 & 3.09e-02 & 4.61e-02 & 6.05e-02 & 6.26e-02 & 8.98e-02 \\
 & $\text{velocity}_{x}$ & 2.53e-02 & 3.09e-02 & 4.61e-02 & 6.05e-02 & 6.26e-02 & 8.99e-02 \\
 & $\text{velocity}_{y}$ & 2.51e-02 & 3.11e-02 & 4.56e-02 & 5.81e-02 & 5.94e-02 & 8.66e-02 \\
\addlinespace
Rayleigh Benard & $\text{buoyancy}$ & 1.84e-03 & 2.15e-03 & 2.33e-03 & 2.92e-03 & 3.41e-03 & 3.61e-03 \\
 & $\text{pressure}$ & 1.07e-04 & 1.28e-04 & 1.29e-04 & 2.13e-04 & 2.50e-04 & 2.48e-04 \\
 & $\text{velocity}_{x}$ & 9.00e-04 & 1.04e-03 & 1.12e-03 & 1.67e-03 & 1.92e-03 & 1.99e-03 \\
 & $\text{velocity}_{y}$ & 8.87e-04 & 1.04e-03 & 1.14e-03 & 1.70e-03 & 1.97e-03 & 2.07e-03 \\
\addlinespace
Planetswe & $\text{height}$ & 2.72e-01 & 3.09e-01 & 7.62e-01 & 3.92e-01 & 4.24e-01 & 1.07e+00 \\
 & $\text{velocity}_{\theta}$ & 1.90e-02 & 2.01e-02 & 5.66e-02 & 2.78e-02 & 2.92e-02 & 8.08e-02 \\
 & $\text{velocity}_{\phi}$ & 1.70e-02 & 1.76e-02 & 5.22e-02 & 2.43e-02 & 2.53e-02 & 7.41e-02 \\
\addlinespace
Acoustic Scattering Maze & $\text{pressure}$ & 7.60e-04 & 8.85e-04 & 1.08e-03 & 8.74e-04 & 1.02e-03 & 1.25e-03 \\
 & $\text{velocity}_{x}$ & 1.12e-04 & 1.34e-04 & 1.77e-04 & 1.26e-04 & 1.54e-04 & 2.00e-04 \\
 & $\text{velocity}_{y}$ & 1.12e-04 & 1.34e-04 & 1.79e-04 & 1.26e-04 & 1.50e-04 & 2.02e-04 \\
\midrule
\multirow{2}{*}{\textbf{Dataset}} & \multirow{2}{*}{\textbf{Variable}} & \multicolumn{3}{c}{\textbf{$\tau = 20$}} & \multicolumn{3}{c}{\textbf{$\tau = 24$}} \\
\cmidrule(lr){3-5} \cmidrule(lr){6-8}
 & & Kastor-L & Kastor-M & Walrus & Kastor-L & Kastor-M & Walrus \\
\midrule
Active Matter & $D_xx$ & 5.55e-05 & 5.48e-05 & 3.51e-04 & 7.42e-05 & 7.04e-05 & 4.03e-04 \\
 & $D_xy$ & 9.02e-02 & 8.51e-02 & 1.29e-01 & 1.25e-01 & 1.35e-01 & 1.83e-01 \\
 & $D_yx$ & 8.86e-02 & 9.93e-02 & 1.28e-01 & 1.20e-01 & 1.28e-01 & 1.74e-01 \\
 & $D_yy$ & 6.28e-02 & 6.26e-02 & 8.26e-02 & 8.69e-02 & 8.94e-02 & 1.13e-01 \\
 & $E_xx$ & 6.35e-02 & 6.38e-02 & 8.40e-02 & 8.20e-02 & 8.58e-02 & 1.10e-01 \\
 & $E_xy$ & 6.35e-02 & 6.38e-02 & 8.40e-02 & 8.20e-02 & 8.58e-02 & 1.10e-01 \\
 & $E_yx$ & 6.28e-02 & 6.26e-02 & 8.26e-02 & 8.69e-02 & 8.94e-02 & 1.13e-01 \\
 & $E_yy$ & 1.02e-01 & 9.93e-02 & 1.34e-01 & 1.34e-01 & 1.37e-01 & 1.70e-01 \\
 & $\text{concentration}$ & 1.05e-01 & 1.03e-01 & 1.35e-01 & 1.29e-01 & 1.30e-01 & 1.66e-01 \\
 & $\text{velocity}_{x}$ & 1.05e-01 & 1.03e-01 & 1.35e-01 & 1.29e-01 & 1.30e-01 & 1.66e-01 \\
 & $\text{velocity}_{y}$ & 1.02e-01 & 9.93e-02 & 1.34e-01 & 1.34e-01 & 1.37e-01 & 1.70e-01 \\
\addlinespace
Rayleigh Benard & $\text{buoyancy}$ & 4.60e-03 & 5.35e-03 & 5.71e-03 & 7.41e-03 & 8.32e-03 & 9.13e-03 \\
 & $\text{pressure}$ & 4.37e-04 & 4.97e-04 & 5.07e-04 & 8.79e-04 & 9.70e-04 & 1.07e-03 \\
 & $\text{velocity}_{x}$ & 2.87e-03 & 3.27e-03 & 3.39e-03 & 4.89e-03 & 5.46e-03 & 5.76e-03 \\
 & $\text{velocity}_{y}$ & 3.01e-03 & 3.46e-03 & 3.66e-03 & 5.21e-03 & 5.92e-03 & 6.33e-03 \\
\addlinespace
Planetswe & $\text{height}$ & 5.04e-01 & 5.51e-01 & 1.36e+00 & 6.22e-01 & 6.87e-01 & 1.60e+00 \\
 & $\text{velocity}_{\theta}$ & 3.60e-02 & 3.82e-02 & 1.03e-01 & 4.40e-02 & 4.72e-02 & 1.23e-01 \\
 & $\text{velocity}_{\phi}$ & 3.13e-02 & 3.30e-02 & 9.56e-02 & 3.91e-02 & 4.21e-02 & 1.17e-01 \\
\addlinespace
Acoustic Scattering Maze & $\text{pressure}$ & 9.79e-04 & 1.14e-03 & 1.40e-03 & 1.07e-03 & 1.24e-03 & 1.55e-03 \\
 & $\text{velocity}_{x}$ & 1.37e-04 & 1.65e-04 & 2.22e-04 & 1.48e-04 & 1.78e-04 & 2.43e-04 \\
 & $\text{velocity}_{y}$ & 1.38e-04 & 1.65e-04 & 2.25e-04 & 1.48e-04 & 1.77e-04 & 2.46e-04 \\
\midrule
\multirow{2}{*}{\textbf{Dataset}} & \multirow{2}{*}{\textbf{Variable}} & \multicolumn{3}{c}{\textbf{$\tau = 28$}} & \multicolumn{3}{c}{\textbf{$\tau = 32$}} \\
\cmidrule(lr){3-5} \cmidrule(lr){6-8}
 & & Kastor-L & Kastor-M & Walrus & Kastor-L & Kastor-M & Walrus \\
\midrule
Active Matter & $D_xx$ & 9.01e-05 & 9.15e-05 & 4.51e-04 & 1.16e-04 & 1.10e-04 & 5.05e-04 \\
 & $D_xy$ & 1.56e-01 & 1.73e-01 & 2.18e-01 & 1.94e-01 & 2.11e-01 & 2.62e-01 \\
 & $D_yx$ & 1.54e-01 & 1.56e-01 & 2.15e-01 & 1.92e-01 & 1.86e-01 & 2.43e-01 \\
 & $D_yy$ & 1.06e-01 & 1.11e-01 & 1.42e-01 & 1.30e-01 & 1.30e-01 & 1.60e-01 \\
 & $E_xx$ & 1.06e-01 & 1.14e-01 & 1.41e-01 & 1.31e-01 & 1.30e-01 & 1.67e-01 \\
 & $E_xy$ & 1.06e-01 & 1.14e-01 & 1.41e-01 & 1.31e-01 & 1.30e-01 & 1.67e-01 \\
 & $E_yx$ & 1.06e-01 & 1.11e-01 & 1.42e-01 & 1.30e-01 & 1.30e-01 & 1.60e-01 \\
 & $E_yy$ & 1.53e-01 & 1.58e-01 & 1.95e-01 & 1.82e-01 & 1.82e-01 & 2.14e-01 \\
 & $\text{concentration}$ & 1.53e-01 & 1.61e-01 & 1.96e-01 & 1.85e-01 & 1.82e-01 & 2.25e-01 \\
 & $\text{velocity}_{x}$ & 1.53e-01 & 1.61e-01 & 1.96e-01 & 1.85e-01 & 1.82e-01 & 2.25e-01 \\
 & $\text{velocity}_{y}$ & 1.53e-01 & 1.58e-01 & 1.95e-01 & 1.82e-01 & 1.82e-01 & 2.14e-01 \\
\addlinespace
Rayleigh Benard & $\text{buoyancy}$ & 1.19e-02 & 1.30e-02 & 1.43e-02 & 1.86e-02 & 2.02e-02 & 2.15e-02 \\
 & $\text{pressure}$ & 1.65e-03 & 1.79e-03 & 1.99e-03 & 3.01e-03 & 3.28e-03 & 3.48e-03 \\
 & $\text{velocity}_{x}$ & 8.52e-03 & 9.23e-03 & 9.93e-03 & 1.45e-02 & 1.56e-02 & 1.65e-02 \\
 & $\text{velocity}_{y}$ & 8.88e-03 & 9.83e-03 & 1.06e-02 & 1.48e-02 & 1.61e-02 & 1.70e-02 \\
\addlinespace
Planetswe & $\text{height}$ & 7.06e-01 & 8.04e-01 & 1.80e+00 & 8.05e-01 & 9.19e-01 & 2.11e+00 \\
 & $\text{velocity}_{\theta}$ & 5.30e-02 & 5.71e-02 & 1.41e-01 & 6.10e-02 & 6.64e-02 & 1.59e-01 \\
 & $\text{velocity}_{\phi}$ & 4.79e-02 & 5.20e-02 & 1.38e-01 & 5.67e-02 & 6.26e-02 & 1.57e-01 \\
\addlinespace
Acoustic Scattering Maze & $\text{pressure}$ & 1.16e-03 & 1.34e-03 & 1.68e-03 & 1.23e-03 & 1.43e-03 & 1.80e-03 \\
 & $\text{velocity}_{x}$ & 1.57e-04 & 1.89e-04 & 2.61e-04 & 1.66e-04 & 2.00e-04 & 2.79e-04 \\
 & $\text{velocity}_{y}$ & 1.58e-04 & 1.89e-04 & 2.65e-04 & 1.68e-04 & 2.00e-04 & 2.82e-04 \\
\label{tab:table_fcrps}
\end{longtable}
\twocolumn

\onecolumn
\begin{longtable}{ll ccc ccc}
\caption{Ensemble Mean RMSE per variable and lead time $\tau$ for final Kastor FGN models compared to Walrus finetuning baseline.} \\
\toprule
\endhead
\bottomrule \multicolumn{8}{r}{\textit{Continued on next page}} \\
\endfoot
\bottomrule
\endlastfoot
\multirow{2}{*}{\textbf{Dataset}} & \multirow{2}{*}{\textbf{Variable}} & \multicolumn{3}{c}{\textbf{$\tau = 4$}} & \multicolumn{3}{c}{\textbf{$\tau = 8$}} \\
\cmidrule(lr){3-5} \cmidrule(lr){6-8}
 & & Kastor-L & Kastor-M & Walrus & Kastor-L & Kastor-M & Walrus \\
\midrule
Active Matter & $D_xx$ & 1.57e-05 & 1.82e-05 & 8.59e-05 & 2.40e-05 & 2.87e-05 & 2.23e-04 \\
 & $D_xy$ & 4.35e-03 & 5.47e-03 & 6.80e-03 & 1.21e-02 & 1.44e-02 & 2.01e-02 \\
 & $D_yx$ & 4.03e-03 & 5.01e-03 & 6.28e-03 & 1.23e-02 & 1.52e-02 & 2.07e-02 \\
 & $D_yy$ & 3.79e-03 & 4.73e-03 & 5.50e-03 & 1.17e-02 & 1.40e-02 & 1.84e-02 \\
 & $E_xx$ & 3.89e-03 & 4.72e-03 & 5.97e-03 & 1.18e-02 & 1.44e-02 & 1.86e-02 \\
 & $E_xy$ & 3.89e-03 & 4.72e-03 & 5.97e-03 & 1.18e-02 & 1.44e-02 & 1.86e-02 \\
 & $E_yx$ & 3.79e-03 & 4.72e-03 & 5.50e-03 & 1.17e-02 & 1.40e-02 & 1.84e-02 \\
 & $E_yy$ & 7.22e-03 & 8.96e-03 & 1.08e-02 & 2.13e-02 & 2.49e-02 & 3.33e-02 \\
 & $\text{concentration}$ & 7.36e-03 & 8.84e-03 & 1.11e-02 & 2.10e-02 & 2.54e-02 & 3.39e-02 \\
 & $\text{velocity}_{x}$ & 7.36e-03 & 8.84e-03 & 1.11e-02 & 2.10e-02 & 2.54e-02 & 3.39e-02 \\
 & $\text{velocity}_{y}$ & 7.22e-03 & 8.97e-03 & 1.08e-02 & 2.13e-02 & 2.49e-02 & 3.33e-02 \\
\addlinespace
Rayleigh Benard & $\text{buoyancy}$ & 2.20e-03 & 2.69e-03 & 3.13e-03 & 4.83e-03 & 5.55e-03 & 5.87e-03 \\
 & $\text{pressure}$ & 6.37e-05 & 7.52e-05 & 7.60e-05 & 1.70e-04 & 1.95e-04 & 1.94e-04 \\
 & $\text{velocity}_{x}$ & 5.78e-04 & 7.20e-04 & 8.28e-04 & 1.63e-03 & 1.92e-03 & 2.05e-03 \\
 & $\text{velocity}_{y}$ & 5.72e-04 & 7.25e-04 & 8.42e-04 & 1.69e-03 & 1.99e-03 & 2.15e-03 \\
\addlinespace
Planetswe & $\text{height}$ & 1.77e-01 & 2.03e-01 & 3.25e-01 & 3.03e-01 & 3.69e-01 & 7.14e-01 \\
 & $\text{velocity}_{\theta}$ & 1.26e-02 & 1.31e-02 & 2.71e-02 & 2.31e-02 & 2.55e-02 & 6.12e-02 \\
 & $\text{velocity}_{\phi}$ & 1.11e-02 & 1.16e-02 & 2.44e-02 & 2.09e-02 & 2.30e-02 & 5.61e-02 \\
\addlinespace
Acoustic Scattering Maze & $\text{pressure}$ & 2.90e-03 & 3.44e-03 & 4.60e-03 & 3.76e-03 & 4.28e-03 & 5.30e-03 \\
 & $\text{velocity}_{x}$ & 4.24e-04 & 5.06e-04 & 6.84e-04 & 4.96e-04 & 5.98e-04 & 7.85e-04 \\
 & $\text{velocity}_{y}$ & 4.23e-04 & 4.99e-04 & 6.84e-04 & 4.90e-04 & 5.86e-04 & 7.95e-04 \\
\midrule
\multirow{2}{*}{\textbf{Dataset}} & \multirow{2}{*}{\textbf{Variable}} & \multicolumn{3}{c}{\textbf{$\tau = 12$}} & \multicolumn{3}{c}{\textbf{$\tau = 16$}} \\
\cmidrule(lr){3-5} \cmidrule(lr){6-8}
 & & Kastor-L & Kastor-M & Walrus & Kastor-L & Kastor-M & Walrus \\
\midrule
Active Matter & $D_xx$ & 3.76e-05 & 4.66e-05 & 3.63e-04 & 6.38e-05 & 6.96e-05 & 4.95e-04 \\
 & $D_xy$ & 3.17e-02 & 4.23e-02 & 5.65e-02 & 8.93e-02 & 8.51e-02 & 1.30e-01 \\
 & $D_yx$ & 3.59e-02 & 4.44e-02 & 6.28e-02 & 8.09e-02 & 9.32e-02 & 1.28e-01 \\
 & $D_yy$ & 3.28e-02 & 3.84e-02 & 5.15e-02 & 6.94e-02 & 7.39e-02 & 9.68e-02 \\
 & $E_xx$ & 3.22e-02 & 3.90e-02 & 5.20e-02 & 7.10e-02 & 7.62e-02 & 1.00e-01 \\
 & $E_xy$ & 3.22e-02 & 3.90e-02 & 5.20e-02 & 7.10e-02 & 7.62e-02 & 1.00e-01 \\
 & $E_yx$ & 3.28e-02 & 3.84e-02 & 5.15e-02 & 6.94e-02 & 7.39e-02 & 9.68e-02 \\
 & $E_yy$ & 5.54e-02 & 6.67e-02 & 8.89e-02 & 1.16e-01 & 1.21e-01 & 1.61e-01 \\
 & $\text{concentration}$ & 5.60e-02 & 6.68e-02 & 9.12e-02 & 1.20e-01 & 1.27e-01 & 1.66e-01 \\
 & $\text{velocity}_{x}$ & 5.60e-02 & 6.68e-02 & 9.12e-02 & 1.20e-01 & 1.27e-01 & 1.66e-01 \\
 & $\text{velocity}_{y}$ & 5.54e-02 & 6.67e-02 & 8.89e-02 & 1.16e-01 & 1.21e-01 & 1.61e-01 \\
\addlinespace
Rayleigh Benard & $\text{buoyancy}$ & 7.64e-03 & 8.81e-03 & 9.04e-03 & 1.12e-02 & 1.28e-02 & 1.30e-02 \\
 & $\text{pressure}$ & 3.48e-04 & 4.08e-04 & 4.00e-04 & 6.35e-04 & 7.30e-04 & 7.07e-04 \\
 & $\text{velocity}_{x}$ & 3.26e-03 & 3.74e-03 & 3.81e-03 & 5.59e-03 & 6.34e-03 & 6.28e-03 \\
 & $\text{velocity}_{y}$ & 3.41e-03 & 3.94e-03 & 4.07e-03 & 6.00e-03 & 6.85e-03 & 6.87e-03 \\
\addlinespace
Planetswe & $\text{height}$ & 4.72e-01 & 5.58e-01 & 1.16e+00 & 6.67e-01 & 7.52e-01 & 1.61e+00 \\
 & $\text{velocity}_{\theta}$ & 3.61e-02 & 4.04e-02 & 9.74e-02 & 5.07e-02 & 5.68e-02 & 1.36e-01 \\
 & $\text{velocity}_{\phi}$ & 3.24e-02 & 3.62e-02 & 8.99e-02 & 4.54e-02 & 5.07e-02 & 1.25e-01 \\
\addlinespace
Acoustic Scattering Maze & $\text{pressure}$ & 4.43e-03 & 4.94e-03 & 6.00e-03 & 4.89e-03 & 5.41e-03 & 6.52e-03 \\
 & $\text{velocity}_{x}$ & 5.38e-04 & 6.47e-04 & 8.69e-04 & 5.77e-04 & 6.98e-04 & 9.27e-04 \\
 & $\text{velocity}_{y}$ & 5.37e-04 & 6.41e-04 & 8.76e-04 & 5.73e-04 & 6.80e-04 & 9.35e-04 \\
\midrule
\multirow{2}{*}{\textbf{Dataset}} & \multirow{2}{*}{\textbf{Variable}} & \multicolumn{3}{c}{\textbf{$\tau = 20$}} & \multicolumn{3}{c}{\textbf{$\tau = 24$}} \\
\cmidrule(lr){3-5} \cmidrule(lr){6-8}
 & & Kastor-L & Kastor-M & Walrus & Kastor-L & Kastor-M & Walrus \\
\midrule
Active Matter & $D_xx$ & 1.04e-04 & 1.07e-04 & 6.14e-04 & 1.39e-04 & 1.36e-04 & 7.12e-04 \\
 & $D_xy$ & 1.58e-01 & 1.54e-01 & 2.14e-01 & 2.23e-01 & 2.40e-01 & 3.09e-01 \\
 & $D_yx$ & 1.57e-01 & 1.73e-01 & 2.14e-01 & 2.22e-01 & 2.29e-01 & 2.86e-01 \\
 & $D_yy$ & 1.21e-01 & 1.22e-01 & 1.52e-01 & 1.65e-01 & 1.70e-01 & 2.04e-01 \\
 & $E_xx$ & 1.20e-01 & 1.22e-01 & 1.53e-01 & 1.57e-01 & 1.64e-01 & 2.00e-01 \\
 & $E_xy$ & 1.20e-01 & 1.22e-01 & 1.53e-01 & 1.57e-01 & 1.64e-01 & 2.00e-01 \\
 & $E_yx$ & 1.21e-01 & 1.22e-01 & 1.52e-01 & 1.65e-01 & 1.70e-01 & 2.04e-01 \\
 & $E_yy$ & 1.92e-01 & 1.92e-01 & 2.43e-01 & 2.51e-01 & 2.56e-01 & 3.04e-01 \\
 & $\text{concentration}$ & 1.96e-01 & 1.96e-01 & 2.43e-01 & 2.42e-01 & 2.46e-01 & 2.98e-01 \\
 & $\text{velocity}_{x}$ & 1.96e-01 & 1.96e-01 & 2.43e-01 & 2.42e-01 & 2.46e-01 & 2.98e-01 \\
 & $\text{velocity}_{y}$ & 1.92e-01 & 1.92e-01 & 2.43e-01 & 2.51e-01 & 2.57e-01 & 3.04e-01 \\
\addlinespace
Rayleigh Benard & $\text{buoyancy}$ & 1.61e-02 & 1.82e-02 & 1.87e-02 & 2.33e-02 & 2.54e-02 & 2.64e-02 \\
 & $\text{pressure}$ & 1.14e-03 & 1.29e-03 & 1.26e-03 & 2.04e-03 & 2.24e-03 & 2.29e-03 \\
 & $\text{velocity}_{x}$ & 8.80e-03 & 9.86e-03 & 9.80e-03 & 1.36e-02 & 1.50e-02 & 1.51e-02 \\
 & $\text{velocity}_{y}$ & 9.65e-03 & 1.09e-02 & 1.10e-02 & 1.52e-02 & 1.70e-02 & 1.72e-02 \\
\addlinespace
Planetswe & $\text{height}$ & 8.56e-01 & 9.71e-01 & 2.04e+00 & 1.06e+00 & 1.22e+00 & 2.47e+00 \\
 & $\text{velocity}_{\theta}$ & 6.49e-02 & 7.29e-02 & 1.71e-01 & 7.98e-02 & 9.02e-02 & 2.04e-01 \\
 & $\text{velocity}_{\phi}$ & 5.78e-02 & 6.52e-02 & 1.59e-01 & 7.14e-02 & 8.14e-02 & 1.93e-01 \\
\addlinespace
Acoustic Scattering Maze & $\text{pressure}$ & 5.28e-03 & 5.81e-03 & 6.96e-03 & 5.59e-03 & 6.14e-03 & 7.34e-03 \\
 & $\text{velocity}_{x}$ & 6.00e-04 & 7.20e-04 & 9.83e-04 & 6.19e-04 & 7.46e-04 & 1.03e-03 \\
 & $\text{velocity}_{y}$ & 6.03e-04 & 7.14e-04 & 1.00e-03 & 6.21e-04 & 7.40e-04 & 1.05e-03 \\
\midrule
\multirow{2}{*}{\textbf{Dataset}} & \multirow{2}{*}{\textbf{Variable}} & \multicolumn{3}{c}{\textbf{$\tau = 28$}} & \multicolumn{3}{c}{\textbf{$\tau = 32$}} \\
\cmidrule(lr){3-5} \cmidrule(lr){6-8}
 & & Kastor-L & Kastor-M & Walrus & Kastor-L & Kastor-M & Walrus \\
\midrule
Active Matter & $D_xx$ & 1.68e-04 & 1.76e-04 & 8.02e-04 & 2.09e-04 & 2.10e-04 & 9.02e-04 \\
 & $D_xy$ & 2.80e-01 & 3.11e-01 & 3.73e-01 & 3.48e-01 & 3.80e-01 & 4.47e-01 \\
 & $D_yx$ & 2.76e-01 & 2.87e-01 & 3.65e-01 & 3.37e-01 & 3.45e-01 & 4.14e-01 \\
 & $D_yy$ & 1.99e-01 & 2.10e-01 & 2.51e-01 & 2.39e-01 & 2.43e-01 & 2.83e-01 \\
 & $E_xx$ & 1.99e-01 & 2.12e-01 & 2.50e-01 & 2.42e-01 & 2.43e-01 & 2.93e-01 \\
 & $E_xy$ & 1.99e-01 & 2.12e-01 & 2.50e-01 & 2.42e-01 & 2.43e-01 & 2.93e-01 \\
 & $E_yx$ & 1.99e-01 & 2.10e-01 & 2.51e-01 & 2.39e-01 & 2.43e-01 & 2.83e-01 \\
 & $E_yy$ & 2.85e-01 & 2.95e-01 & 3.50e-01 & 3.35e-01 & 3.40e-01 & 3.83e-01 \\
 & $\text{concentration}$ & 2.85e-01 & 3.01e-01 & 3.50e-01 & 3.37e-01 & 3.39e-01 & 3.98e-01 \\
 & $\text{velocity}_{x}$ & 2.85e-01 & 3.01e-01 & 3.50e-01 & 3.37e-01 & 3.39e-01 & 3.98e-01 \\
 & $\text{velocity}_{y}$ & 2.85e-01 & 2.95e-01 & 3.50e-01 & 3.35e-01 & 3.40e-01 & 3.83e-01 \\
\addlinespace
Rayleigh Benard & $\text{buoyancy}$ & 3.32e-02 & 3.54e-02 & 3.66e-02 & 4.72e-02 & 5.02e-02 & 5.04e-02 \\
 & $\text{pressure}$ & 3.53e-03 & 3.82e-03 & 3.89e-03 & 6.16e-03 & 6.65e-03 & 6.58e-03 \\
 & $\text{velocity}_{x}$ & 2.13e-02 & 2.28e-02 & 2.32e-02 & 3.31e-02 & 3.52e-02 & 3.52e-02 \\
 & $\text{velocity}_{y}$ & 2.35e-02 & 2.56e-02 & 2.59e-02 & 3.61e-02 & 3.89e-02 & 3.87e-02 \\
\addlinespace
Planetswe & $\text{height}$ & 1.23e+00 & 1.45e+00 & 2.89e+00 & 1.42e+00 & 1.67e+00 & 3.36e+00 \\
 & $\text{velocity}_{\theta}$ & 9.59e-02 & 1.09e-01 & 2.37e-01 & 1.10e-01 & 1.27e-01 & 2.68e-01 \\
 & $\text{velocity}_{\phi}$ & 8.68e-02 & 9.93e-02 & 2.26e-01 & 1.02e-01 & 1.18e-01 & 2.58e-01 \\
\addlinespace
Acoustic Scattering Maze & $\text{pressure}$ & 5.85e-03 & 6.40e-03 & 7.65e-03 & 6.08e-03 & 6.65e-03 & 7.97e-03 \\
 & $\text{velocity}_{x}$ & 6.40e-04 & 7.68e-04 & 1.07e-03 & 6.55e-04 & 7.87e-04 & 1.11e-03 \\
 & $\text{velocity}_{y}$ & 6.45e-04 & 7.68e-04 & 1.10e-03 & 6.66e-04 & 7.92e-04 & 1.14e-03 \\
\label{tab:table_ensrmse}
\end{longtable}
\twocolumn

\end{document}